\documentclass{article}

\usepackage{microtype}
\usepackage{graphicx}
\usepackage{subfigure}
\usepackage{booktabs} 

\usepackage{multirow}
\usepackage{hyperref}
\usepackage{graphicx}
\usepackage{pythonhighlight}

\usepackage{amsmath,amsthm,amsfonts,bm}

\newtheorem{theorem}{Theorem}[section]

\newtheorem{proposition}[theorem]{Proposition}
\newtheorem{lemma}[theorem]{Lemma}

\usepackage[capitalize]{cleveref}
\crefname{section}{Sec.}{Secs.}
\Crefname{section}{Section}{Sections}
\Crefname{table}{Table}{Tables}
\crefname{table}{Tab.}{Tabs.}
\crefname{proposition}{Proposition}{Propositions}

\def\eqref#1{equation~\ref{#1}}

\def\1{\bm{1}}

\DeclareMathAlphabet{\mathsfit}{\encodingdefault}{\sfdefault}{m}{sl}
\SetMathAlphabet{\mathsfit}{bold}{\encodingdefault}{\sfdefault}{bx}{n}

\def\gH{{\mathcal{H}}}

\def\sP{{\mathbb{P}}}

\def\sR{{\mathbb{R}}}
\def\sS{{\mathbb{S}}}

\newcommand{\E}{\mathbb{E}}

\usepackage[accepted]{icml2022}

\icmltitlerunning{Prototype-Anchored Learning for Learning with Imperfect Annotations}

\begin{document}

\twocolumn[
\icmltitle{Prototype-Anchored Learning for Learning with Imperfect Annotations}




\begin{icmlauthorlist}
\icmlauthor{Xiong Zhou}{hit}
\icmlauthor{Xianming Liu}{hit,pcl}
\icmlauthor{Deming Zhai}{hit}
\icmlauthor{Junjun Jiang}{hit,pcl}
\icmlauthor{Xin Gao}{kaust,pcl}
\icmlauthor{Xiangyang Ji}{thu}
\end{icmlauthorlist}

\icmlaffiliation{hit}{Harbin Institute of Technology, Harbin, China}
\icmlaffiliation{pcl}{Peng Cheng Laboratory, Shenzhen, China}
\icmlaffiliation{kaust}{King Abdullah University of Science and Technology, Thuwal, Saudi Arabia}
\icmlaffiliation{thu}{Tsinghua University, Beijing, China}

\icmlcorrespondingauthor{Xianming Liu}{csxm@hit.edu.cn}

\icmlkeywords{Machine Learning, ICML}

\vskip 0.3in
]


\printAffiliationsAndNotice{} 

\begin{abstract}
The success of deep neural networks greatly relies on the availability of large amounts of high-quality annotated data, which however are difficult or expensive to obtain. The resulting labels may be class imbalanced, noisy or human biased. It is challenging to learn unbiased classification models from imperfectly annotated datasets, on which we usually suffer from overfitting or underfitting. In this work, we thoroughly investigate the popular softmax loss and margin-based loss, and offer a feasible approach to tighten the generalization error bound by maximizing the minimal sample margin. We further derive the optimality condition for this purpose, which indicates how the class prototypes should be anchored. Motivated by theoretical analysis, we propose a simple yet effective method, namely prototype-anchored learning (PAL), which can be easily incorporated into various learning-based classification schemes to handle imperfect annotation. We verify the effectiveness of PAL on class-imbalanced learning and noise-tolerant learning by extensive experiments on synthetic and real-world datasets.

\end{abstract}

\section{Introduction}

Over the past few years, deep neural networks have achieved impressive performance in various tasks of computer vision \cite{goodfellow2016deep}, such as image classification, segmentation, object detection, etc. One critical factor that attributes to the success of deep learning is the availability of large amounts of annotated training data. However, in many real-world scenarios, it would be difficult to attain perfect supervision for fully supervised learning due to a variety of limitations, such as inaccurate human annotation \cite{Han2020ASO}, difficulty in collecting enough samples for all classes \cite{Zhang2021DeepLL}, etc. In this work, we specially consider the task of image classification under two imperfect annotation scenarios that are commonly encountered in practical applications:

\textbf{Imbalanced Annotation.} Real-world data usually exhibits highly-skewed class distribution, due to the nature that some classes are easy to collect sufficient examples while many classes are associated with only a few ones \cite{Zhang2021DeepLL}. For instance, in plant disease classification, it is comparatively easier to obtain images for common diseases than rare diseases during data collection. This results in that a few majority classes contain most of the data while massive minority classes only contain a scarce amount of instances. This issue would make naive learning biased towards the majority classes while with poor accuracy on the minority ones, posing a challenge for generalization.

\textbf{Inaccurate Annotation.} Data annotation tasks that require a high level of competency make the acquisition of hand-labeled data time-consuming and expensive. One common and economical way to collect large amounts of training data is through online queries \cite{liu2011noise} or crowdsourcing \cite{arpit2017closer}, which however inevitably introduce noisy labels due to that no domain expertise is involved and a variety of human biases exist. This issue would lead deep neural networks to overfit mislabeled data, which seriously hampers the generalization ability of neural networks.

Learning unbiased classification models from imperfectly annotated datasets is a challenging problem. Specifically, for learning with imbalanced annotation, the main concern is how to prevent overfitting to the majority classes and achieve a better trade-off between the majority and minority classes. The difficulty lies in that the spanned feature space of the majority classes is spread while that of the minority classes is concentrated, which hampers the generalization of the learned classifier boundaries due to the lack of intra-class compactness and inter-class separation, as shown in \cref{fig:imb-CE}.
For learning with noisy labels, the key is to prevent overfitting to mislabeled data, as well as achieve a good trade-off between robustness and the fitting ability. The softmax loss enjoys the advantage of fitting ability due to that the optimizer puts more emphasis on ambiguous samples, which however suffers from the overfitting effect on noisy labels. The symmetric losses, such as MAE \cite{symmetric} and normalized losses \cite{ma2020normalized}, have theoretical guarantees to be noise-tolerant, which however usually suffer from underfitting on complicated datasets. 

In this work, we thoroughly investigate the softmax loss and its variant---margin-based losses, and claim that to tighten the generalization error bound one can maximize the minimal sample margin $\gamma_{\min}$ in \cref{samle-margin-definition}.  We further derive the optimality condition of prototypes to obtain the largest $\gamma_{\min}$, which indicates that the class prototypes $\bm{w}_1,...,\bm{w}_k$ should satisfy $\bm{w}_i^{\top}\bm{w}_j=\frac{-1}{k-1}$, $\forall i\neq j$, where $k$ is the class number. Motivated by the theoretical analysis, we propose a simple yet effective method, namely \textit{prototype-anchored learning} (PAL).
Specifically, we derive a classifier composed of anchored prototypes, which are predefined by a simple implementation that only requires knowing the number of classes and the feature dimension. Due to its simplicity, PAL can be easily incorporated into various learning-based classification schemes. For class-imbalanced learning, PAL can be used in the feature representation learning of decoupling methods \cite{kang2019decoupling,zhang2021disalign}, as well as in tandem with other performance-boosting approaches and modules. For noise-tolerant learning, the application of PAL is not such straightforward as class-imbalanced learning. 
Instead, we suggest the \textit{feature normalized and prototype-anchored learning} (FNPAL) strategy, which can be combined with traditional losses to boost their ability in handling noisy labels. Extensive experiments are provided to demonstrate the superiority of PAL in handling imperfect annotations.

Our main contributions are highlighted as follows:
\begin{itemize}
    \item We propose a theoretically sound, simple yet effective scheme---prototype-anchored learning, which can be easily embedded into various learning based classification under imperfect annotations.
    \item For class-imbalanced learning, we theoretically prove that PAL can implicitly guarantee balanced feature representations,  and empirically verify that PAL can combine with existing methods to boost their performance significantly.
    
    \item For noise-tolerant learning, we extend the classical symmetric condition to a more general theorem and reveals that PAL can lead to a tighter bound than that without PAL. We provide extensive experimental results to demonstrate its superiority.
    
\end{itemize}

\section{Related Works}
\subsection{Class-imbalanced Learning}
A mainstream paradigm in class-imbalanced learning is class re-balancing, which seeks to balance the training sample numbers of different classes, such as re-sampling \cite{  han2005borderline, liu2008exploratory}, re-weighting \cite{huang2016learning, pmlr-v80-ren18a, cui2019class, byrd2019effect, shu2019meta}, and logit adjustment \cite{ Ren2020balms, hong2021disentangling, menon2020long}. However, the main drawback of class re-balancing methods is that most of them usually improve tail-class performance while impairing head-class performance \cite{zhang2021deep}. Recently, two-stage approaches \cite{cao2019learning, kang2019decoupling, BBN, zhong2021improving} composed of representation learning and classifier learning have achieved significant improvement compared to one-stage methods. For the two-stage learning framework, \citet{kang2019decoupling} and \citet{BBN} found that instance-balanced sampling gives the best and most generalizable representation while explicit class-balanced strategies may introduce adverse effects. However, balanced representations and decision boundaries cannot be guaranteed by instance-balanced sampling, while they being more balanced would improve the performance. For instance, \citet{zhong2021improving} found that mixup can benefit the learning of minority classes and remedy over-confidence by balancing classifier weight norms. In this work, we theoretically propose an implicit re-balancing method to achieve balanced representations.

\subsection{Noise-tolerant Learning}
Noise-tolerant learning, or called learning with noisy labels,  aims to train a robust model in the presence of noisy labels. To alleviate the impact of label noise, one popular research line is to design robust loss functions. This approach has been pursued in a large body of work \cite{nonconvex, Wang2019ImprovingMA, liu2020peer, Lyu2020CurriculumLR, menon2020can, Feng2020CanCE} that embraces new loss functions, especially symmetric losses and their variants \cite{unhingedloss, symmetric, gce, wang2019sce, ma2020normalized, zhou2021asymmetric}. However, the fitting ability of the existing symmetric loss functions is restricted by the symmetric condition \cite{gce} such that other strategies to enhance the fitting ability are necessary \cite{BERAUC, ma2020normalized, zhou2021learning}. In a sense, the existing methods of learning with noisy labels mainly seeks a trade-off between fitting ability and robustness.

\begin{figure*}[!h]
  \centering
  \subfigure[]{
    \label{fig:CE}
    \includegraphics[scale=0.12]{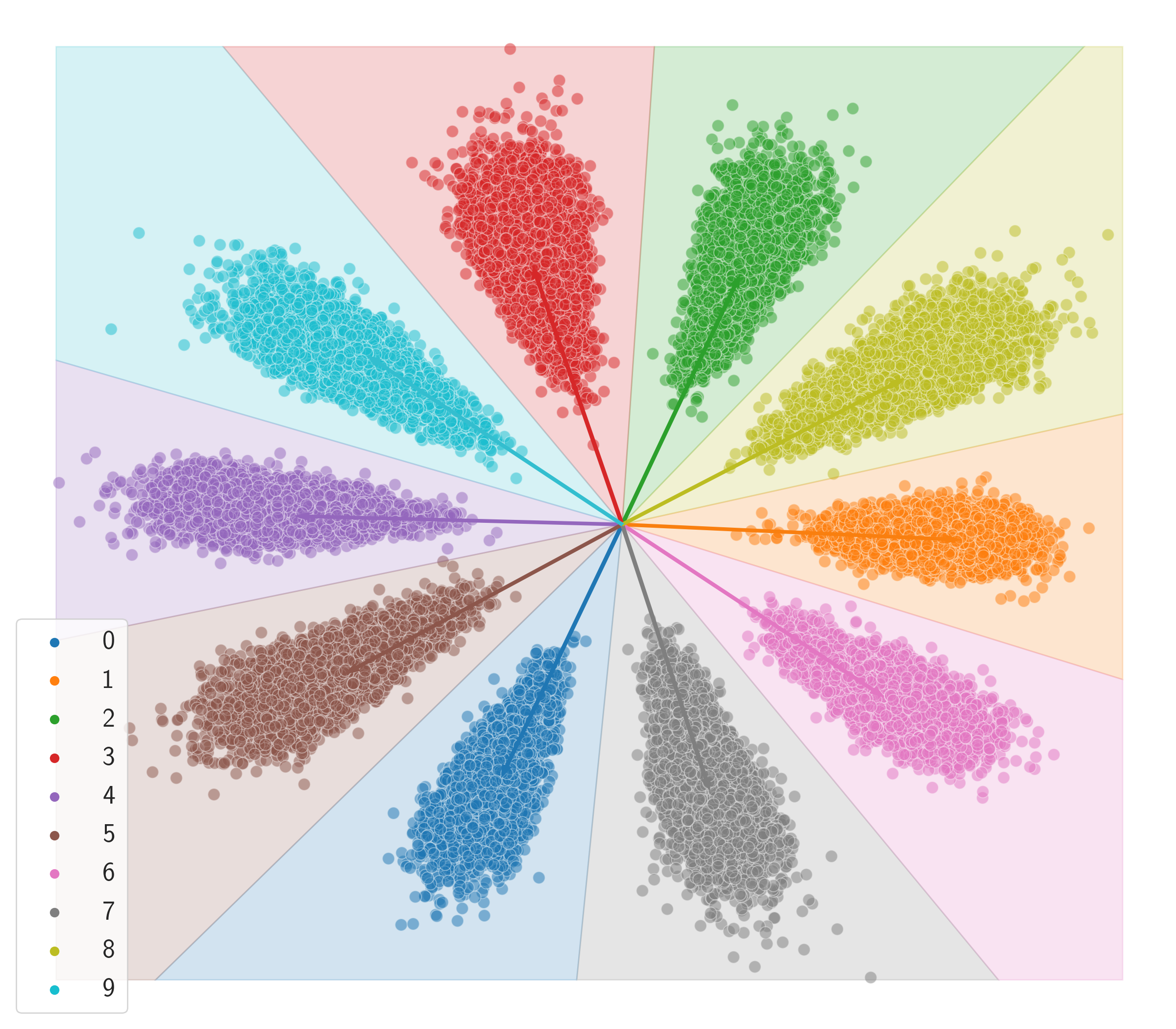}
  }
  \subfigure[]{
    \label{fig:imb-CE}
    \includegraphics[scale=0.12]{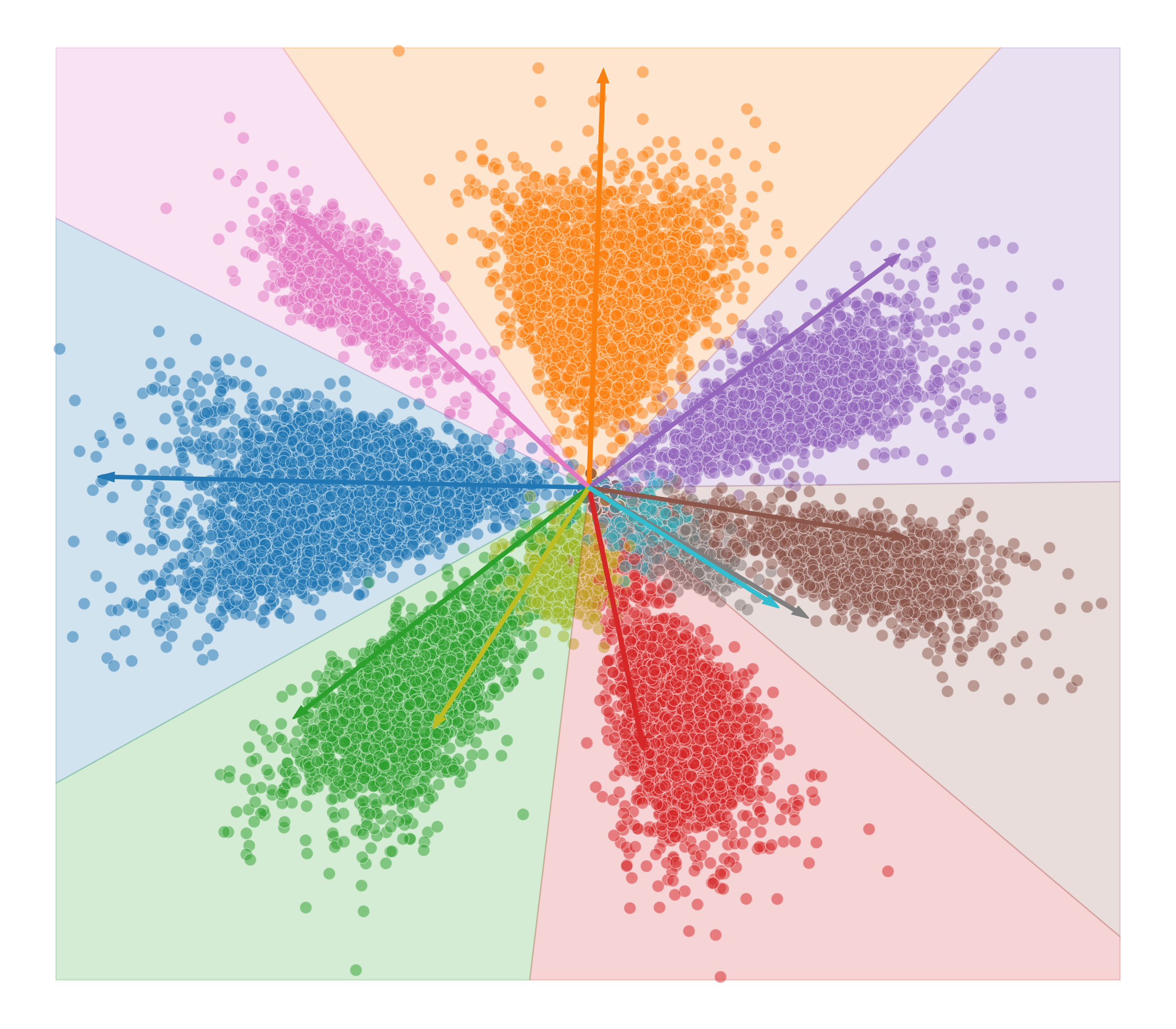}
  }
  \subfigure[]{
    \label{fig:imb-Norm}
    \includegraphics[scale=0.12]{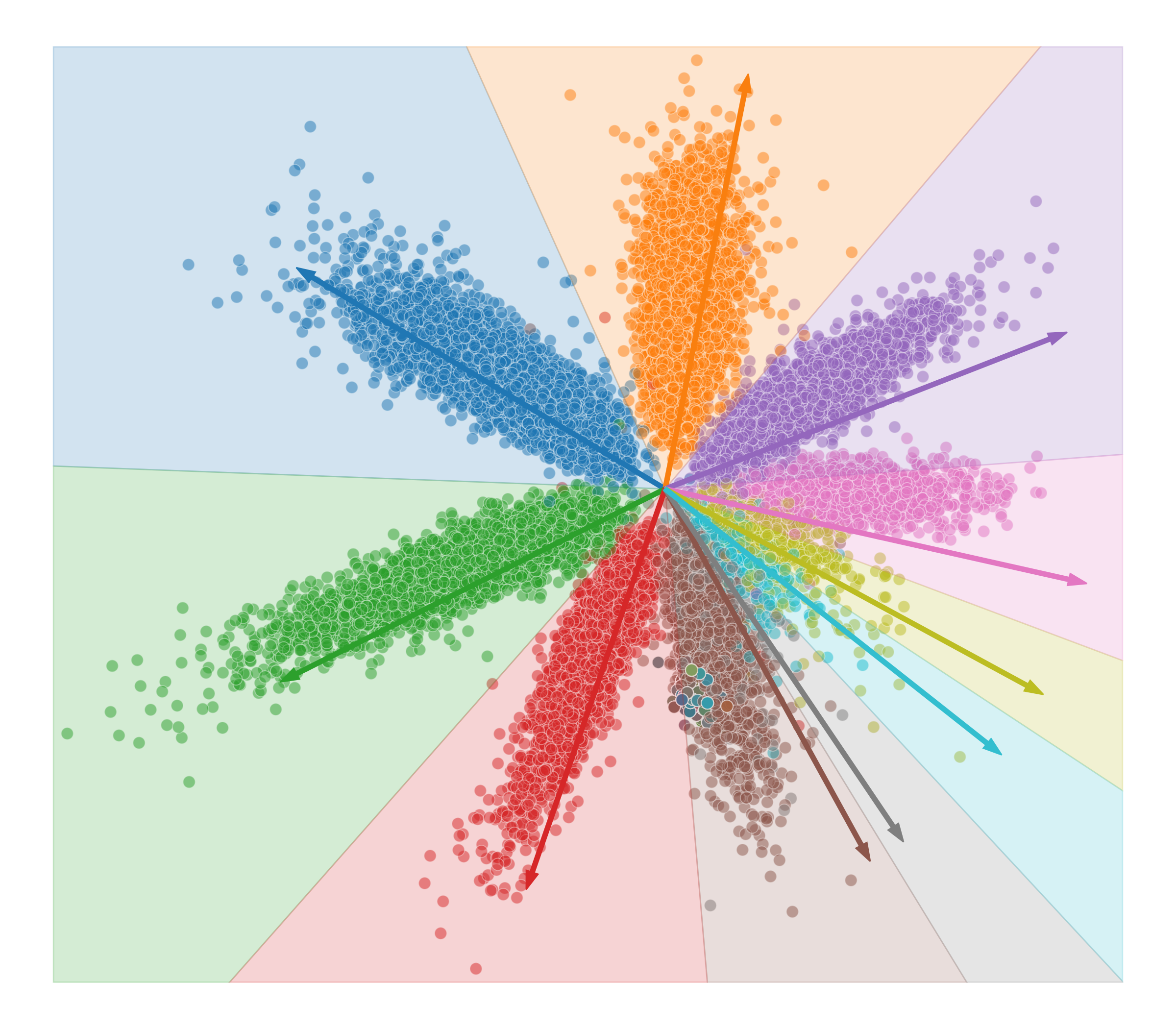}
  }
  \subfigure[]{
    \label{fig:imb-Norm-fix}
    \includegraphics[scale=0.12]{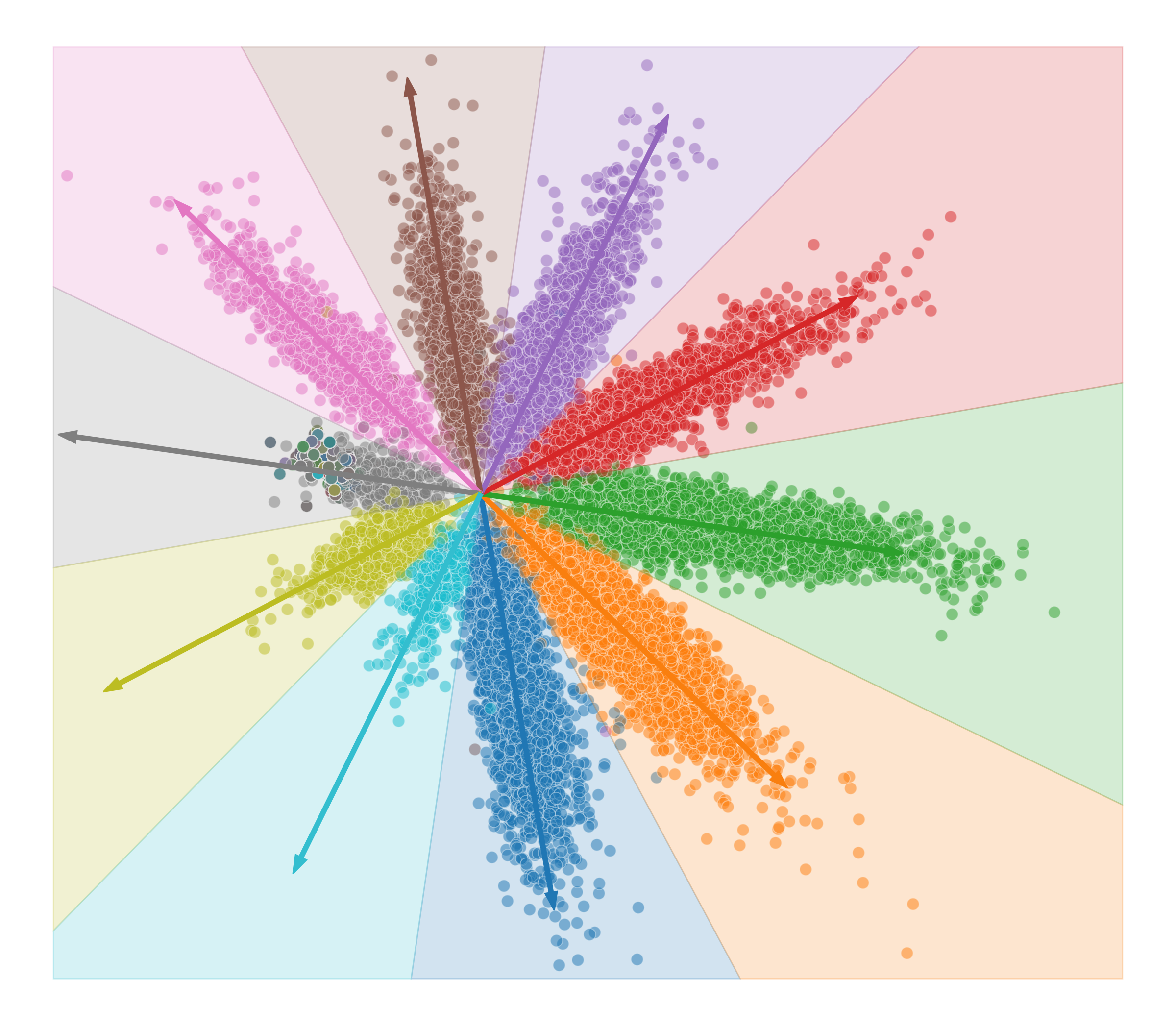}
  }
  \vskip-10pt
  \caption{Visualization on MNIST (a) and long-tailed MNIST (b-d) with imbalance ratio 0.1. The learned representations, class prototypes and decision regions are represented by points, vectors and areas in different colors, respectively. The corresponding color for each class is provided in (a), which is also used in (b-d). Specifically,  (a) denotes the class-balanced case by CE, where features and prototypes are optimized to be perfectly balanced. (b) denotes the class-imbalanced case by CE, where the majority classes (``0-3") occupy most of the feature space, the representations of minority classes (``7-9") are narrow, and the majority classes have larger norms and angular distance from other prototypes, and the reverse on the minority classes. (c) denotes the class-imbalanced case by CE with feature and prototype normalization, where minority classes are squeezed into a small space, while the majority classes take up a lot, but the learned representations and decision regions are more separated clearly than (b). (d) denotes the class-imbalanced case by CE with proposed feature normalized and prototype-anchored learning, the decision regions and feature distribution are more balanced.}
  \label{fig:mnist}
\end{figure*}

\section{Preliminaries and Motivation}

\subsection{Preliminaries}
For multi-class classification problem, let $\mathcal{X}\subseteq \mathbb{R}^m$ denote the feature space and $\mathcal{Y} = \{1,\cdots,k\}$ denote the label space, we are usually given a labeled dataset $D=\{(\bm{x}_i, y_i)\}_{i=1}^N$ to train models, where $(\bm{x}_i, y_i)$ are drawn from the joint distribution $\mathcal D$. 

\textbf{Softmax Loss.} The most popular loss used for classification is the softmax loss, for $(\bm{x}_i, y_i)$ which is formulated as:
\begin{equation}
    \label{softmax-loss}
    L_i=-\log \frac{\exp(\bm{w}_{y_i}^\top \bm{z}_i)}{\sum_{j=1}^k\exp(\bm{w}_j^\top \bm{z}_i)},
\end{equation}
where $\bm{z}_i=\bm{\phi}_{\Theta}(\bm{x}_i)\in\sR^d$ is the learned feature representation, $\bm{\phi}_\Theta$ denotes the feature extraction network with parameters $\Theta$, and $W=(\bm{w}_1,...,\bm{w}_k)\in\sR^{d\times k}$ represents the classifier implemented with a linear layer. For brevity, we use ``prototypes" to denote the weight vectors of classifier.

Intuitively, the softmax loss promotes the learned feature representation $\bm{z}_i$ to be close to the corresponding prototype $\bm{w}_{y_i}$ and apart away from other prototypes. However, it may lead to feature representations without enough discriminativeness by overly enlarging the norm $\|\bm{z}_i\|_2$ or $\|\bm{w_j}\|_2$, especially when imperfect annotation exists. For instance, when $\mathbb{P}(y)$ is highly skewed, \textit{i.e.}, imbalanced distribution, the majority classes occupy most of the feature space while the  minority classes are narrow, as illustrated in \cref{fig:imb-CE}. An approach to remedy this effect is performing normalization on both feature vectors and prototypes \cite{wang2017normface}, \textit{i.e.}, restricting them on the unit sphere $\sS^{d-1}$, which 
can improve discriminativeness to some extent (see \cref{fig:imb-Norm}).

\subsubsection{Margin-based Loss}
Based on the spherical constraint, a popular research line to encourage more discriminative features is to introduce an additional margin into softmax loss to draw a more strict decision boundary \cite{wang2018cosface, deng2019arcface, cao2019learning}. The margin-based loss is formulated as
\begin{equation}
    \label{margin-based-loss}
    L_{\bm{\alpha}}=-\log \frac{\exp(s\bm{w}^{\mathrm{T}}_{y}\bm{z}+\alpha_{y})}{\exp(s\bm{w}^{\mathrm{T}}_{y}\bm{z}+\alpha_{y})+\sum\limits_{j\neq y}\exp(s\bm{w}^{\mathrm{T}}_j\bm{z})},
\end{equation}
where both $\bm{w}_j$ and $\bm{z}\in\sS^{d-1}$, $s$ is the inversion of the temperature parameter, and $\alpha_{y}$ is the introduced per-class margin that depends on the distribution of $y$. 
For instance, in one well-known variant of margin-based loss---the label-distribution-aware margin (LDAM) loss \cite{cao2019learning}, the per-class margin is defined as $\alpha_{y}\propto \mathbb{P}(y)^{-1/4}$.


\subsubsection{Sample Margin}
\label{samle-margin-definition}
According to the definition in \cite{koltchinskii2002empirical, cao2019learning}, 
for the network ${f}(\bm {x}; \Theta,W)=W^{\top}\bm \phi_{\Theta}(\bm {x}):\sR^m\rightarrow \sR^k$ that outputs $k$ logits, the margin of a sample $(\bm{x}, y)$ is defined as
\begin{equation}
    \label{sample-margin}
    \gamma(\bm{x}, y)= {f}(\bm{x})_y-\max_{j\not=y} {f}(\bm{x})_j= \bm{w}_y^{\top}\bm{z} - \max_{j\not=y} \bm{w}_j^{\top} \bm{z},
\end{equation}
Let $S_j=\{i:y_i=j\}$ denote the sample indices corresponding to class $j$. We further define the sample margin for class $j$ as $\gamma_j = \min_{i\in S_j}\gamma(\bm{x}_i, y_i)$, and the minimal sample margin over the entire dataset is $\gamma_{\min}=\min\{\gamma_1,...,\gamma_k\}$.

Let $L_{\gamma, j}[f]=\sP_{\bm{x}\sim \mathcal P_j}[\max_{j'\not=j} f(\bm{x})_{j'}>f(\bm{x})_j-\gamma]$ denote the hard margin loss on samples from class $j$, and let $\hat{L}_{\gamma,j}$ denote its empirical variant. When the training dataset is separable (which means that there exists $f$ such that $\gamma_{\min}>0$ for all training samples), \citet{cao2019learning} provided a class-balanced generalization error bound: for $\gamma_j>0$ and all $f\in\mathcal F$, with a high probability, one have
\begin{equation}
    \label{generalization-bound}
    \begin{aligned}
    &\sP_{(\bm{x}, y)}[f(\bm x)_y<\max_{l\not=y}f(\bm x)_l]\\
    \le&\frac{1}{k}\sum_{j=1}^k\left(\hat{L}_{\gamma_j,j}[f]+\frac{4}{\gamma_j}\hat{\mathfrak{R}}_{j}(\mathcal F) + \varepsilon_{j}(\gamma_j)\right).
    \end{aligned}
\end{equation}
In the right-hand side, $\frac{4}{\gamma_j}\hat{\mathfrak{R}}_{j}(\mathcal F)$ that denotes the empirical Rademacher complexity has a big impact on the value of the generalization bound. \citet{cao2019learning} suggest that large margins $\{\gamma_j\}_{j=1}^k$ for all classes should be encouraged in order to tighten the generalization bound.

\subsection{Theoretical Motivation}
\label{sample-margin-sec}
Inspired by \citet{zhou2022learning}, we turn to \textit{tighten the generalization bound by maximizing the minimal sample margin $\gamma_{\min}$}. In this way, we can achieve the larger margins $\{\gamma_j\}$ for all classes. To this end, we have the following theorems:
\begin{lemma}[\textbf{The optimality condition of prototypes to maximize $\gamma_{\min}$}]
\label{largest-sample-margin}
 If $\bm{w}_1,...,\bm{w}_k, \bm{z}_1,...,\bm{z}_N\in\sS^{d-1}$ ($2\le k\le d+1$), then the maximum of the minimal sample margin $\gamma_{\min}$ is $\frac{k}{k-1}$, which is uniquely obtained if $\bm{z}_i=\bm{w}_{y_i}$, $\forall i$, and $\bm{w}_i^{\top}\bm{w}_j=\frac{-1}{k-1}$, $\forall i\neq j$.
\end{lemma}

This lemma indicates that, once the maximum of $\gamma_{\min}$ is obtained, the class prototypes $\bm{w}_1,...,\bm{w}_k$ will satisfy $\bm{w}_i^{\top}\bm{w}_j=\frac{-1}{k-1}$, $\forall i\neq j$. In other words, the angular between any two class prototypes will be $\arccos{\frac{-1}{k-1}}$, which is actually also the solution of the \textit{best-packing problem} on the sphere \cite{borodachov2019discrete}. This conclusion can also be derived for the loss in \cref{margin-based-loss} with the same per-label margins when learning from balanced datasets.

\begin{theorem}
\label{balanced-case-for-per-margin}
For balanced datasets (\textit{i.e.}, each class has the same number of samples), if $\bm{w}_1,...,\bm{w}_k, \bm{z}_1,...,\bm{z}_N\in\sS^{d-1}$,  ($2\le k\le d+1$), then learning with $L_{\bm{\alpha}}$ that has the same per-class margins (\textit{i.e.}, $\alpha_j=\alpha$, $\forall j\in[k]$) will deduce $\bm{z}_i=\bm{w}_{y_i}$, $\forall i$, and $\bm{w}_i^{\top}\bm{w}_j=\frac{-1}{k-1}$, $\forall i\neq j$.
\end{theorem}

This theorem shows that, on balanced datasets, by minimizing the margin-based loss with the same per-class margins, we can achieve the maximum of $\gamma_{\min}$, and thus make the distribution of prototypes the most discriminative, and also make each feature concentrate on the corresponding prototype. However, on imbalanced datasets, this wonderful effect cannot be enjoyed anymore. As shown in \cref{fig:imb-Norm}, the learned representation of minority classes are squeezed into a narrow space, while that of majority classes occupy a large part of the feature space.

In the class-imbalanced setting, LDAM \citep{cao2019learning} intuitively assigns different margins to each class to enforce a large margin between the minority and majority classes. However, there is no theoretical guarantee that we will obtain class-balanced features and prototypes such as those used in learning with balanced datasets. Moreover, different per-class margins in LDAM cannot guarantee Fisher consistency \cite{lin2004note} (or called classification calibration \cite{bartlett2006convexity}), since it allows shifting the decision boundary away from the minority classes. We can derive the following theorem:
\begin{theorem}
\label{LDAM-not-CC}
Under class-imbalanced data distribution, LDAM is not classification-calibrated.
\end{theorem}
As a corroboration, \citet{menon2020long} empirically claims that the existence of different per-class margins would sacrifice the consistency with the Bayes-optimal solution and thus result in sub-optimal solutions even in simple settings. 

\subsection{Prototype-anchored Learning}
As analyzed above, when learning with imperfect annotations, the current softmax loss and margin-based losses suffer from the lack of enough discrimination or Fisher consistency. Lemma \ref{largest-sample-margin} provides the optimality condition of prototypes to maximize the minimal sample margin. Inspired by the derived condition $\bm{w}_i^{\top}\bm{w}_j=\frac{-1}{k-1}$, $\forall i\neq j$, we propose a simple yet effective method, namely \textit{prototype-anchored learning} (PAL).
Specifically, we derive a classifier composed of predefined prototypes satisfying the optimality condition in Lemma \ref{largest-sample-margin}, and the classifier is anchored during training, \textit{i.e.}, performs no gradient updates. In practice, to generate these prototypes we randomly initialize $\{\bm{z}_i\}_{i=1}^N$ and $\{\bm{w}_i\}_{i=1}^k$ in $L_{\bm{\alpha}}$ ($\bm{\alpha}=0$) with a balanced setting (\textit{i.e.}, $N=k$ and $y_i=i$), and then directly minimize $L_{\bm{\alpha}}$ to obtain the optimal prototypes $\{\bm{w}_i\}_{i=1}^k$ according to Theorem \ref{balanced-case-for-per-margin}. The pseudo code can be seen in \cref{pseudocode}.

Prototype-anchored learning can be regarded as a parameterized loss with respect to the feature representations. As shown in \cref{softmax-loss}, the softmax loss parameterized by prototypes $(\bm{w}_1,...,\bm{w}_k)$ enlarges the inner product $\bm{w}_{y_i}^{\top}\bm{z}_i$ and shrinks other inner products $\bm{w}_j^{\top}\bm{z_i}$, $\forall j\neq y_i$. When these prototypes are anchored, the classification problem is transformed to a feature alignment problem, which is intuitively more stable than jointly learning feature extractor and classifier prototypes, especially for learning with imperfect data.

\section{Applications of PAL}

Due to its simplicity, PAL can be easily incorporated into various learning-based classification schemes. In this section, we present how to apply PAL in handling imperfect annotations. We specially consider two scenarios---class-imbalanced learning and learning with noisy labels. 

\begin{figure*}[!t]
\centering
    \subfigure[CE with wd=0.0]{
        \label{fig:ECE-CE-0.0-m}
        \includegraphics[scale=0.23]{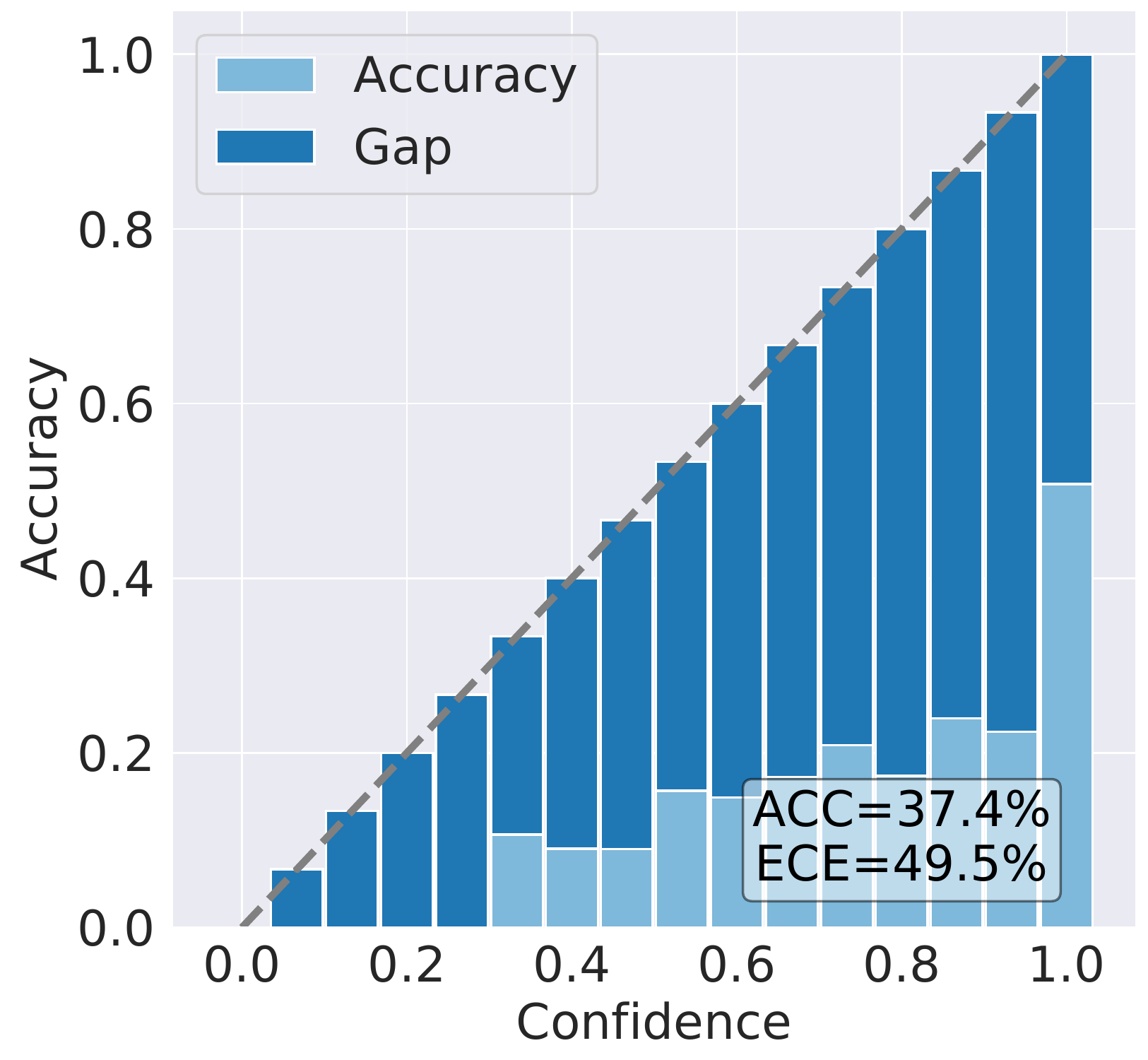}
    }
    \subfigure[CE with wd=5e-5]{
        \label{fig:ECE-CE-5e-5-m}
        \includegraphics[scale=0.23]{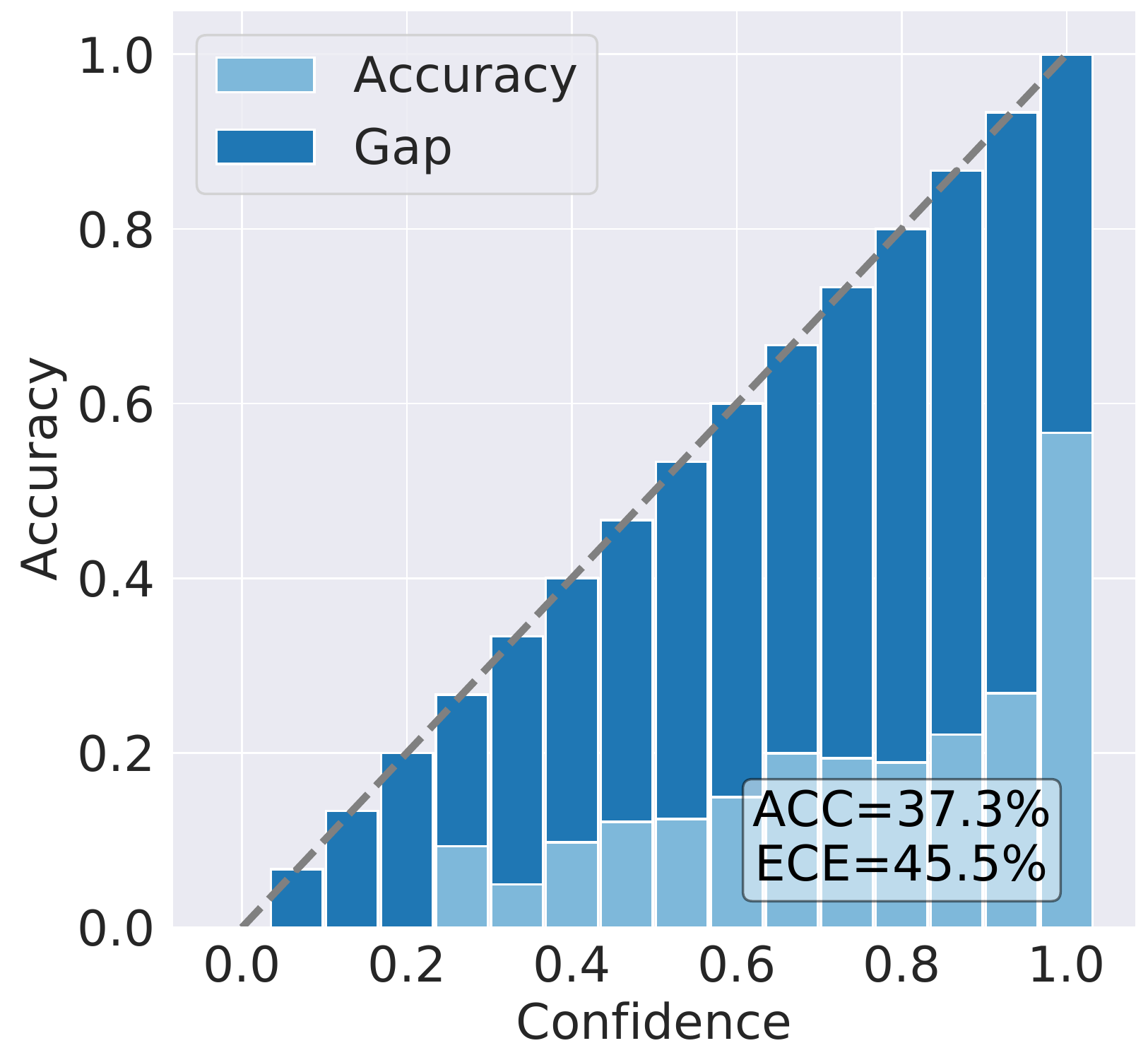}
    }
    \subfigure[CE with wd=5e-4]{
        \label{fig:ECE-CE-5e-4-m}
        \includegraphics[scale=0.23]{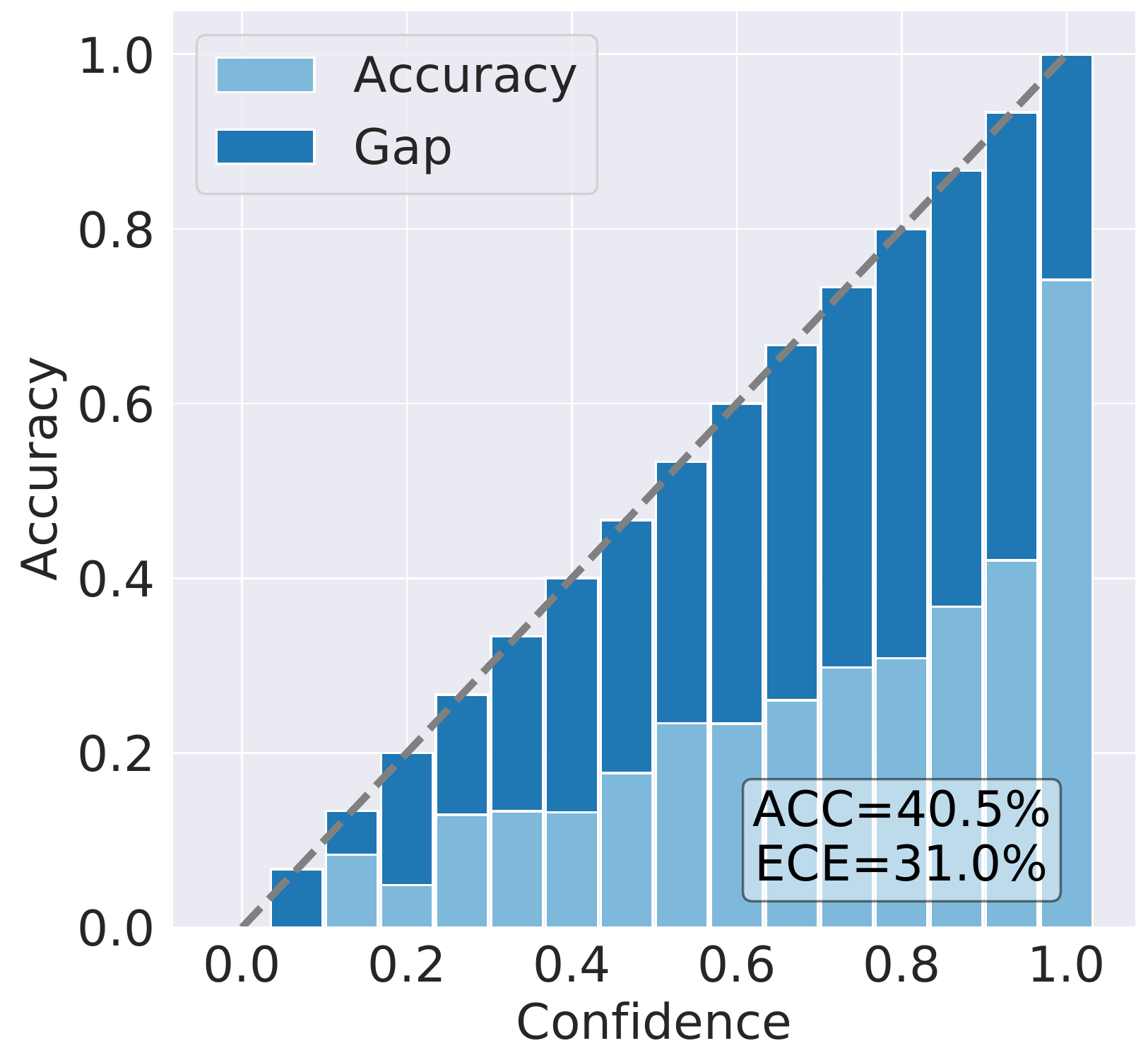}
    }
    \subfigure[CE with wd=1e-3]{
        \label{fig:ECE-CE-1e-3-m}
        \includegraphics[scale=0.23]{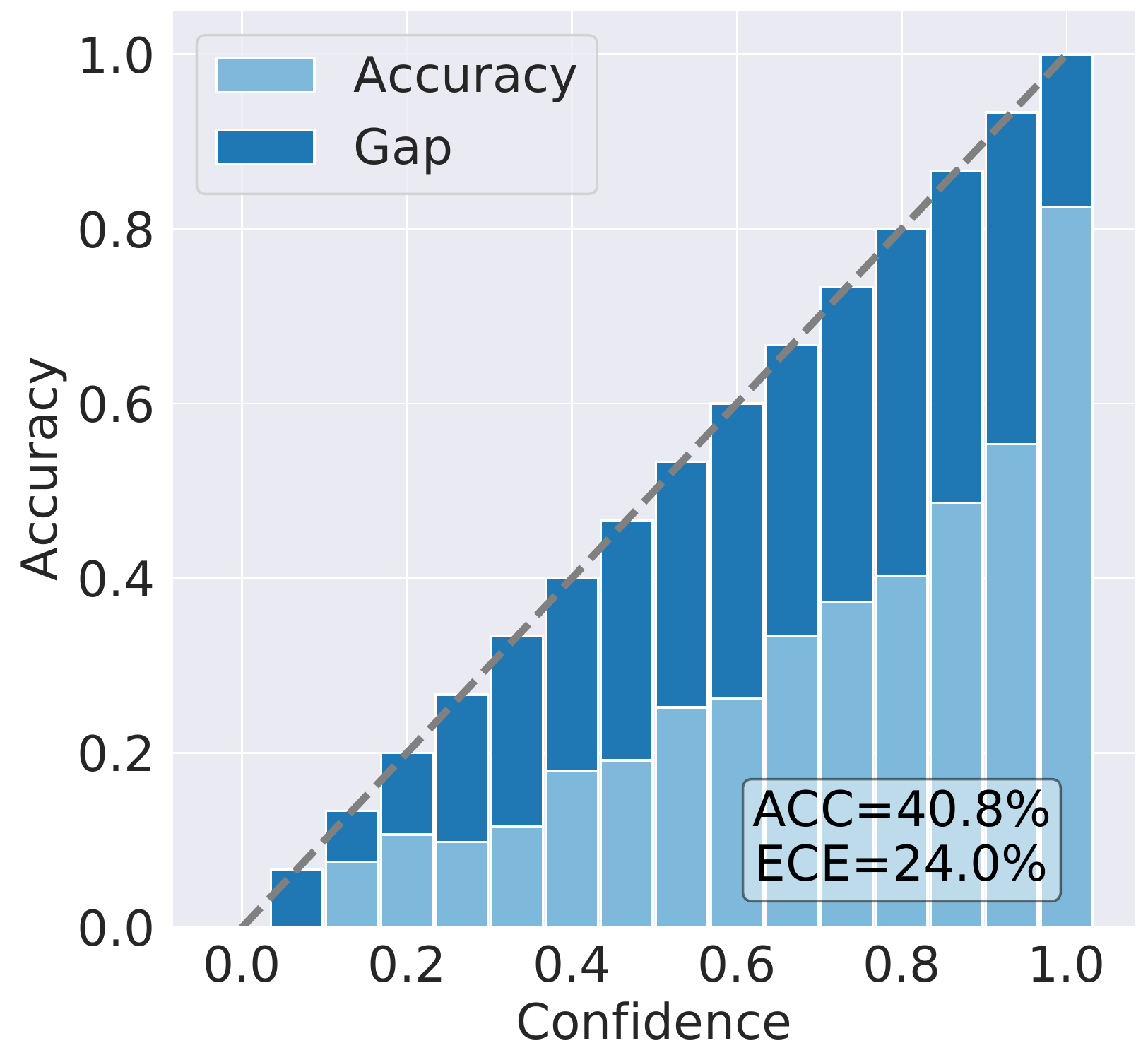}
    }
  \caption{Reliability diagrams of ResNet-32 trained by CE on CIFAR-100-LT with imbalance ratio 100 under different weight decays (wds).  As can be seen, an appropriate larger weight decay can improve both accuracy and confidence.}
  \label{fig:ece-ce}
\end{figure*}

\begin{figure}[tpb]
  \centering
  \subfigure[prototype norms]{
    \label{fig:CE-weights}
    \includegraphics[scale=0.18]{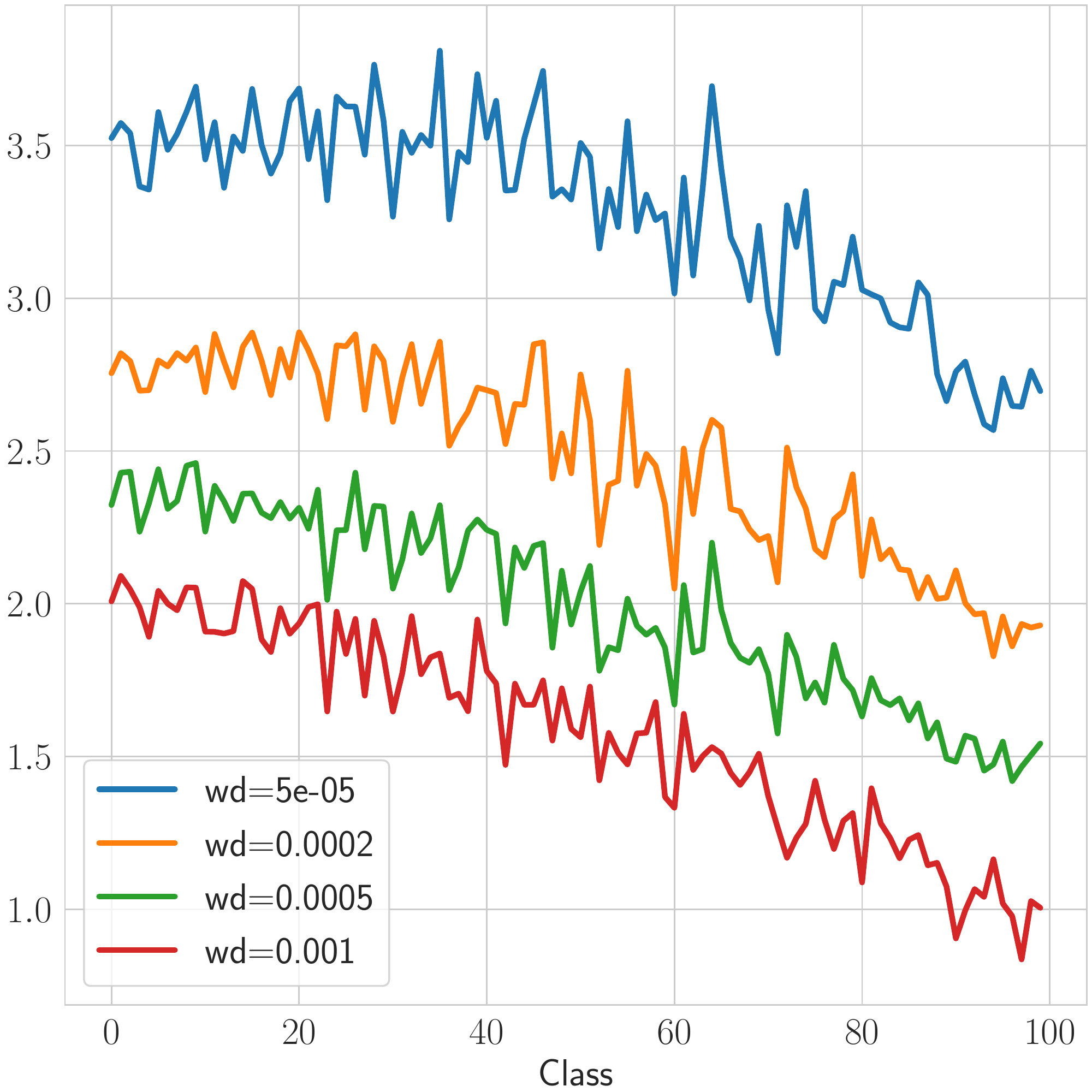}
  }
  \subfigure[histogram of feature norms]{
    \label{fig:CE-vhist}
    \includegraphics[scale=0.18]{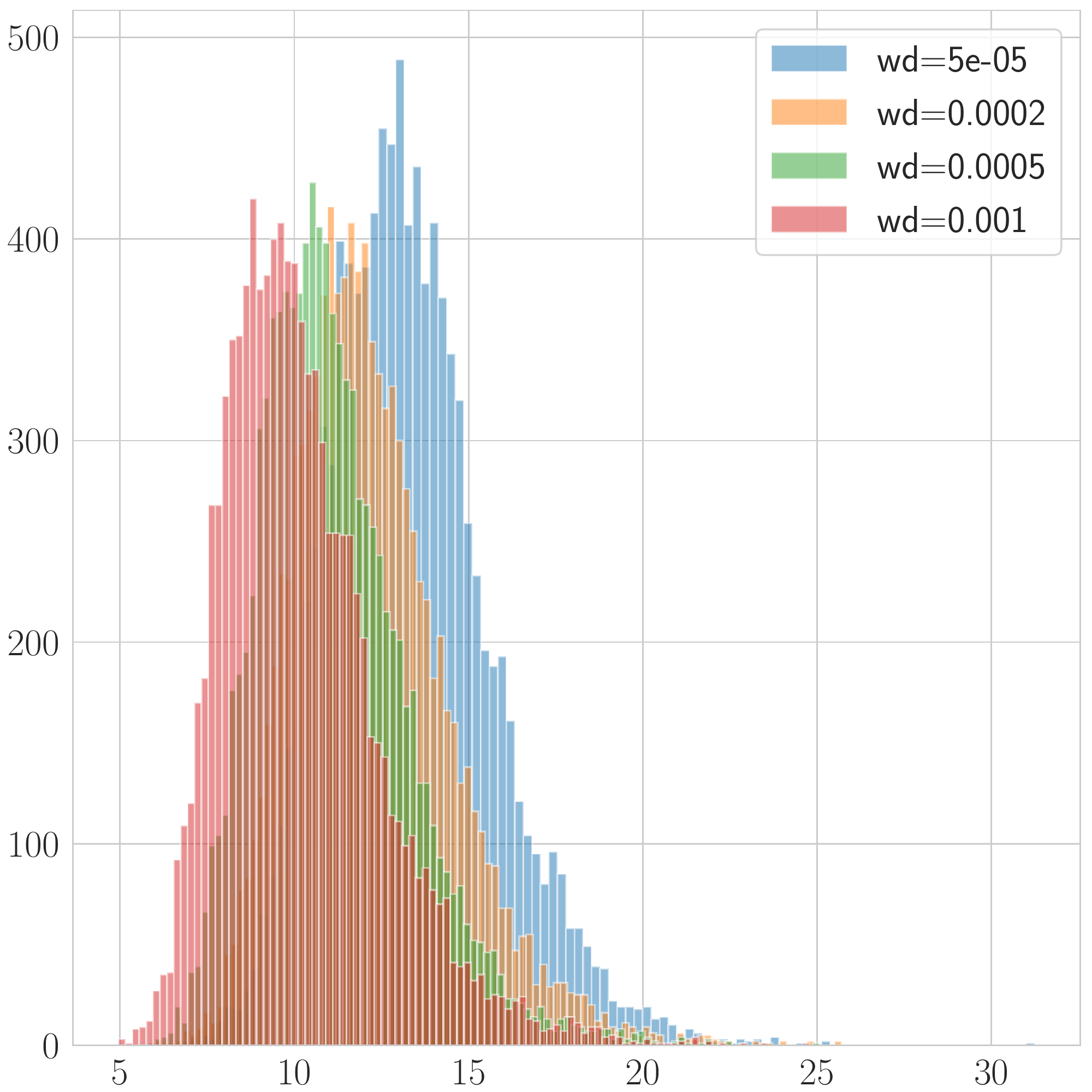}
  }
  \vskip-10pt
  \caption{Illustration of prototypes norms and feature norms by CE trained on CIFAR-100-LT with imbalance ratio 100 under different weight decays. As can be seen, the larger weight decay usually leads to smaller prototype norms and feature norms.}
  \label{fig:wd}
\end{figure}

\subsection{Class-imbalanced Learning}
Theorem \ref{balanced-case-for-per-margin} proves that learning with $L_{\bf{\alpha}}$ will lead to prototypes that satisfy $\bm{w}_i^{\top}\bm{w}_j=\frac{-1}{k-1}$, $\forall i\neq j$, which however works for balanced datasets. For imbalanced datasets, a phenomenon termed \textit{Minority Collapse} would occur \cite{fang2021exploring}. In the following, we demonstrate in theory that the anchored prototypes can alleviate minority collapse in imbalanced training and acquire the maximum of $\gamma_{\min}$:


\begin{theorem}
\label{largest-margin-imbalanced-case}
For imbalanced or balanced datasets, if $\bm{z}_1,...,\bm{z}_N, \bm{w}_1,...,\bm{w}_k\in\sS^{d-1}$  ($2\le k\le d+1$), where $\bm{w}_1,...,\bm{w}_k$ are anchored and satisfy that $\bm{w}_i^{\top}\bm{w}_j=\frac{-1}{k-1}$, $\forall i\neq j$, then learning with $L_{\bm{\alpha}}$ will deduce $\bm{z}_i=\bm{w}_{y_i}$, $\forall i$, and obtain the maximum of the minimal sample margin $\gamma_{\min}=\frac{k}{k-1}$.
\end{theorem}
\vskip-5pt

Theorem \ref{largest-margin-imbalanced-case} offers a solid theoretical guarantee to obtain the maximal $\gamma_{\min}$ on imbalanced datasets by PAL.

In class-imbalanced learning, a fruitful avenue of exploration is the strategy of decoupling the learning procedure into representation learning and classification training \cite{kang2019decoupling}, which works in a two-stage manner: firstly learn feature representation from the original imbalanced data, and then retrain the classifier using class-balanced sampling with the first-stage representation extractor frozen. The state-of-the-art decoupling methods usually utilize the softmax loss for the representation learning \cite{kang2019decoupling, zhong2021improving}, which does not explicitly perform normalization on features. This seems to invalidate the normalization condition of Theorem 3.1. Fortunately, in deep neural networks, there are some implicit biases that actually conduct restriction on features and prototypes, such as weight decay \cite{fang2021exploring}. \Cref{fig:wd} shows that weight decay can facilitate not only explicitly learning small weights of networks but also implicitly producing small feature representations. 
Moreover, we empirically find that \textit{an appropriately large weight decay can mitigate class imbalance} by preventing the excessively large norms that cause over-confidence (see more analysis in \cref{role-of-weight-decay-in-imbalanced-learning}) in \cref{fig:ece-ce}. This makes our PAL still work well for the softmax loss in imbalanced setting with limited feature norms. Formally, we have the following theorem:
\begin{theorem}
\label{largest-margin-ce-imbalanced-case}
For imbalanced or balanced datasets, if $\|\bm{z}_i\|\le B$, $\forall i\in[1,n]$, and the class prototypes $\bm{w}_1,...,\bm{w}_k \in \mathbb{S}^{d-1}$  ($2\le k\le d+1$) are anchored to satisfy $\bm{w}_i^{\top}\bm{w}_j=\frac{-1}{k-1}$, $\forall i\neq j$, then learning with the softmax loss will deduce $\frac{\bm{z}_i^{\top}\bm{w}_{y_i}}{\|\bm{z}_i\|_2}=1$, $\forall i$, and obtain the maximum of $\gamma_{\min}$.
\end{theorem}

According to this theoretical guarantee, PAL can be used in the feature representation learning of the first stage, as well as in tandem with other performance-boosting approaches and modules. Through such a simple operation, as demonstrated by extensive experiments in \Cref{imbalanced-exp}, PAL can significantly boost the performance of the existing long-tailed classification methods.


\subsection{Learning with Noisy Labels}

In noise-tolerant learning, the most popular family of loss functions is the symmetric loss \cite{Manwani, unhingedloss, symmetric}, which satisfies
\begin{equation}
    \sum\nolimits_{i=1}^k L(f(\bm{x}), i)=C, \forall  \bm{x}\in \mathcal X, \forall f.
\end{equation}
where $C$ is a constant. This symmetric condition theoretically guarantees the noise tolerance by risk minimization on a symmetric loss function \cite{symmetric} under symmetric label noise, \textit{i.e.}, the global minimizer of the noisy $L$-risk $R_L^\eta(f)=\mathbb{E}_{\mathcal D}[(1-\eta_{\bm{x}})L(f(\bm{x}), y)+\sum_{i\neq y}\eta_{\bm{x},i}L(f(\bm{x}), i)]$ also minimizes the $L$-risk $R_L(f)=\mathbb{E}_{\mathcal D}L(f(\bm{x}), y)$, where $\eta_{\bm{x,i}}$ denotes the probability (or called noise rate) of flipping label $y$ into label $i$.

\noindent\textbf{Negative-signed Sample Logit Loss.} 
In this work, we propose a novel loss for noise-tolerant learning based on sample logit $f(\bm{x})=W^\top \bm{\phi}_{\Theta}(\bm{x})$, called \textit{negative-signed sample logit loss} (NSL), which is defined as:
\begin{equation}
    L_{\mathrm{NSL}}(f(\bm{x}), i)=-f(\bm x)_i = -\bm{w}_i^\top \bm{\phi_{\Theta}(\bm{x})}.
\end{equation}
Importantly, when $\sum_{i=1}^k \bm{w}_i=0$, we can simply derive that $\sum_{i=1}^k L_{\mathrm{NSL}}(f(\bm{x}), i)=0$. Thus, NSL also serves as a symmetric loss.
However, NSL only contains the fitting term that encourages each feature to be close to the corresponding prototype, which would lead to a trivial solution when jointly training $\Theta$ and $W$. Fortunately, when prototypes are anchored, this issue can be easily handled as the solution of the best packing problem (which implicitly satisfies $\sum_{i=1}^k \bm{w}_i=0$).  Specifically, we have:
\begin{proposition}
\label{NSL-learning-with-noisy-labels}
If the prototypes $\bm{w}_1,...,\bm{w}_k\in\sS^{d-1}$ ($2\le k\le d+1$) are anchored to satisfy $\bm{w}_i^{\top}\bm{w}_j=\frac{-1}{k-1}$, $\forall i\neq j$,  $L_{\mathrm{NSL}}(f(\bm{x}), i)$ is symmetric. More specifically, we have $\sum_{i=1}^k L_{\mathrm{NSL}}(f(\bm{x}),i)=0$, and learning with $L_{\mathrm{NSL}}$ will lead to the maximum of $\gamma_{\min}$ under symmetric label noise.
\end{proposition}

More generally, given the anchored prototypes $\bm{w}_1,...,\bm{w}_k$,  we can provide a risk bound for the loss functions satisfying that $L_W(\bm{z})=\sum_{i=1}^k L(W^\top\bm{z},i)$ is $\lambda$-Lipschitz\footnote{that is, $\|L_W(\bm{z}_1)-L_W(\bm{z}_2)\|_2\le \lambda \| \bm{z}_1-\bm{z}_2\|_2$}:
\begin{theorem}
\label{lipschitz-symmetric}
In a multi-class classification problem, given $\bm{w}_1,...,\bm{w}_k$, if $\bm{z}=\bm{\phi}_{\Theta}(x)$ is norm-bounded by $B$, \textit{i.e.}, $\|\bm{z}\|_2\le B$, then for any loss $L(\bm{z},i)$ satisfying $L_W(\bm{z})=\sum_{i=1}^k L(W^\top\bm{z},i)$ is $\lambda$-Lipschitz, we have the following risk bound under symmetric label noise with $\eta < \frac{k-1}{k}$:
\begin{equation}
    \label{risk-bound}
    R_L(\hat{f})-R_L(f^*)\le \frac{2\eta \lambda B}{(1-\eta)k-1},
\end{equation}
where $\hat{f}$ and $f^*$ denote a global minimize of $R_L^\eta(f)$ and $R_L(f)$, respectively.
\end{theorem}
\cref{lipschitz-symmetric} naturally encompasses the symmetric condition in \citep[Theorem 1]{symmetric}. We know that if the loss $L$ is symmetric, then $\lambda=0$, and thus we have $R_L(\hat{f})=R_L(f^*)$. 
\begin{figure}[tbp]
        \centering
        \subfigure[ $\eta=0.6$]{
        \label{fig:CE-cifar10_s0.6}
        \includegraphics[scale=0.18]{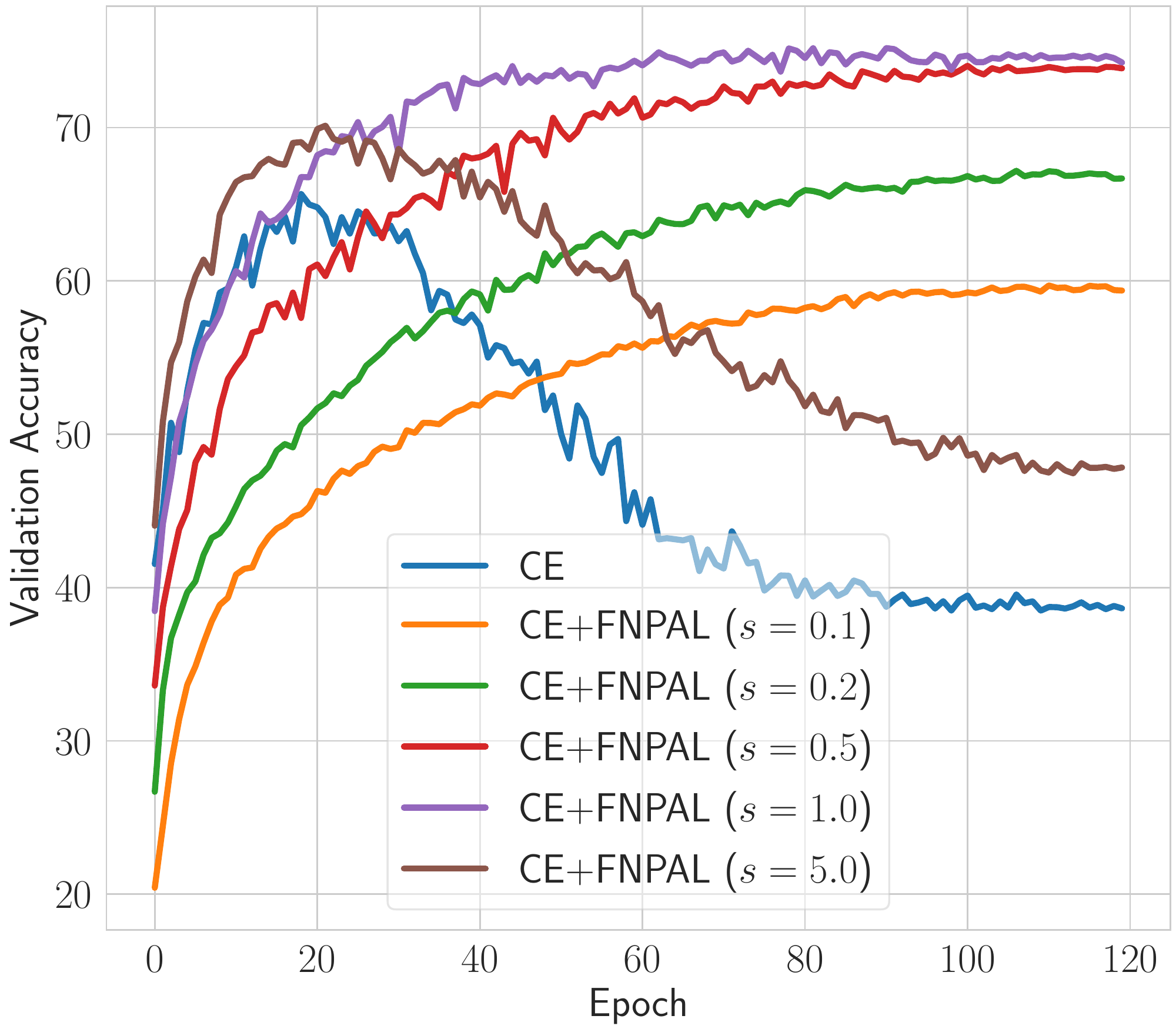}
          }
      \subfigure[$\eta=0.8$]{
        \label{fig:CE-cifar10_s0.8}
        \includegraphics[scale=0.18]{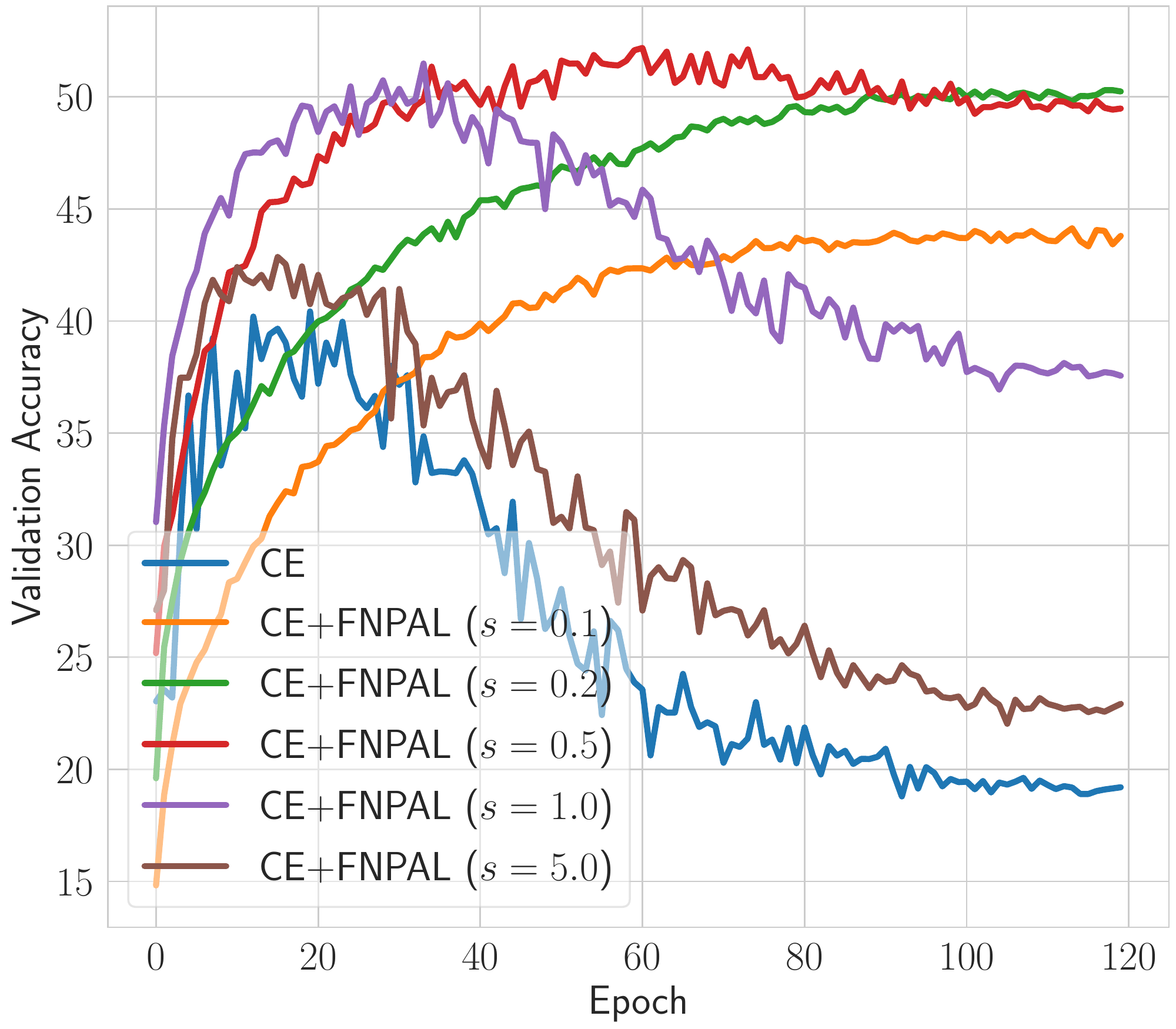}
      }
      \vskip-8pt
      \caption{Validation accuracies of CE and CE+FNPAL (with different scaled parameter $s$) on CIFAR-10 with symmetric label noise. As can be seen, CE exhibits significant overfitting after epoch 20 while CE+FNPAL shows more robustness and achieve better performance. To conclude, as $s$ decreases, the robustness increases. This empirically demonstrates that small feature norms can significantly mitigate label noise when prototypes are anchored. On the other hand, the fitting ability decreases as $s$ increases. That is, $s$ introduces a trade-off between fitting ability and robustness. Thus, an appropriate small $s$ is preferred for better performance.}
      \label{CE+FNPAL}
\end{figure}

{For general losses, when $B$ is fixed, the risk bound will depend on the Lipschitz constant of $L_W(\bm{z})$ and noise rate $\eta$. As can be seen, the Lipschitz constant depends on the choice of $W$. Actually, our PAL strategy in which the prototypes are anchored can provide a smaller $\lambda$ than learning with unanchored prototypes, especially for some losses like CE that are usually not Lipschitz continuous. This will be described later.} 
We also note that the feature norm $B$ can be naturally adjusted by tuning the scaled parameter $s$ under the spherical constraint on features and prototypes (\textit{e.g.}, $L_{\bm{\alpha}}$ in \cref{margin-based-loss}), which can be served as a parameter to trade off fitting ability and robustness.


We further validate the new trade-off by adjusting the inversion of temperature parameter. As illustrated in \cref{CE+FNPAL}, an appropriate small feature norm can achieve good robustness and accuracy. However, it can be observed that the same parameter $s$ at different noise rates does not perform exactly the same. For example, when $s=0.2$, it obtains the best accuracy at $\eta=0.8$ but encounters underfitting at $\eta=0.6$. This is mainly due to that smaller noise rates require more fitting power. We propose to set $s=O(\frac{1}{\eta+\epsilon})$, which makes the risk bound in \cref{risk-bound} be noise-independent.

Based on the above discussion, finally we suggest the \textit{feature normalized and prototype-anchored learning} (FNPAL) strategy that performs $\ell_2$ normalization on features and anchored prototypes, which can be combined with traditional losses to boost their ability on noise-tolerant learning. Specifically, for cross-entropy loss, we have the following proposition:

\begin{proposition}
\label{risk-bound-CE}
In a multi-class classification problem, let $\bm{w}_1,...,\bm{w}_k\in\sS^{d-1}$  ($2\le k\le d+1$) satisfy $\bm{w}_i^{\top}\bm{w}_j=\frac{-1}{k-1}$, $\forall i\neq j$, if $\bm{z}=\bm{\phi}_{\Theta}(x)$ is norm-bounded by $B$, \textit{i.e.}, $\|\bm{z}\|_2\le B$, then we have the following risk bound for the CE loss defined in \cref{softmax-loss} under symmetric label noise with $\eta < \frac{k-1}{k}$:
\begin{equation}
    R_L(\hat{f})-R_L(f^*)\le \frac{2c\eta k(1-t)B}{k-1+t(k-1)^2},
\end{equation}
where $c=\frac{k-1}{(1-\eta)k-1}$, $t=\exp(-\frac{kB}{k-1})$, $\hat{f}$ and $f^*$ denote the global minimum of $R_L^\eta(f)$ and $R_L(f)$, respectively.
\end{proposition}
\textbf{Remark.} Proposition \ref{risk-bound-CE} provides a FNPAL-based risk bound that relies on the Lipschitz constant $\lambda_{\text{PAL}}=\frac{k(1-\exp(-\frac{kB}{k-1}))}{1+(k-1)\exp(-\frac{kB}{k-1})}$. Actually, FNPAL indicates a tighter risk bound by a smaller Lipschitz constant than unnormalized or unanchored cases:
(1) when $\bm{w}$ is not normalized, we can easily know that $L_W(\bm{z})$ is not Lipschitz continuous as $\bm{w}_i=t\bm{z}$ and $t\rightarrow \infty$; (2) when $\bm{w}_i$ is normalized and unanchored (\textit{w.l.o.g}, $\|\bm{w}_i\|_2=1$) and $\bm{z}$ is unnormalized, we have the Lipschitz constant $\sup_{W,\bm{z}}\|\frac{\partial L_W(\bm{z})}{\partial \bm{z}}\|_2\ge k >\lambda_{\text{PAL}}$;
(3) when $\bm{w}_i$ and $\bm{z}$ are normalized and unanchored ($\|\bm{w}_i\|_2=1$, $\|\bm{z}\|_2=B$), we have $\sup_{W,\bm{z}}\|\frac{\partial L_W(\bm{z})}{\partial \bm{z}}\|_2\ge \frac{2(\exp(2B)-1)}{\frac{\exp(2B)}{k-1}+1} > \lambda_{\text{PAL}}$. More details are provided in \cref{a-tighter-bound-indicated-by-PAL}. 

Similar risk bounds can also be derived for GCE \cite{gce} and Focal loss \cite{lin2017focal}. Please see \cref{risk-bound-ce-gce-focal} for more details.

\begin{table}[htbp]
    \small
    \centering
    \vskip-10pt
    \caption{Validation accuracy (\%) on ImageNet-LT. The results with positive gains are \textbf{boldfaced} and the best one is \underline{underlined}.}
    \label{tab:lt-imagenet}
    \begin{tabular}{c|ccc|c}
        \toprule
         Method &  Many & Medium & Few & All\\
         \midrule
         CE & 66.8 & 36.9 & 7.1 & 43.6\\
         FL & 64.3 & 37.1 & 8.2 & 43.7\\
         OLTR & 51.0 & 40.8 & 20.8 & 41.9\\
         Causal Norm & 65.2 & 47.7 & 29.8 & 52.0\\
         Balanced Softmax  & 63.6 & 48.4 & 32.9 & 52.1\\
         LADE & 65.1 & 48.9 & 33.4 & 53.0\\
        cRT+mixup & 63.9 & 49.1 & 30.2 & 51.7\\
        LWS+mixup & 62.9 & 49.8 & 31.6 & 52.0\\
        MiSLAS & 61.7 & 51.3 & 35.8 & 52.7 \\
        \midrule
        \textbf{CE+PAL}  & \textbf{69.0} & \textbf{42.5} & \textbf{11.0} & \textbf{47.6}\\
        \textbf{MiSLAS+PAL} & \textbf{64.0} & \textbf{51.6} & 32.4 & \underline{\textbf{53.3}}\\
         \bottomrule
    \end{tabular}
    \vskip-10pt
\end{table}
\section{Experiments}
In this section, we empirically investigate the effectiveness of prototype-anchored learning on two tasks: long-tailed classification and learning with noisy labels. The benchmark datasets include benchmark datasets: MNIST \cite{MNIST}, CIFAR-10/-100 \cite{CIFAR} and ImageNet \cite{deng2009imagenet} as well as a real-world long-tailed and noisy dataset WebVision \cite{li2017webvision}.

\begin{table}[!t]
\small
\centering
\caption{Validation accuracy (\%) on CIFAR-100-LT. The results with positive gains are \textbf{boldfaced} and the best one is \underline{underlined}.}
\label{tab:lt-cifar100}
\begin{tabular}{c|ccc}
    \toprule
     \multirow{3}{*}{Method} &  \multicolumn{3}{c}{CIFAR-100-LT}\\
     \cmidrule{2-4}
      & \multicolumn{3}{c}{Imbalance Ratio}\\
      &  100 & 50 & 10\\
      \midrule
      CE & 38.4 & 43.9 & 55.8\\
      FL & 38.4 & 44.3 & 55.8\\
      CE+mixup & 39.6 & 45.0 & 58.2\\
      LDAM & 41.8 & 45.6 & 59.0\\
      
      BBN & 42.6 & 47.1 & 59.2\\
      cRT+mixup & 45.1 & 50.9 & 62.1\\
      LWS+mixup & 44.2 & 50.7 & 62.3\\
      PCSoftmax & 45.3 & 49.5 & 61.2\\
      LADE & 45.4 & 49.5 & 59.0 \\
      LDAM+DRW & 45.2 & 50.0 & 60.8\\
      MiSLAS & 47.0 & 52.3 & 63.2\\
      TADE & 49.8 & 53.9 & 63.6\\
      \midrule
      \textbf{CE+PAL} & \textbf{39.2} & \textbf{44.4} & \textbf{57.3}\\
      \textbf{CE+mixup+PAL} & \textbf{41.9} & \textbf{47.2} & \textbf{60.2}\\
      \textbf{LDAM+PAL} & \textbf{42.7} & \textbf{46.9} & \textbf{59.4}\\
      \textbf{LDAM+DRW+PAL} & \textbf{45.9} & \textbf{50.1} & \textbf{61.4}\\
      \textbf{MiSLAS+PAL} & \underline{\textbf{50.6}} & \underline{\textbf{55.3}} & \underline{\textbf{64.2}}\\
     \bottomrule
\end{tabular}
\end{table}

\subsection{Evaluation on Long-tailed Classification}
\label{imbalanced-exp}
We evaluate our approach on benchmarks CIFAR-100-LT and ImageNet-LT with artificially created long-tailed settings. We follow the controllable data imbalance strategy in \cite{cao2019learning, zhong2021improving} by reducing the number of training examples per class and keeping the validation set unchanged. The imbalance ratio $\rho=\max_i n_i/\min_i n_i$ denotes the ratio between sample sizes of the most frequent and the least frequent classes.

\noindent\textbf{Baselines.} We compare the proposed method against previous one-stage methods of Focal loss (FL) \cite{lin2017focal}, LDAM \cite{cao2019learning}, OLTR \cite{liu2019large}, and LADE \cite{hong2021disentangling}, against two-stage methods including LDAM+DRW \cite{cao2019learning}, cRT \cite{kang2019decoupling}, BNN \cite{zhou2020bbn}, and MiSLAS \cite{zhong2021improving}, and an ensemble method TADE \cite{zhang2021test}. We simply embed PAL into one-stage methods CE and LDAM, and a two-stage method MiSLAS. For MiSLAS, we only use PAL in the representation learning stage.

\begin{table*}[!t]
\small
\centering
\caption{Validation accuracies (\%) of different methods on benchmark datasets with clean or symmetric label noise ($\eta\in[0.2, 0.4, 0.6, 0.8]$). The results (mean $\pm$ std) are reported over 3 random runs. The results with positive gains are \textbf{boldfaced} and the best one is \underline{underlined}.}
\label{symmetric-noise}
\begin{tabular}{c|c|c|cccc}
    \toprule
     \multirow{2}*{Dataset} & \multirow{2}*{Method} & \multirow{2}*{Clean ($\eta=0.0$)} & \multicolumn{4}{c}{Symmetric Noise Rate ($\eta$)}  \\
     ~ & ~ & ~ & 0.2 & 0.4 & 0.6 & 0.8\\
     \midrule
     \multirow{10}*{MNIST}
      & CE & 99.17 $\pm$ 0.04 & 91.40 $\pm$ 0.11 & 74.36 $\pm$ 0.29 & 49.32 $\pm$ 0.70 & 22.32 $\pm$ 0.15\\
    ~ & FL & 99.16 $\pm$ 0.02 & 91.49 $\pm$ 0.20 & 75.28 $\pm$ 0.10 & 50.25 $\pm$ 0.70 & 22.68 $\pm$ 0.14 \\
    ~ & GCE & 99.15 $\pm$ 0.02 & 98.90 $\pm$ 0.03 & 96.81 $\pm$ 0.23 & 81.39 $\pm$ 0.64 & 33.07 $\pm$ 0.31\\
    ~ & SCE & 99.28 $\pm$ 0.07 & 98.91 $\pm$ 0.12 & 97.60 $\pm$ 0.22 & 88.00 $\pm$ 0.50 & 47.32 $\pm$ 0.99\\
    ~ & NCE+MAE & 99.42 $\pm$ 0.02 & 99.18 $\pm$ 0.08 & 98.47 $\pm$ 0.21 & 95.52 $\pm$ 0.04 & 73.05 $\pm$ 0.59 \\
    ~ & NCE+RCE & 99.40 $\pm$ 0.04 & \underline{99.24} $\pm$ 0.01 & 98.44 $\pm$ 0.11 & 95.77 $\pm$ 0.09 & 74.80 $\pm$ 0.28\\
    ~ & NFL+RCE & 99.37 $\pm$ 0.01 & 99.16 $\pm$ 0.03 & 98.55 $\pm$ 0.05 & 95.62 $\pm$ 0.24 & 74.67 $\pm$ 0.97\\
    \cmidrule{2-7}
    ~ & NSL & 99.24 $\pm$ 0.03 & 98.99 $\pm$ 0.03 & 98.58 $\pm$ 0.11 & 95.99 $\pm$ 0.24 & 59.77 $\pm$ 1.98\\
    ~ & \textbf{CE+FNPAL} & 99.24 $\pm$ 0.05 & \textbf{99.05 $\pm$ 0.04} & \textbf{98.66 $\pm$ 0.04} & \textbf{97.62 $\pm$ 0.15} & \textbf{79.23 $\pm$ 0.87}\\
    ~ & \textbf{SCE+FNPAL} & 99.27 $\pm$ 0.04 & \textbf{99.06 $\pm$ 0.05} & \underline{\textbf{98.76 $\pm$ 0.09}} & \underline{\textbf{97.94 $\pm$ 0.07}} & \underline{\textbf{88.56 $\pm$ 1.07}}\\
    ~ & \textbf{NCE+RCE+FNPAL} & 99.29 $\pm$ 0.04 & 99.04 $\pm$ 0.07 & 98.11 $\pm$ 0.09 & 94.84 $\pm$ 0.08 & \textbf{79.70 $\pm$ 1.06}\\
    ~ & \textbf{NFL+RCE+FNPAL} & 99.29 $\pm$ 0.06 & 99.02 $\pm$ 0.05 & 98.32 $\pm$ 0.14 & 95.38 $\pm$ 0.11 & \textbf{76.06 $\pm$ 0.58}\\
    \midrule
    \multirow{10}*{CIFAR10} 
      & CE & 90.36 $\pm$ 0.25 & 74.78 $\pm$ 0.68 & 57.95 $\pm$ 0.12 & 38.21 $\pm$ 0.12 & 18.89 $\pm$ 0.43\\
    ~ & FL & 89.69 $\pm$ 0.25 & 74.19 $\pm$ 0.23 & 57.35 $\pm$ 0.27 & 38.11 $\pm$ 0.76 & 19.39 $\pm$ 0.44\\
    ~ & GCE & 89.37 $\pm$ 0.29 & 87.05 $\pm$ 0.21 & 82.43 $\pm$ 0.10 & 68.05 $\pm$ 0.07 & 25.21 $\pm$ 0.28\\
    ~ & SCE & 91.24 $\pm$ 0.19 & 87.34 $\pm$ 0.01 & 79.84 $\pm$ 0.43 & 61.09 $\pm$ 0.19 & 27.19 $\pm$ 0.34\\
    ~ & NCE+MAE & 89.02 $\pm$ 0.09 & 87.06 $\pm$ 0.17 & 83.92 $\pm$ 0.16 & 76.47 $\pm$ 0.25 & 45.01 $\pm$ 0.31\\
    ~ & NCE+RCE & 91.12 $\pm$ 0.14 & 89.21 $\pm$ 0.00 & 86.03 $\pm$ 0.14 & 80.04 $\pm$ 0.26 & 51.67 $\pm$ 1.38\\
    ~ & NFL+RCE & 91.03 $\pm$ 0.15 & 89.10 $\pm$ 0.16 & 86.20 $\pm$ 0.19 & 79.58 $\pm$ 0.08 & 50.03 $\pm$ 2.78\\
    \cmidrule{2-7}
    ~ & NSL & 88.07 $\pm$ 0.12 & 86.46 $\pm$ 0.02 & 83.27 $\pm$ 0.13 & 76.17 $\pm$ 0.40 & 46.74 $\pm$ 0.72\\
    ~ & \textbf{CE+FNPAL} & 90.69 $\pm$ 0.11 & \textbf{86.34 $\pm$ 0.37} & \textbf{81.30 $\pm$ 0.29} & \textbf{72.77 $\pm$ 0.41} & \textbf{51.46 $\pm$ 1.10} \\
    ~ & \textbf{SCE+FNPAL} & 91.11 $\pm$ 0.13 & 87.30 $\pm$ 0.06 & \textbf{82.68 $\pm$ 0.22} & \textbf{73.49 $\pm$ 0.42} & \textbf{51.99 $\pm$ 1.10}\\
    ~ & \textbf{NCE+RCE+FNPAL} & 90.88 $\pm$ 0.10 & \textbf{89.34 $\pm$ 0.15} & \textbf{86.65 $\pm$ 0.21} & \textbf{80.28 $\pm$ 0.07} & \underline{\textbf{57.21 $\pm$ 0.22}}\\
    ~ & \textbf{NFL+RCE+FNPAL} & 91.16 $\pm$ 0.25 & \underline{\textbf{89.49 $\pm$ 0.32}} & \underline{\textbf{86.66 $\pm$ 0.08}} & \underline{\textbf{80.33 $\pm$ 0.15}} & \textbf{56.23 $\pm$ 0.15}\\
     \midrule
     \multirow{10}*{CIFAR100} 
       & CE & 70.41 $\pm$ 1.17 & 55.64 $\pm$ 0.17 & 40.39 $\pm$ 0.46 & 22.00 $\pm$ 1.23 & 7.37 $\pm$ 0.16\\
     ~ & FL & 70.56 $\pm$ 0.59 & 56.02 $\pm$ 0.80 & 40.41 $\pm$ 0.39 & 22.11 $\pm$ 0.30 & 7.70 $\pm$ 0.20\\
     ~ & GCE & 63.06 $\pm$ 1.00 & 62.15 $\pm$ 0.66 & 57.11 $\pm$ 1.43 & 45.99 $\pm$ 1.00 & 18.32 $\pm$ 0.36\\
     ~ & SCE & 70.41 $\pm$ 0.63 & 55.05 $\pm$ 0.68 & 39.60 $\pm$ 0.14 & 21.53 $\pm$ 0.72 & 7.82 $\pm$ 0.30 \\
     ~ & NCE+MAE & 67.16 $\pm$ 0.13 & 52.34 $\pm$ 0.12 & 35.81 $\pm$ 0.42 & 19.29 $\pm$ 0.29 & 7.31 $\pm$ 0.23\\
     ~ & NCE+RCE & 68.09 $\pm$ 0.26 & 64.32 $\pm$ 0.40 & 58.11 $\pm$ 0.63 & 45.94 $\pm$ 1.31 & \underline{25.22 $\pm$ 0.08}\\
     ~ & NFL+RCE & 67.58 $\pm$ 0.39 & 64.48 $\pm$ 0.50 & 57.86 $\pm$ 0.12 & 46.74 $\pm$ 0.59 & 24.55 $\pm$ 0.47\\
     \cmidrule{2-7}
     ~ & NSL & 70.08 $\pm$ 0.19 & 65.30 $\pm$ 0.36 & 56.77 $\pm$ 0.52 & 41.21 $\pm$ 1.01 & 12.16 $\pm$ 0.96\\
     ~ & \textbf{CE+FNPAL} & 71.69 $\pm$ 0.27 & \textbf{65.38 $\pm$ 0.17} & \textbf{57.24 $\pm$ 0.36} & \textbf{41.35 $\pm$ 0.19} & \textbf{12.12 $\pm$ 0.88}\\
     ~ & \textbf{SCE+FNPAL} & 70.87 $\pm$ 0.45 & \textbf{65.30 $\pm$ 0.15} & \textbf{55.10 $\pm$ 0.45} & \textbf{39.73 $\pm$ 0.04} & \textbf{11.70 $\pm$ 0.53}\\
     ~ & \textbf{NCE+RCE+FNPAL} & 69.29 $\pm$ 0.32 & \textbf{65.53 $\pm$ 0.30} & \textbf{60.53 $\pm$ 0.27} & \textbf{49.73 $\pm$ 0.64} & 24.54 $\pm$ 0.28\\
     ~ & \textbf{NFL+RCE+FNPAL} & 69.53 $\pm$ 0.05 & \underline{\textbf{65.94 $\pm$ 0.32}} & \underline{\textbf{60.89 $\pm$ 0.60}} & \underline{\textbf{50.10 $\pm$ 0.40}} & 24.15 $\pm$ 1.06\\
     \bottomrule
\end{tabular}
\end{table*}

\noindent\textbf{Results.}  The experiments follow the settings in \cite{zhong2021improving}. Tables \ref{tab:lt-imagenet} and \ref{tab:lt-cifar100} report the validation accuracy results of each method on CIFAR-100-LT and ImageNet-LT, respectively. As can be seen, our proposed PAL obviously improve the performance on both one-stage and two-stage methods. Particularly, the PAL-based MiSLAS outperform the ensemble method TADE and achieve the best results on CIFAR-100-LT, which demonstrate that PAL can significantly improve representation learning.

\subsection{Evaluation on Learning with Noisy Labels}
We evaluate the proposed method on MNIST, CIFAR-10/-100 with synthesis noisy labels. We follow the noise generation, networks and training settings in \cite{ma2020normalized}.

\begin{figure}[!t]
    \centering
    \subfigure[Validation accuracy]{
    \label{fig:cifar10_s0.6_norm}
    \includegraphics[scale=0.18]{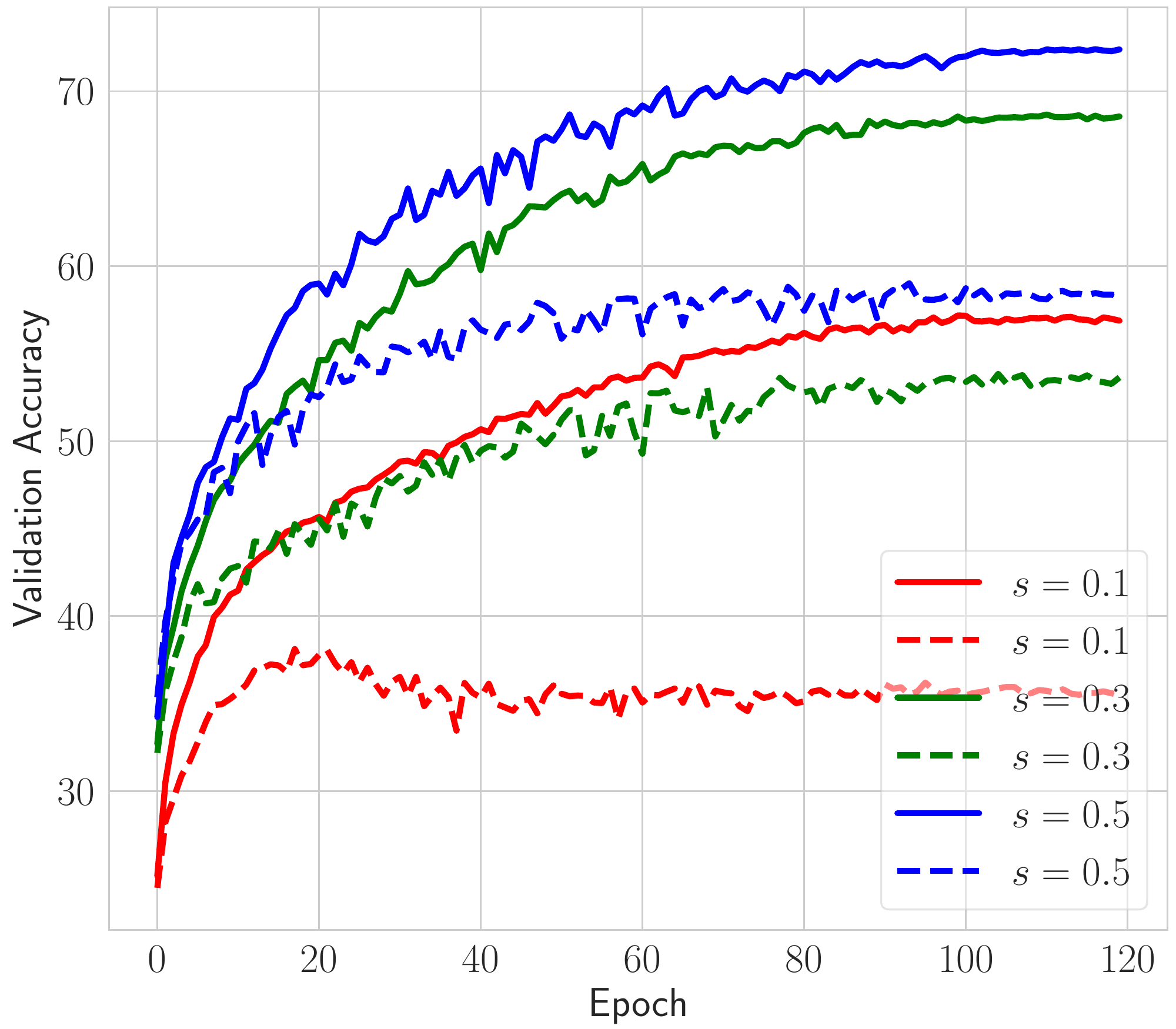}
      }
    \subfigure[The minimum angle]{
    \label{fig:cifar10_margin}
    \includegraphics[scale=0.18]{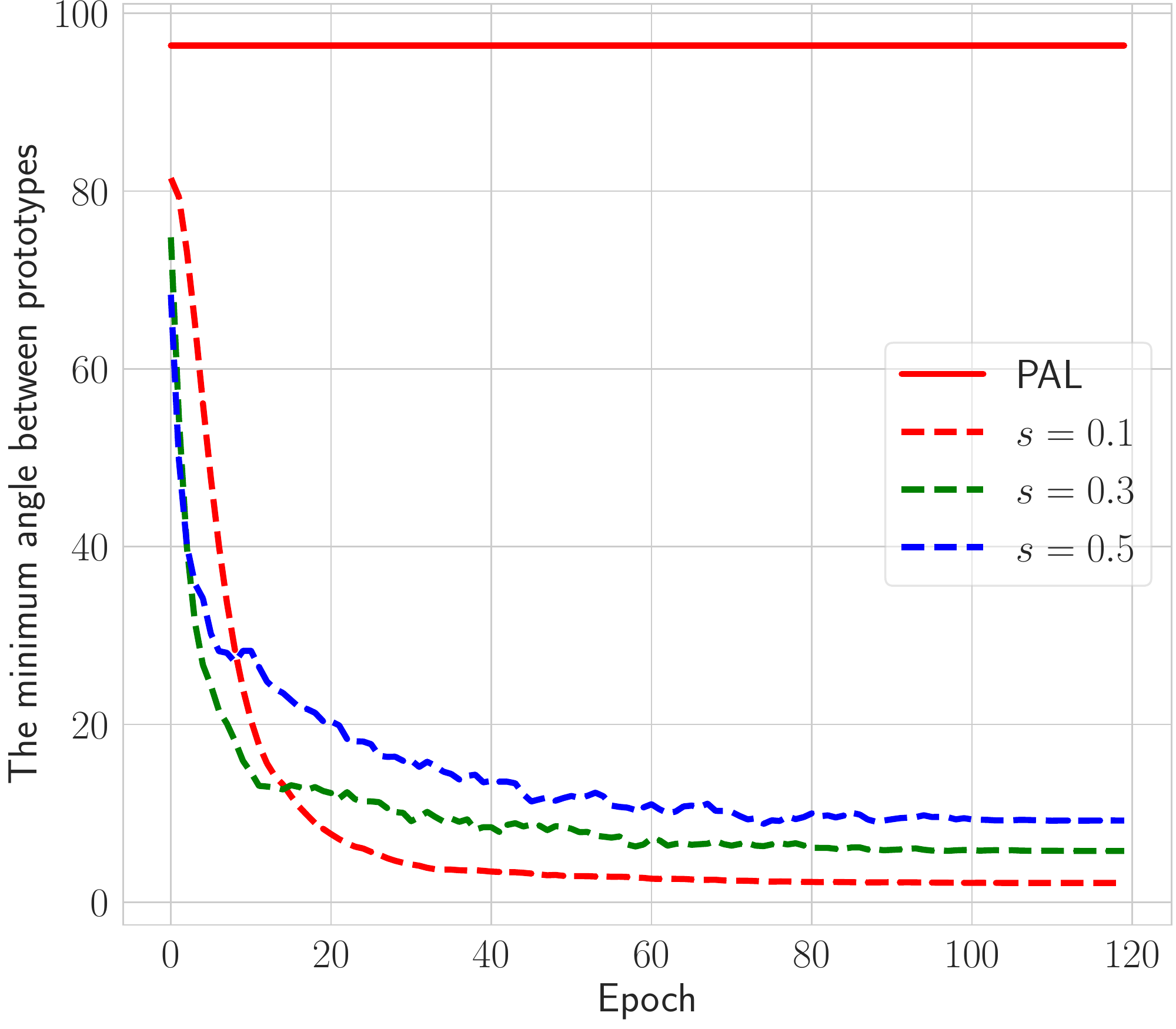}
    }
    \vskip-10pt
    \caption{Validation of FNPAL in mitigating label noise. Solid and dashed lines denote the curve of CE+FNPAL and CE+FNPN, respectively. (a) and (b) denote the curves of validation accuracy and the minimum angle between prototypes, respectively.}
    \label{FNPAL-is-better}
\end{figure}

\textbf{Baselines.} We consider several state-of-the-art methods: Generalized Cross Entropy (GCE) \cite{gce}, Symmetric Cross Entropy (SCE) \cite{wang2019sce}, and the Active Passive Loss (APL)  \cite{ma2020normalized} including NCE+MAE, NCE+RCE, and NFL+RCE. We also compare the commonly used losses CE and Focal Loss (FL). The best parameter settings for all baselines are verified by \cite{ma2020normalized}, thus we follow them. For our feature-normalized PAL (FNPAL) with CE and SCE, we utilize the noise-aware way, \textit{i.e.}, $s=\frac{0.25}{0.05+\eta}$, and $s=10$ for APLs.

\begin{table}[!t]
\vskip-8pt
\small
\setlength{\tabcolsep}{0.8mm}
\centering
\caption{Validation accuracies (\%) of different methods on benchmark datasets with asymmetric label noise ($\eta\in[0.2, 0.3, 0.4]$). The results (mean $\pm$ std) are reported over 3 random runs. C10 and C100 denote CIFAR-10 and CIFAR-100, respectively. The results with positive gains are \textbf{boldfaced} and the best one is \underline{underlined}. $^*$ denotes the method training with FNPAL.}
\label{asymmetric-noise}
\begin{tabular}{c|c|ccc}
    \toprule
     \multirow{2}*{Dataset} & \multirow{2}*{Method} & \multicolumn{3}{c}{Asymmetric Noise Rate ($\eta$)}  \\
     ~ & ~ & 0.2 & 0.3 & 0.4\\
     \midrule
     \multirow{9}*{MNIST} & 
        CE  & 94.42 $\pm$ 0.28 & 88.74 $\pm$ 0.07 & 81.45 $\pm$ 0.09\\
    ~ & FL  & 94.18 $\pm$ 0.05 & 88.87 $\pm$ 0.30 & 81.79 $\pm$ 0.36\\
    ~ & GCE & 96.57 $\pm$ 0.15 & 89.20 $\pm$ 0.11 & 81.92 $\pm$ 0.45\\
    ~ & SCE & 98.13 $\pm$ 0.17 & 93.59 $\pm$ 0.24 & 84.42 $\pm$ 0.57\\
    ~ & NCE+MAE & 98.87 $\pm$ 0.10 & 97.00 $\pm$ 0.15 & 91.33 $\pm$ 0.65\\
    ~ & NCE+RCE & 98.91 $\pm$ 0.08 & 97.37 $\pm$ 0.20 & 86.95 $\pm$ 4.27\\
    ~ & NFL+RCE & 98.87 $\pm$ 0.05 & 97.41 $\pm$ 0.12 & 92.35 $\pm$ 0.26\\
    \cmidrule{2-5}
    ~ & NSL & 98.79 $\pm$ 0.05 & 95.85 $\pm$ 0.12 & 87.41 $\pm$ 0.62\\
     ~ & \textbf{CE$^*$} & \textbf{98.58 $\pm$ 0.05} & \textbf{96.94 $\pm$ 0.27} & \textbf{91.77 $\pm$ 0.94}\\
     ~ & \textbf{SCE$^*$} & \underline{\textbf{99.08 $\pm$ 0.02}} & \underline{\textbf{98.43 $\pm$ 0.13}} & \textbf{92.11 $\pm$ 0.65}\\
     ~ & \textbf{NCE+RCE$^*$} & \textbf{98.93 $\pm$ 0.04} & \textbf{97.91 $\pm$ 0.29} & \textbf{94.37 $\pm$ 0.54}\\
     ~ & \textbf{NFL+RCE$^*$} & \textbf{99.02 $\pm$ 0.10} & \textbf{98.18 $\pm$ 0.19} & \underline{\textbf{94.86 $\pm$ 0.18}}\\
    \midrule
     \multirow{9}*{C10} & CE & 83.34 $\pm$ 0.32 & 79.58 $\pm$ 0.22 & 74.61 $\pm$ 0.42\\
     ~ & FL & 83.31 $\pm$ 0.23 & 79.67 $\pm$ 0.08 & 74.19 $\pm$ 0.40\\
     ~ & GCE & 85.91 $\pm$ 0.02 & 80.56 $\pm$ 0.30 & 74.49 $\pm$ 0.34\\
     ~ & SCE & 86.47 $\pm$ 0.14 & 81.61 $\pm$ 0.17 & 75.52 $\pm$ 0.28\\
     ~ & NCE+MAE & 86.52 $\pm$ 0.28 & 83.74 $\pm$ 0.19 & 76.75 $\pm$ 0.33\\
    ~ & NCE+RCE & 88.61 $\pm$ 0.27 & 85.55 $\pm$ 0.01 & 79.25 $\pm$ 0.33\\
    ~ & NFL+RCE & 88.58 $\pm$ 0.15 & 85.53 $\pm$ 0.26 & 79.61 $\pm$ 0.07\\
     \cmidrule{2-5}
     ~ & NSL & 86.23 $\pm$ 0.23 & 83.56 $\pm$ 0.15 & 78.63 $\pm$ 0.28\\
     ~ & \textbf{CE$^*$} & \textbf{86.11 $\pm$ 0.19} & \textbf{83.06 $\pm$ 0.06} & \textbf{77.41 $\pm$ 0.27}\\
     ~ & \textbf{SCE$^*$} & \textbf{87.32 $\pm$ 0.21} & \textbf{84.18 $\pm$ 0.10} & \textbf{78.15 $\pm$ 0.34}\\
     ~ & \textbf{NCE+RCE$^*$} & \underline{\textbf{89.01 $\pm$ 0.18}} & \textbf{85.97 $\pm$ 0.19} & \textbf{79.51 $\pm$ 0.19}\\
     ~ & \textbf{NFL+RCE$^*$} & \textbf{88.64 $\pm$ 0.17} & \underline{\textbf{86.20 $\pm$ 0.09}} & \underline{\textbf{79.66 $\pm$ 0.60}}\\
     \midrule
     \multirow{9}*{C100} & CE & 57.67 $\pm$ 0.89 & 50.63 $\pm$ 0.57 & 41.95 $\pm$ 0.12\\
     ~ & FL & 58.37 $\pm$ 0.56 & 51.44 $\pm$ 0.79 & 42.16 $\pm$ 0.53\\
     ~ & {GCE} & 59.56 $\pm$ 0.64 & 54.22 $\pm$ 1.18 & 42.18 $\pm$ 0.77\\
     ~ & SCE & 58.15 $\pm$ 0.54 & 50.58 $\pm$ 0.27 & 41.57 $\pm$ 0.19\\
     ~ & NCE+MAE & 52.03 $\pm$ 0.15 & 44.45 $\pm$ 0.22 & 36.86 $\pm$ 0.46 \\
    ~ & NCE+RCE & 62.93 $\pm$ 0.42 & 55.70 $\pm$ 0.14 & 42.73 $\pm$ 0.58\\
    ~ & NFL+RCE & 62.94 $\pm$ 0.39 & 55.83 $\pm$ 0.46 & 42.63 $\pm$ 0.13\\
     \cmidrule{2-5}
     ~ & NSL & 58.16 $\pm$ 0.46 & 50.35 $\pm$ 0.41 & 40.38 $\pm$ 0.60\\
     ~ & \textbf{CE$^*$} & \textbf{57.93 $\pm$ 0.18} & \textbf{51.45 $\pm$ 0.92} & \textbf{47.53 $\pm$ 0.94}\\
     ~ & \textbf{SCE$^*$} & \textbf{57.61 $\pm$ 0.22} & \textbf{50.74 $\pm$ 0.51} & \textbf{44.31 $\pm$ 0.22}\\
     ~ & \textbf{NCE+RCE$^*$} & \underline{\textbf{64.63 $\pm$ 0.32}} & \textbf{57.43 $\pm$ 0.53} & \textbf{49.77 $\pm$ 1.11}\\
     ~ & \textbf{NFL+RCE$^*$} & \textbf{64.32 $\pm$ 0.59} & \underline{\textbf{57.92 $\pm$ 0.65}} & \underline{\textbf{50.19 $\pm$ 1.60}}\\
     \bottomrule
\end{tabular}
\end{table}

\textbf{Feature-normalized PAL is better than one without PAL.} As aforementioned, feature normalization is important to mitigate label noise but is not sufficient for fitting ability. Proposition \ref{risk-bound-CE} shows that anchored prototypes will provide a tighter risk bound. To verify this, we run experiments on CIFAR-10 with 0.6 symmetric label noise by CE with feature normalization and prototype normalization (CE+FNPN) and CE with feature-normalized PAL (CE+FNPAL). Under the same $s$, CE+FNPAL always shows better performance than CE+FNPN as shown in \cref{fig:cifar10_s0.6_norm}. The main reason is that the prototypes of CE+FNPN are defective during training. In \cref{fig:cifar10_margin}, the minimum angle between prototypes of CE+FNPN is very small during training, \textit{i.e.}, there are at least two classes that are hard to classify.

\textbf{Results.} Tables \ref{symmetric-noise} and \ref{asymmetric-noise} report the validation accuracy results of different methods on the benchmark datasets with symmetric label noise and asymmetric label noise, respectively. As can be seen, our proposed FNPAL can significantly improve the baselines in most label noise settings on MNIST, CIFAR10/-100.  For example, CE+FNPAL performs much better than the original CE in all noisy cases. Moreover, we also add FNPAL to APLs (NCE+RCE and NFL+RCE), which surprisingly brings obvious positive gains and achieves the state-of-the-art methods in most cases. The results demonstrate that FNPAL can be robust enough to embed into the existing methods and get the outstanding performance for symmetric and asymmetric label noise.
 
\subsection{Evaluation on a Real-world Dataset}
We also evaluate our proposed prototype-anchored learning on a real-world imbalanced and noisy dataset WebVision 1.0 \cite{li2017webvision}. We follow the "mini" setting in \cite{jiang2018mentornet, ma2020normalized} that takes the first 50 concepts of the Google resized image subset as the training dataset and further validates the trained ResNet-50 \cite{he2016deep} on the same 50 concepts in validation set.

\begin{table}[htbp]
\vskip-12pt
\setlength{\tabcolsep}{0.65mm}
\small
\centering
\caption{Top-1 validation accuracies (\%) on mini-WebVision.}
\label{lnl-webvision}
\begin{tabular}{c|ccccccc}
    \toprule
    Method & CE & FL & NCE+RCE & \textbf{NSL} & \textbf{CE+PAL} & \textbf{CE+FNPAL}\\
    \midrule
    Acc & 62.60 & 63.80 & 66.32 & \textbf{69.56} & \textbf{68.92} & \textbf{69.69} \\
    \bottomrule
\end{tabular}
\end{table}

\textbf{Results.} The top-1 validation accuracies of different methods are reported in Table \ref{lnl-webvision}. Based on PAL, our proposed NSL shows very good performance, while CE+PAL significantly improve CE and outperform NCE+RCE a lot. Moreover, CE with feature-normalized PAL (CE+FNPAL) also brings 0.77 gains to CE+PAL. The results demonstrate that PAL can effectively help the trained model against real-world class imbalance and label noise. 


\section{Conclusion}
In this paper, we presented a simple yet effective method for learning with imperfect annotations, which is also theoretically sound. We formulate the goal of tightening the margin-based generalization bound as maximizing the minimal sample margin $\gamma_{\min}$, and provide the optimality condition of prototypes to maximize $\gamma_{\min}$, from which the proposed prototype-anchored learning (PAL) is derived.
Our PAL strategy can be easily embedded into various learning based classification under imperfect annotations. We specially consider two scenarios--class-imbalanced learning and learning with noisy labels.  Extensive experiments are provided to demonstrate the advantage of PAL. In future work, we will further explore the application of PAL on other classification problems under imperfect annotation.


\section*{Acknowledgments}
This work was supported by National Key Research and Development Project under Grant 2019YFE0109600, National Natural Science Foundation of China under Grants 61922027 and 6207115, and Beijing Municipal Science and Technology Commission Grant Z201100005820005.

\bibliography{preference}
\bibliographystyle{icml2022}

\onecolumn
\icmltitle{Appendix for ``Prototype-anchored Learning for Learning with Imperfect Data"}

\appendix
\section{Proof for Theorems and Propositions}
\subsection{Proof for Lemma \ref{largest-sample-margin}}
\begin{proof}
According to the definition of $\gamma_{\min}$, we have
$$
    \begin{aligned}
    \mathop{\arg\max}_{\bm{w}}\max_{\bm{z}}\gamma_{\min}&=\mathop{\arg\max}_{\bm{w}}\max_{\bm{z}}\min_{i}\bm{w}_{y_i}^\top\bm{z}_i-\max_{j\not=y_i}\bm{w}_j^\top\bm{z}_i\\
    &=\mathop{\arg\max}_{\bm{w}}\min_{i}\max_{\bm{z}_i}\bm{w}_{y_i}^\top\bm{z}_i-\max_{j\not=y_i}\bm{w}_j^\top\bm{z}_i\\
    &=\mathop{\arg\max}_{\bm{w}}\min_{i}\max_{\bm{z}_i}\bm{w}_{y_i}^\top\bm{z}_i-\bm{w}_{k}^\top\bm{z}_i\\
    &=\mathop{\arg\max}_{\bm{w}}\min_{i}\|\bm{w}_{y_i}-\bm{w}_{k}\|_2
    \end{aligned},
$$
where  $k\in\mathop{\arg\max}_{j\not=y_i} \bm{w}_{j}^{\top}\bm{z}_i$, and $\bm z_i=\frac{\bm{w}_{y_i}-\bm{w}_{k}}{\|\bm{w}_{y_i}-\bm{w}_{k}\|_2}$. Notice that $\bm{w}_{k}^\top\bm z_i=-\frac{1}{2}\|\bm w_{y_i}-\bm {w}_{k}\|_2$, then $k=\arg\min_{j\not =y_i} \|\bm w_{y_i}^\top-\bm {w}_j\|_2$. Therefore, we have
$$
    \begin{aligned}
    \mathop{\arg\max}_{\bm{w}}\max_{\bm{z}}\gamma_{\min}&=\max_{\bm{w}}\min_{i}\min_{k\not=y_i} \|\bm{w}_{y_i}-\bm{w}_k  \|_2=\mathop{\arg\max}_{\bm{w}}\min_{i\not=j} \|\bm{w}_{i}-\bm{w}_j\|_2,
    \end{aligned}
$$
i.e., maximizing $\gamma_{\min}$ will provide the solution of the Tammes Problem. According to the proof of \citet[Theorem 3.3.1]{borodachov2019discrete}, we have the above solution satisfies that $\bm{w}_i^{\top}\bm{w}_j=\frac{-1}{k-1}$, $\forall i\neq j$. Then we have $\bm{z}_i=\arg\max_{\bm{z}\in \sS^{d-1}}\bm{w}_{y_i}^\top \bm{z}-\max_{j\neq y}\bm{w}_j^\top \bm{z}=\bm{w}_{y_i}$, and $\gamma_{\min}=-\frac{k}{k-1}$. 

\end{proof}

\subsection{Proof for \cref{balanced-case-for-per-margin}}
\begin{proof}
Since the function $\exp$ is strictly convex, using the Jensen's inequality, we have
$$
\begin{aligned}
L&=\frac{1}{N}\sum_{i=1}^N-\log\frac{\exp(s\bm{w}_{y_i}^{\top}\bm{z}_i+\alpha)}{\exp(s\bm{w}_{y_i}^{\top}\bm{z}_i+\alpha)+\sum_{j\not=i}\exp(s\bm{w}_j^{\top}\bm{z}_i)}\\
&\ge \frac{1}{N}\sum_{i=1}^N-\log\frac{\exp(s\bm{w}_{y_i}^{\top}\bm{z}_i+\alpha)}{\exp(s\bm{w}_{y_i}^{\top}\bm{z}_i+\alpha)+(k-1)\exp(\frac{s}{k-1}\sum_{j\not=i}\bm{w}_j^{\top}\bm{z}_i)}\\
\end{aligned}
$$
Let $\overline{\bm{w}}=\frac{1}{k}\sum_{i=1}^k\bm{w}_i$, and $\sigma=\frac{k}{k-1}$, then we have
$$
\begin{aligned}
L&\ge \frac{1}{N}\sum_{i=1}^N\log\left[1+(k-1)\exp(s\sigma(\overline{\bm{w}}-\bm{w}_{y_i})^{\top}\bm{z}_i-\alpha\right]\\
&\ge \frac{1}{N}\sum_{i=1}^N \log[1+(k-1)\exp(-s\sigma\|\overline{\bm{w}}-\bm{w}_{y_i}\|_2-\alpha)]\\
&=\frac{1}{k}\sum_{i=1}^k \log[1+(k-1)\exp(-s\sigma\|\overline{\bm{w}}-\bm{w}_i\|_2-\alpha)]
\end{aligned},
$$
where we use the facts that  $(\overline{\bm{w}}-\bm{w}_{y_i})^{\top}\bm{z}_i\ge -\|\overline{\bm{w}}-\bm{w}_i\|_2$ when $\bm{z}_i\in\mathbb{S}^{d-1}$. Due the convexity of the function $\log[1+\exp(ax+b)]$ ($a>0$), we use the Jensen's inequality and obtain that
$$
\begin{aligned}
L&\ge\log \left[1+(k-1)\exp\left(-\frac{s\sigma}{k}\sum_{i=1}^{k}\|\overline{\bm{w}}- \bm{w}_i\|_2-\alpha\right)\right]\\
&\ge \log \left[1+(k-1)\exp\left(-\frac{s}{k-1}\sqrt{k\sum_{i=1}^{k}\|\overline{\bm{w}}- \bm{w}_i\|_2^2}-\alpha\right)\right]\\
&= \log\left[1+(k-1)\exp\left(-\frac{s}{k-1}\sqrt{k(k-k\|\overline {\bm{w}}\|_2^2\|_2^2)}-\alpha\right)\right]\\
&\ge \log[1 + (k-1)\exp(-s\sigma -\alpha)]\\
\end{aligned},
$$
where in the second inequality we used the Cauchy–Schwarz inequality, and the third inequality is based on that $k(k-k\|\overline {\bm{w}}\|_2^2\|_2^2\le k^2$.
According to the above derivation, the equality holds if and only if $\forall i$,  $\bm{w}_{1}^{\top}\bm{z}_i=...=\bm{w}_{y_i-1}^{\top}\bm{z}_i=\bm{w}_{y_i+1}^{\top}\bm{z}_i=...=\bm{w}_{k}^{\top}\bm{z}_i$, $\bm{w}_{y_i}^{\top}\bm{z}_i=1$, $\bm{z}_i=-\frac{\overline{\bm{w}}-\bm{w}_{y_i}}{\|\overline{\bm{w}}-\bm{w}_{y_i}\|_2}$,  $\|\overline{\bm{w}}-\bm{w}_1\|_2=...=\|\overline{\bm{w}}-\bm{w}_k\|_2$, and $\overline{\bm{w}}=0$. The condition can be simplified as \ $\forall i\not =j$, $\bm{w}_i^{\top}\bm{w}_j=\frac{-1}{k-1}$, and $\bm{z}_i=\bm{w}_{y_i}$ when $2\le d$ and $2\le k\le d+1$.
\end{proof}

\subsection{Proof for \cref{LDAM-not-CC}}
\begin{proof}
Considering the binary classification problem, the conditional risk with respect to the margin-based loss in Eq. \ref{margin-based-loss} is
\begin{equation}
    \begin{aligned}
    C(\eta_{\bm{x}}; \phi_\Theta(\bm{x}),\bm{w}_+,\bm{w}_-)
    =&-\eta_{\bm{x}}\log \frac{\exp(s\bm{w}_{+}^{\mathrm{T}}\phi_\Theta(\bm{x})+\alpha_+)}{\exp(s\bm{w}_{+}^{\mathrm{T}}\phi_\Theta(\bm{x})+\alpha_+)+\exp(s\bm{w}_{-}^{\mathrm{T}}\phi_\Theta(\bm{x}))}\\
    &-(1-\eta_{\bm{x}})\log \frac{\exp(s\bm{w}_{-}^{\mathrm{T}}\phi_\Theta(\bm{x})+\alpha_{-})}{\exp(s\bm{w}_{+}^{\mathrm{T}}\phi_\Theta(\bm{x}))+\exp(s\bm{w}_{-}^{\mathrm{T}}\phi_\Theta(\bm{x})+\alpha_{-})}
    \end{aligned}
\end{equation}
where $\bm{z}=\phi_\Theta(\bm{x}),\bm{w}_{+},\bm{w}_{-}\in\mathbb{S}^{d-1}$, $\alpha_{+}=\frac{C}{\pi_{+}^{1/4}}$, and $\alpha_{-}=\frac{C}{\pi_{-}^{1/4}}$. The Lagrangian function is $\mathcal L(\eta_{\bm{x}}; \bm{z},\bm{w}_+,\bm{w}_-)=C(\eta_{\bm{x}}; \phi_\Theta(\bm{x}),\bm{w}_+,\bm{w}_-) + \lambda(\|\bm{z}\|_2^2-1)$, and the gradient of $\mathcal L$ with respect to $\bm{z}$ can be derived as follows:
\begin{equation}
    \begin{aligned}
    \frac{\partial \mathcal L(\eta_{\bm{x}}; \bm{z},\bm{w}_+,\bm{w}_-)}{\partial \bm{z}}
    =&2\lambda \bm{z} + \frac{s\eta_{\bm{x}}\exp(s(\bm{w}_{-}-\bm{w}_{+})^{\mathrm{T}}\bm{z}-\alpha_{+})}{1+\exp(s(\bm{w}_{-}-\bm{w}_{+})^{\mathrm{T}}\bm{z}-\alpha_{+})}(\bm{w}_{-}-\bm{w}_{+})\\
    &+\frac{s(1-\eta_{\bm{x}})\exp(s(\bm{w}_{+}-\bm{w}_{-})^{\mathrm{T}}\bm{z}-\alpha_{-})}{1+\exp(s(\bm{w}_{+}-\bm{w}_{-})^{\mathrm{T}}\bm{z}-\alpha_{-})}(\bm{w}_{+}-\bm{w}_{-})\
    \end{aligned},
\end{equation}
therefore, the minimum of $C(\eta_{\bm{x}}; \phi_\Theta,\bm{w}_+,\bm{w}_-)$ will be obtained at $\frac{\bm{w}_{+}-\bm{w}_{-}}{\|\bm{w}_{+}-\bm{w}_{-}\|_2}$ or $-\frac{\bm{w}_{+}-\bm{w}_{-}}{\|\bm{w}_{+}-\bm{w}_{-}\|_2}$. 

As can be noticed, the prediction followed LDAM is +1 if $\frac{\exp(s\bm{w}_+^\top\bm{z})}{\exp(s\bm{w}_+^\top\bm{z})+\exp(s\bm{w}_-^\top\bm{z})}>\frac{\exp(s\bm{w}_-^\top\bm{z})}{\exp(s\bm{w}_+^\top\bm{z})+\exp(s\bm{w}_-^\top\bm{z})}$, and vice versa. This indicates that the Bayes-optimal prediction of LDAM is dependent on the sign of $(\bm w_+-\bm w_-)^{\mathrm{T}}\bm{z}$ with $\bm{z}=\arg\min_{\bm{z}\in\sS^{d-1}}C(\eta_{\bm{x}}; \bm{z},\bm{w}_+,\bm{w}_-)$. In the following, we are going to confirm the sign.

Assume that the minimizer of $C(\eta_{\bm{x}}; \phi_\Theta,\bm{w}_+,\bm{w}_-)$ is $t\cdot \frac{\bm{w}_{+}-\bm{w}_{-}}{\|\bm{w}_{+}-\bm{w}_{-}\|_2}$, where $t\in\{-1,+1\}$. Then the Bayes-optimal prediction of LDAM depends on the value of $t$. Actually, the value of $t$ relies not only on $\eta_{\bm{x}}$, but also on $\alpha_+$ and $\alpha_-$. 


Let $r=s\|\bm{w}_{+}-\bm{w}_{-}\|_2\in[0,2s]$, and $g(t;\eta_{\bm x}, r, \alpha_+,\alpha_-)=\eta_{\bm{x}}\log\left[1+\exp(-tr-\alpha_{+})\right]+(1-\eta_{\bm{x}})\log\left[1+\exp(tr-\alpha_{-})\right]$, then we have
\begin{equation}
    \begin{aligned}
    t &= \text{sign}(g(-1;\eta_{\bm x}, r, \alpha_+,\alpha_-)-g(1;\eta_{\bm x}, r, \alpha_+,\alpha_-))\\
    &= \text{sign}\bigg(\eta_{\bm{x}}\log \frac{1+\exp(r-\alpha_+)}{1+\exp(-r-\alpha_+)}+(1-\eta_{\bm{x}})\log\frac{1+\exp(-r-\alpha_-)}{1+\exp(r-\alpha_-)}\bigg)\\
    &=\text{sign}\left(\eta_{\bm{x}}-\frac{\log\frac{1+\exp(r-\alpha_-)}{1+\exp(-r-\alpha_-)}}{\log \frac{1+\exp(r-\alpha_+)}{1+\exp(-r-\alpha_+)}+\log\frac{1+\exp(r-\alpha_-)}{1+\exp(-r-\alpha_-)}}\right)
    \end{aligned}
\end{equation}
where $\text{sign}(\cdot)$ denotes the signum function.

As can be seen, if $a_+=a_-$, then the Bayes-optimal prediction of LDAM is $t=\text{sign}(\eta_{\bm{x}}-\frac{1}{2})$, which is the Bayes-optimal classifier, then LDAM is calibrated. However, under imbalanced data distribution, the definition is $\alpha_+\neq \alpha_-$, we know that $\frac{\log\frac{1+\exp(r-\alpha_-)}{1+\exp(-r-\alpha_-)}}{\log \frac{1+\exp(r-\alpha_+)}{1+\exp(-r-\alpha_+)}+\log\frac{1+\exp(r-\alpha_-)}{1+\exp(-r-\alpha_-)}}$ will not be equal to $\frac{1}{2}$, so LDAM is not calibrated.
\end{proof}

\subsection{Proof for \cref{largest-margin-imbalanced-case}}
\begin{proof}
For the margin-based loss $L_{\bm{\alpha}}=-\log \frac{\exp(s\bm{w}^{\mathrm{T}}_{y}\bm{z}+\alpha_{y})}{\exp(s\bm{w}^{\mathrm{T}}_{y}\bm{z}+\alpha_{y})+\sum\limits_{j\neq y}\exp(s\bm{w}^{\mathrm{T}}_j\bm{z})}$, we have
$$
\begin{aligned}
L_{\bm{\alpha}}(\bm{z}, y)&=-\log \frac{\exp(s\bm{w}^{\mathrm{T}}_{y}\bm{z}+\alpha_{y})}{\exp(s\bm{w}^{\mathrm{T}}_{y}\bm{z}+\alpha_{y})+\sum\limits_{j\neq y}\exp(s\bm{w}^{\mathrm{T}}_j\bm{z})}\\
&\ge -\log \frac{\exp(s\bm{w}^{\mathrm{T}}_{y}\bm{z}+\alpha_{y})}{\exp(s\bm{w}^{\mathrm{T}}_{y}\bm{z}+\alpha_{y})+(k-1)\exp[\frac{s}{k-1}(\sum_{j\neq y}\bm{w}_j)^{\mathrm{T}}\bm{z}]}\\
&= -\log \frac{\exp(s\bm{w}^{\mathrm{T}}_{y}\bm{z}+\alpha_{y})}{\exp(s\bm{w}^{\mathrm{T}}_{y}\bm{z}+\alpha_{y})+(k-1)\exp(-\frac{s}{k-1}\bm{w}_y^{\mathrm{T}}\bm{z})}\\
&= \log \left[1+(k-1)\exp\left(-\frac{sk}{k-1}\bm{w}_y^{\mathrm{T}}\bm{z}-\alpha_y\right)\right]\\
&\ge \log \left[1+(k-1)\exp\left(-\frac{sk}{k-1}-\alpha_y\right)\right]
\end{aligned}
$$
where in the first inequality we used the Jensen's inequality when $\exp(\cdot)$ is convex, the second equality is according to the condition that $\sum_{j=1}^k \bm{w}_j=0$ since $\bm{w}_i^{\top}\bm{w}_j=\frac{-1}{k-1}$, $\forall i\neq j$, and the last equality comes from the fact that $\bm{w}_y^{\mathrm{T}}\bm{z}\le 1$.

Therefore, we have the lower bound of the risk $\frac{1}{N}\sum_{i=1}^N L_{\bm{\alpha}}(\bm{z}_i,y_i;W)\ge \frac{1}{N}\sum_{i=1}^N \left[1+(k-1)\exp\left(-\frac{sk}{k-1}-\alpha_y\right)\right]$, where the equality holds if and only if $\forall i$, $\bm{w}_1^{\mathrm{T}}\bm{z}_i=\cdots=\bm{w}_{y_i-1}^{\mathrm{T}}\bm{z}_i=\bm{w}_{y_i+1}^{\mathrm{T}}\bm{z}_i=\cdots=\bm{w}_{k}^{\mathrm{T}}\bm{z}_i$, and $\bm{w}_{y_i}^{\mathrm{T}}\bm{z}_i=1$. Then we have $\gamma_{\min}=\frac{k}{k-1}$.
\end{proof}

\subsection{Proof for Theorem \cref{largest-margin-ce-imbalanced-case}}
\begin{proof}
Similar to the proof for Theorem 3.1, for the margin-based loss $L_{\bm{\alpha}}=-\log \frac{\exp(s\bm{w}^{\mathrm{T}}_{y}\bm{z}+\alpha_{y})}{\exp(s\bm{w}^{\mathrm{T}}_{y}\bm{z}+\alpha_{y})+\sum\limits_{j\neq y}\exp(s\bm{w}^{\mathrm{T}}_j\bm{z})}$, we have
$$
\begin{aligned}
L_{\bm{\alpha}}(\bm{z}, y)&\ge  \log \left[1+(k-1)\exp\left(-\frac{sk}{k-1}\bm{w}_y^{\mathrm{T}}\bm{z}-\alpha_y\right)\right]\\
&\ge \log \left[1+(k-1)\exp\left(-\frac{sk\|\bm{z}\|_2}{k-1}-\alpha_y\right)\right]\\
&\ge \log\left[1+(k-1)\exp\left(-\frac{skB}{k-1}-\alpha_y\right)\right]
\end{aligned}
$$
where the derivation above comes from the fact that $\bm{w}_y^{\mathrm{T}}\bm{z}\le \|\bm{z}\|_2\le B$.

Therefore, we have the lower bound of the risk $\frac{1}{N}\sum_{i=1}^N L_{\bm{\alpha}}(\bm{z}_i,y_i;W)\ge \frac{1}{N}\sum_{i=1}^N \left[1+(k-1)\exp\left(-\frac{skB}{k-1}-\alpha_y\right)\right]$, where the equality holds if and only if $\forall i$, $\bm{w}_1^{\mathrm{T}}\bm{z}_i=\cdots=\bm{w}_{y_i-1}^{\mathrm{T}}\bm{z}_i=\bm{w}_{y_i+1}^{\mathrm{T}}\bm{z}_i=\cdots=\bm{w}_{k}^{\mathrm{T}}\bm{z}_i$, $\frac{\bm{w}_{y_i}^{\mathrm{T}}\bm{z}_i}{\|\bm{z}_i\|_2}=1$, and $\|\bm{z}_i\|_2=B$ Then we have $\gamma_{\min}=\frac{kB}{k-1}$.
\end{proof}


\subsection{Proof for \cref{lipschitz-symmetric}}
\begin{proof}
Recall that for any $f$, we have $L(f(\bm{x}),i)=L(W^{\top}\bm{\phi}_{\Theta}(\bm{x}), i)$, $R_L(f)=\E_{\bm{x}, y}L(f(\bm{x}),y)$, and $L_W(\bm{\phi}_{\Theta}(\bm{x}))=\sum_{i=1}^k L(W^{\top}\bm{\phi}_{\Theta}(\bm{x}), i)$. Then for symmetric or uniform label noise, the noisy risk $R_L^\eta(f)$ satisfies that
$$
\begin{aligned}
    R_L^\eta(f)&=\E_{\bm{x},\tilde{y}}L(f(\bm{x}),\tilde{y})\\
    &=\E_{\bm{x}}\E_{y|\bm{x}}\E_{\tilde{{y}}|\bm{x},y} L(f(\bm{x}),\tilde{y})\\
    &=\E_{\bm{x}}\E_{y|\bm{x}}\left[(1-\eta)L(f(\bm{x},y)+\frac{\eta}{k-1}\sum_{i\neq y}L(f(\bm{x}),y)\right]\\
    &=(1-\eta)R_L(f)+\frac{\eta}{k-1}\left[\E_{\bm{x}}L_W(\bm{\phi}_{\Theta}(\bm{x}))-R_L(f)\right]
\end{aligned},
$$
that is, 
$$
    \begin{aligned}
    (1-\frac{\eta k}{k-1})R_L(f)=R_L^\eta(f)-\frac{\eta}{k-1}\E_{\bm{x}} L_W(\bm{\phi}_{\Theta}(\bm{x}))
    \end{aligned}.
$$
For $f^*\in \arg\min_{f\in \gH} R_L(f)$ and $\hat{f} \in \arg\min_{f\in gH}R_L^\eta (f)$, we then obtain
$$
    \begin{aligned}
    (1-\frac{\eta k}{k-1})[R_L(\hat{f})-R_L(f^*)]&= R_L^\eta(\hat{f})-R_L^\eta(f^*)-\frac{\eta}{k-1}[\E_{\bm{x}}L_W(\bm{\phi}_{\hat{\Theta}}(\bm{x}))-\E_{\bm{x}}L_W(\bm{\phi}_{\Theta^*}(\bm{x}))]\\
    &\le \frac{\eta}{k-1} \E_{\bm{x}}|L_W(\bm{\phi}_{\hat{\Theta}}(\bm{x}))-L_W(\bm{\phi}_{\Theta^*}(\bm{x}))|\\
    &\le \frac{\eta\lambda}{k-1}\E_{\bm{x}}\| \bm{\phi}_{\hat{\Theta}}(\bm{x})-\bm{\phi}_{\Theta^*}(\bm{x})\|_2\le \frac{2\eta\lambda B}{k-1}
    \end{aligned},
$$
where we use the facts that $R_L^\eta(\hat{f})-R_L^\eta(f^*)\le 0$, $L_{W}(\bm{z})$ is $\lambda$-Lipschitz, and $\| \bm{\phi}_{\hat{\Theta}}(\bm{x})-\bm{\phi}_{\Theta^*}(\bm{x})\|_2\le 2B$ when $\|\bm{\phi}_{\Theta}(\bm{x})\|_2=B$. Therefore, rearranging the above equation will obtain the risk bound in \cref{risk-bound}.
\end{proof}

\subsection{Proof for Proposition \ref{risk-bound-CE}}
\label{risk-bound-ce-gce-focal}
\begin{proof}
For the softmax loss in \cref{softmax-loss}, we have
\begin{equation}
\label{ce}
    L_W(\bm{z})=-\sum_{i=1}^k\log \frac{\exp(\bm{w}_i^\top \bm{z})}{\sum_{j=1}^k \exp(\bm{w}_j^\top\bm{z})},
\end{equation}
then the derivative of $L_W(\bm{z})$ with respect to $\bm{z}$ is
$$
    \frac{\partial L_W(\bm{z})}{\partial \bm{z}}=\sum_{i=1}^k\left[\bm{w}_i-\sum_{j=1}^k \frac{\exp(\bm{w}_j^\top\bm{z})}{\sum_{t=1}^k\exp(\bm{w}_t^\top\bm{z})}\bm{w}_j\right],
$$
since $\bm{w}_1,...,\bm{w}_k\in\sS^{d-1}$ satisfy $\bm{w}_i^{\top}\bm{w}_j=\frac{-1}{k-1}$, then $\sum_{i=1}^k \bm{w}_i=\bm{0}$ and $\frac{\partial L_W(\bm{z})}{\partial \bm{z}}=k\sum_{j=1}^k \frac{\exp(\bm{w}_i^\top\bm{z})}{\sum_{i=1}^k\exp(\bm{w}_j^\top\bm{z})}\bm{w}_i$. The Lipschitz constant of $L_W(\bm{z})$ will depend on the upper bound of $\|\frac{\partial L_W(\bm{z})}{\partial \bm{z}}\|_2$, and this bound exists when $\bm{w}_i\in \sS^{d-1}$ and $\|\bm{z}\|_2\le B$. Actually, let $\alpha_i=\frac{\exp(\bm{w}_i^\top\bm{z})}{\sum_{j=1}^k\exp(\bm{w}_j^\top\bm{z})}$, then $\alpha_i\in [\frac{1}{1+(k-1)\exp(\frac{kB}{k-1})}, \frac{1}{1+(k-1)\exp(-\frac{kB}{k-1})}]$, $\sum_{i=1}^k\alpha_i=1$, and we have
$$
\begin{aligned}
\left\|\frac{\partial L_W({\bm{z}})}{\partial \bm{z}}\right\|_2^2&=k^2\| \sum_{i=1}^k \alpha_i \bm{w}_i\|_2^2=k^2 \sum_{i=1}^k\sum_{j=1}^k \alpha_i\alpha_j \bm{w}_i^\top \bm{w}_j\\
&=k^2 \left[\sum_{i=1}^k\alpha_i^2 -\frac{1}{k-1}\sum_{i\neq j}\alpha_i\alpha_j\right]=k^2 \left[\frac{k}{k-1}\sum_{i=1}^k\alpha_i^2-\frac{1}{k-1}\right]\\
&\le k^2\left[\frac{k}{k-1}\left(\frac{1}{[1+(k-1)\exp(-\frac{kB}{k-1})]^2}+\frac{k-1}{[k-1+\exp(\frac{kB}{k-1})]^2}\right)-\frac{1}{k-1}\right]\\
& = k^2\left(\frac{1-\exp(-\frac{kB}{k-1})}{1+(k-1)\exp(-\frac{kB}{k-1})}\right)^2
\end{aligned},
$$
thus, the Lipschitz constant of $L_W(\bm{z})$ is $\lambda_{\text{PAL}}=\frac{k(1-\exp(-\frac{kB}{k-1}))}{1+(k-1)\exp(-\frac{kB}{k-1})}$, then we have the following risk according to \cref{lipschitz-symmetric}:
\begin{equation}
    R_L(\hat{f})-R_L(f^*)\le \frac{2c\eta k(1-t)B}{k-1+t(k-1)^2},
\end{equation}
where $c=\frac{k-1}{(1-\eta)k-1}$, and $t=\exp(-\frac{kB}{k-1})$.
\end{proof}

\subsubsection{More Analysis about Other Losses}
For a general loss, we usually have $L_i = \ell(\alpha_i)$, where $\alpha_i=\frac{\exp(\bm{w}_i^\top\bm{z})}{\sum_{j=1}^k\exp(\bm{w}_j^\top\bm{z})}\in [0,1]$ denotes the predicted probability of class $i$. Then the derivative of $L=\sum_{i=1}^k L_i$ with respect to $\bm{z}$ can be derived as
\begin{equation}
    \begin{aligned}
    \frac{\partial L}{\partial \bm{z}}&=\sum_{i=1}^k \frac{\partial L_i}{\partial \alpha_i} \frac{\partial \alpha_i}{\partial \bm{z}}=\sum_{i=1}^k \frac{\partial \ell(\alpha_i)}{\partial \alpha_i} \cdot\alpha_i\left(\bm{w}_i-\sum_{j=1}^k\alpha_j \bm{w}_j\right)\\
    &=\left(\sum_{i=1}^k \alpha_i\cdot\frac{\partial \ell(\alpha_i)}{\partial \alpha_i}\right)\left(\sum_{i=1}^k\frac{\alpha_i\cdot\frac{\partial \ell(\alpha_i)}{\partial \alpha_i}}{\sum_{j=1}^k \alpha_i\cdot\frac{\partial \ell(\alpha_j)}{\partial \alpha_j}}\bm{w}_i-\sum_{i=1}^k\alpha_i\bm{w}_i\right)
    \end{aligned}.
\end{equation}
As can be seen, to make $L$ be Lipschitz continuous, the key point is to guarantee that $\sum_{i=1}^k \alpha_i\cdot\frac{\partial \ell(\alpha_i)}{\partial \alpha_i}$ and $\frac{\alpha_i\cdot\frac{\partial \ell(\alpha_i)}{\partial \alpha_i}}{\sum_{j=1}^k \alpha_i\cdot\frac{\partial \ell(\alpha_j)}{\partial \alpha_j}}$ are also bounded. In the following, we will analyze GCE \cite{gce} and Focal loss \cite{lin2017focal}.

\paragraph{GCE.}  GCE is formulated as $\ell_i(\alpha_i)=\frac{1-\alpha_i^q}{q}$ ($0< q\le 1$). The derivative of $\ell_i$ w.r.t $\alpha_i$ is $\frac{\partial \ell_i}{\partial \alpha_i}=-\alpha_i^{q-1}$, then $\frac{\alpha_i\cdot\frac{\partial \ell(\alpha_i)}{\partial \alpha_i}}{\sum_{j=1}^k \alpha_i\cdot\frac{\partial \ell(\alpha_j)}{\partial \alpha_j}}=\frac{\alpha_i^q}{\sum_{j=1}^k\alpha_j^q} \in[0,1]$ is bounded, and we also have $\sum_{i=1}^k \alpha_i\cdot\frac{\partial \ell(\alpha_i)}{\partial \alpha_i}=-\sum_{i=1}^k \alpha_i^q \in[-1, 0]$. Thus, GCE loss is Lipschitz continuous.

\paragraph{Focal Loss.} Focal loss is formulated as $\ell_i=-(1-\alpha_i)^\gamma\log \alpha_i$ ($\gamma >0$). We have $\alpha_i\frac{\partial \ell_i}{\partial \alpha_i}=-(1-\alpha_i)^{\gamma-1}(1-\alpha_i-\gamma\alpha_i\log \alpha_i)<0$. As can be seen, when $\alpha_i\rightarrow 0$ ($\gamma >0$) or $\alpha_i\rightarrow 1$ ($\gamma <1$), $\alpha_i\frac{\partial \ell_i}{\partial \alpha_i}$ will be unbounded. This issue can be alleviated when $\bm{w}_1,...,\bm{w}_k\in\sS^{d-1}$ satisfy $\bm{w}_i^{\top}\bm{w}_j=\frac{-1}{k-1}$, since we have $\alpha_i\in [\frac{1}{1+(k-1)\exp(\frac{kB}{k-1})}, \frac{1}{1+(k-1)\exp(-\frac{kB}{k-1})}]$, where $\|\bm{z}\|_2\le B$.



\subsubsection{About the Tighter Bound Indicated by PAL}
\label{a-tighter-bound-indicated-by-PAL}
In this section, we will prove that PAL will lead to a smaller Lipschitz constant of $L_{W}$ in \cref{ce} than unnormalized or unacnhored cases. Note that $\lambda_{\text{PAL}}=\frac{k(1-\exp(-\frac{kB}{k-1}))}{1+(k-1)\exp(-\frac{kB}{k-1})}$, we have:
\begin{enumerate}
    \item \textbf{When $\bm{w}$ is unnormalized}. The Lipschitz constant of $L_{W}$ is $\sup_{W,\bm{z}}\left\|\sum_{i=1}^k\bm{w}_i-k\sum_{j=1}^k \frac{\exp(\bm{w}_j^\top\bm{z})}{\sum_{t=1}^k\exp(\bm{w}_t^\top\bm{z})}\bm{w}_j\right\|_2$, which can be infinitely large as $\bm{w}_i=t\bm{z}$ ($t\rightarrow \infty$) and $\bm{w}_j$ ($j\neq i$) is fixed. 
    \item \textbf{When $\bm{w}_i$ is normalized but unanchored (\textit{w.l.o.g}, $\|\bm{w}_i\|_2=1$ and $\bm{z}$ is unnormalized).} We have the Lipschitz constant $\sup_{W,\bm{z}}\|\frac{\partial L_{W}(\bm{z})}{\partial \bm{z}}\|_2\ge k$ (which can be obtained at $\sum_{i=1}^k\bm{w}_i=0$ and $\max_i \frac{\exp(\bm{w}_i^T\bm{z})}{\sum_{j=1}^k \exp(\bm{w}_j^T\bm{z})}=1$) that is larger than $\lambda_{\text{PAL}}$; 
    \item \textbf{When both $\bm{w}_i$ and $\bm{z}$ are normalized and unanchored ($\|\bm{w}_i\|_2=1$, $\|\bm{z}\|_2=B$).} Considering that $\bm{w}_1=\frac{\bm{z}}{\|\bm{z}\|_2}$, and $\bm{w}_2=...\bm{w}_k=-\frac{\bm{z}}{\|\bm{z}\|_2}$, we have
    $$
        \left\|\frac{\partial L_{W}(\bm{z})}{\partial \bm{z}}\right\|_2=\frac{2(\exp(2B)-1)}{\frac{\exp(2B)}{k-1}+1} > \lambda_{\text{PAL}},
    $$
    where $\frac{2(\exp(2B)-1)}{\frac{\exp(2B)}{k-1}+1} > \frac{k(1-\exp(-\frac{kB}{k-1}))}{1+(k-1)\exp(-\frac{kB}{k-1})}$ is equivalent to 
    \begin{equation}
       \begin{aligned}
            &(\exp(2B)-1)\left[1+(k-1)\exp\left(\frac{-k}{k-1}B\right)\right] > \frac{k}{2}\left[1-\exp\left(\frac{-k}{k-1}B\right)\right]\left[\frac{\exp(2B)}{k-1}+1\right]\\
            \Leftrightarrow&\left(1-\frac{k}{2k-2}\right)\exp(2B)+\left(k-1+\frac{k}{2k-2}\right)\exp\left(\frac{k-2}{k-1}B\right) > \frac{k}{2}+1+\left(k-1-\frac{k}{2}\right)\exp\left(-\frac{k}{k-1}B\right)
       \end{aligned},
    \end{equation}
    since $k\ge 2$ and $B>0$, we have $\text{LHS}>(1-\frac{k}{2k-2})+ (k-1+\frac{k}{2k-2})=k\ge \frac{k}{2} +1 + \left(k-1-\frac{k}{2}\right)\exp\left(-\frac{k}{k-1}B\right)=\text{RHS}$.
\end{enumerate}

Therefore, PAL will lead to a smaller Lipschitz constant, and then a tighter risk bound.

\section{More Experimental Details, Analysis and Results}
\label{more-experiments}
\begin{figure*}[!t]
\centering
    \subfigure[CE with wd=0.0]{
        \label{fig:ECE-CE-0.0}
        \includegraphics[scale=0.23]{figures/confidences/CE/CE_0.0.pdf}
    }
    \subfigure[CE with wd=5e-5]{
        \label{fig:ECE-CE-5e-5}
        \includegraphics[scale=0.23]{figures/confidences/CE/CE_5e-05.pdf}
    }
    \subfigure[CE with wd=5e-4]{
        \label{fig:ECE-CE-5e-4}
        \includegraphics[scale=0.23]{figures/confidences/CE/CE_0.0005.pdf}
    }
    \subfigure[CE with wd=1e-3]{
        \label{fig:ECE-CE-1e-3}
        \includegraphics[scale=0.23]{figures/confidences/CE/CE_0.001.pdf}
    }
  \\
   \subfigure[CE+DRW with wd=0.0]{
        \label{fig:ECE-CE-DRW-0.0}
        \includegraphics[scale=0.23]{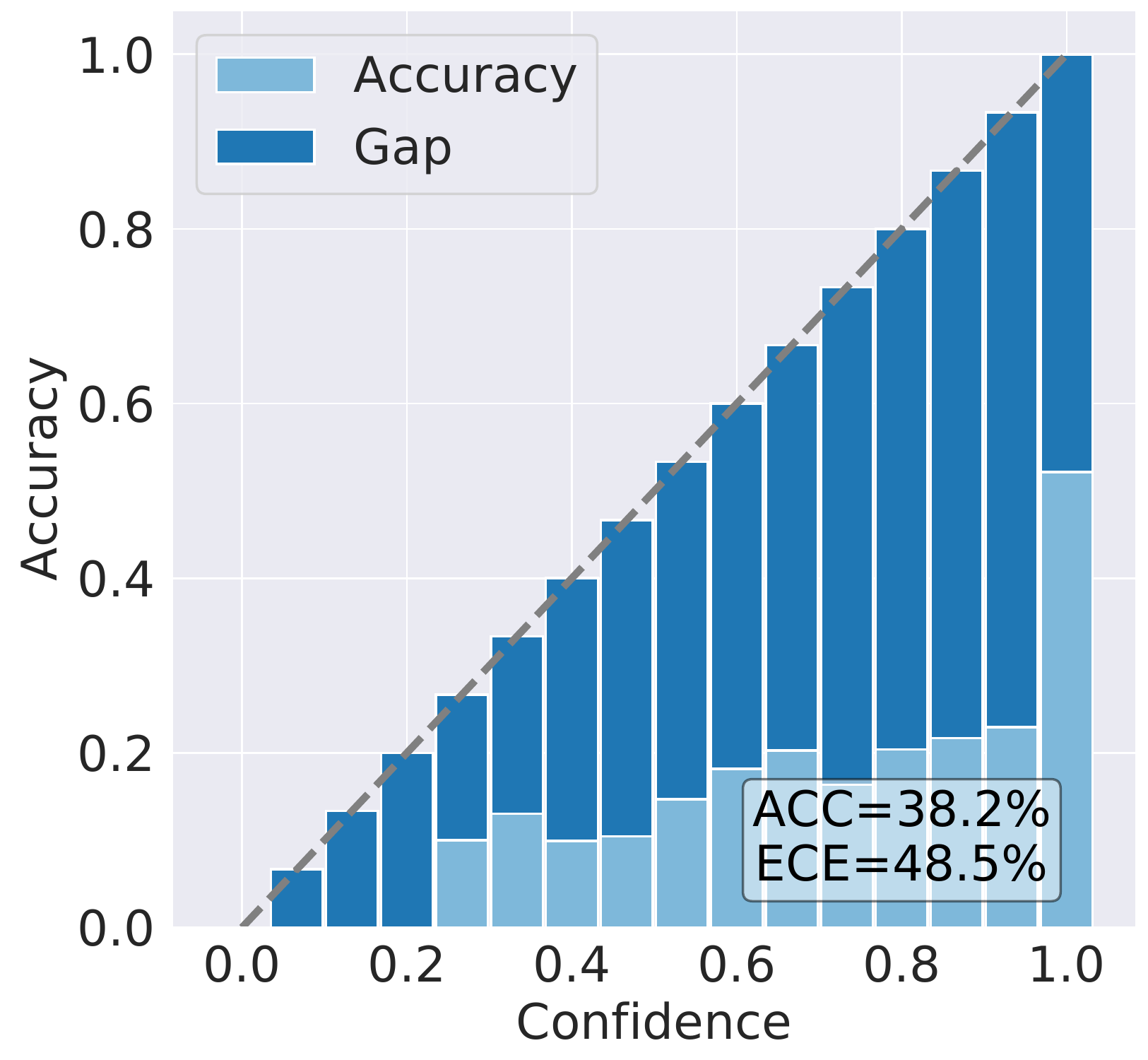}
    }
    \subfigure[CE+DRW with wd=5e-5]{
        \label{fig:ECE-CE-DRW-5e-5}
        \includegraphics[scale=0.23]{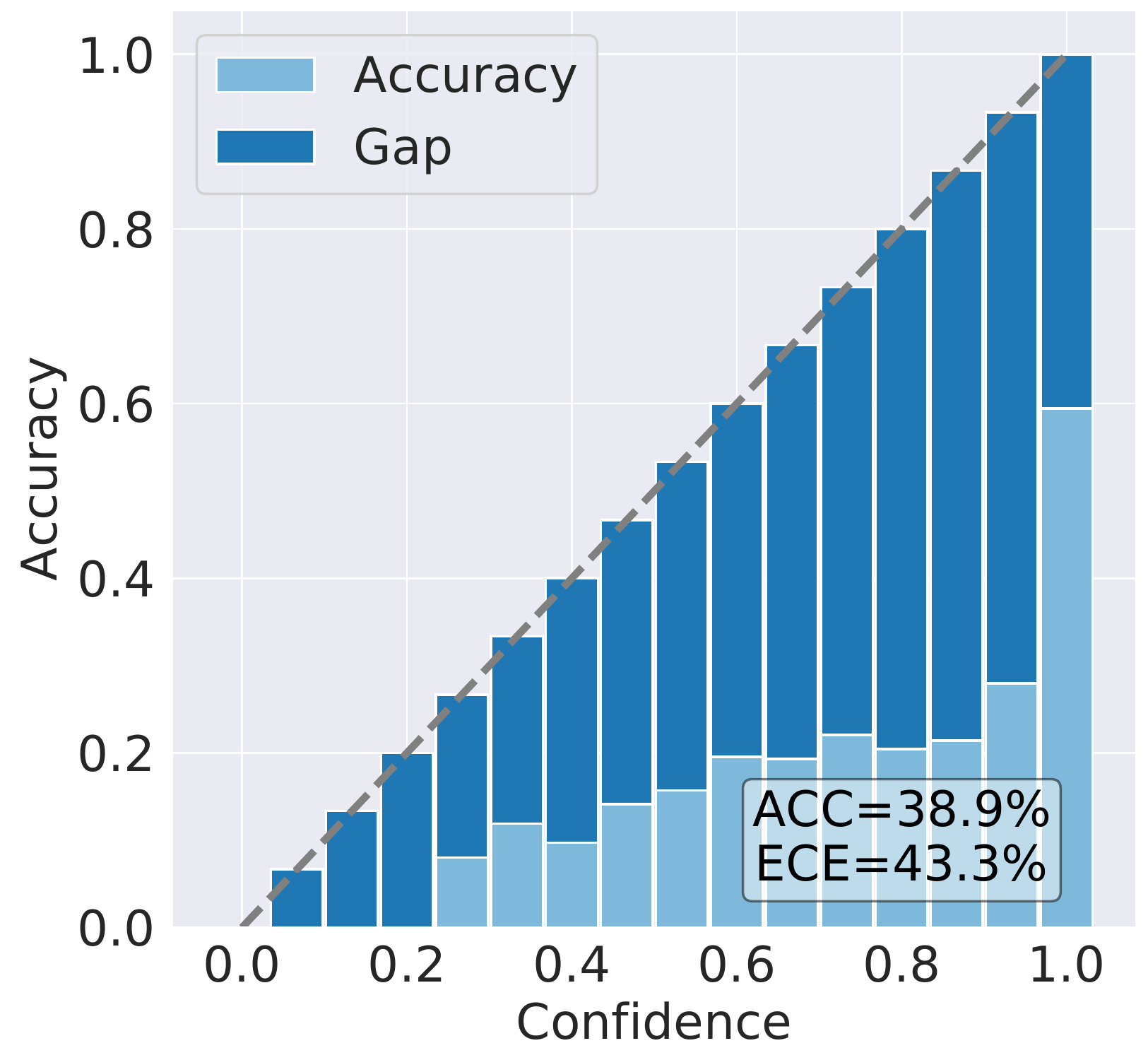}
    }
    \subfigure[CE+DRW with wd=5e-4]{
        \label{fig:ECE-CE-DRW-5e-4}
        \includegraphics[scale=0.23]{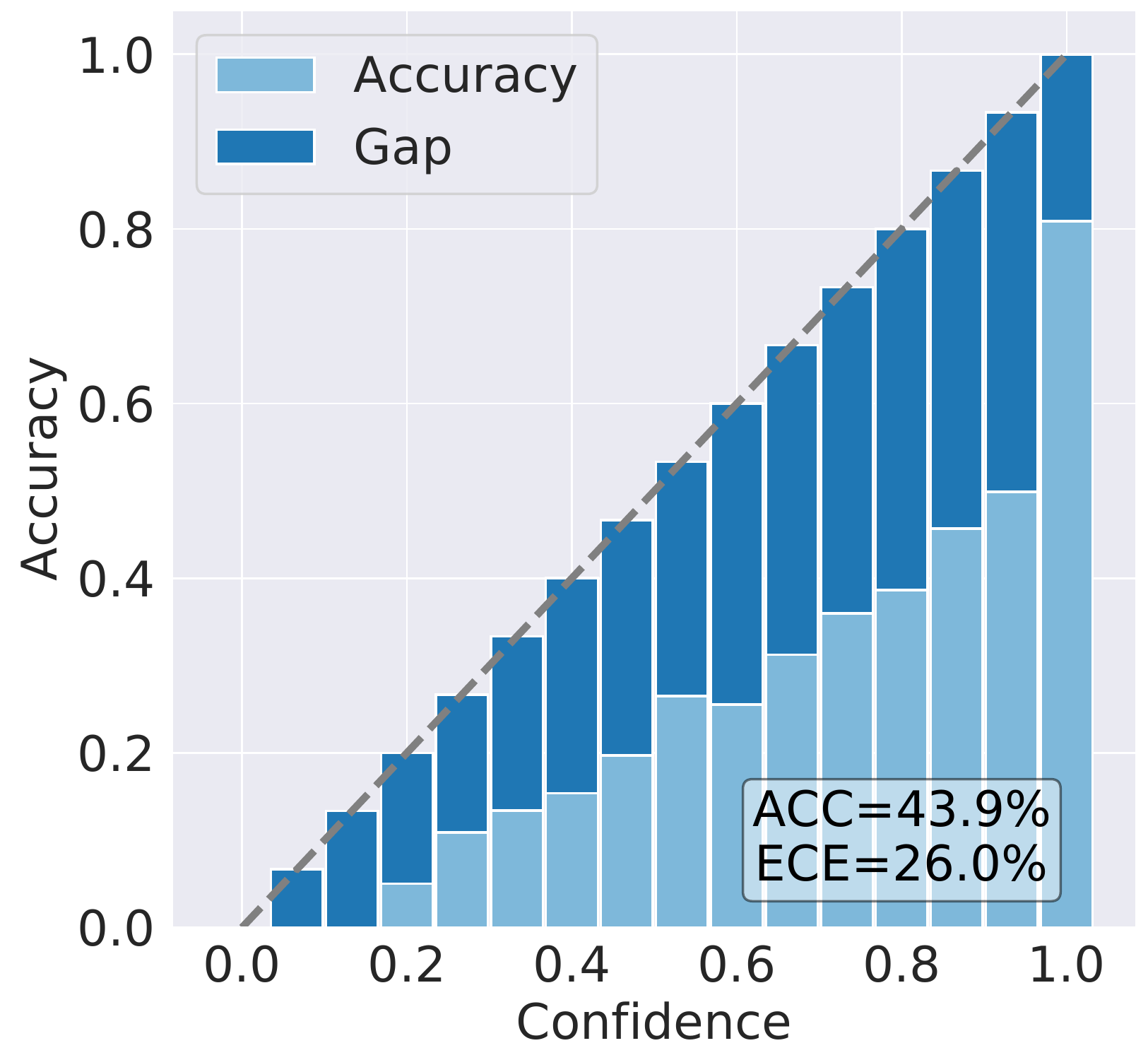}
    }
    \subfigure[CE+DRW with wd=1e-3]{
        \label{fig:ECE-CE-DRW-1e-3}
        \includegraphics[scale=0.23]{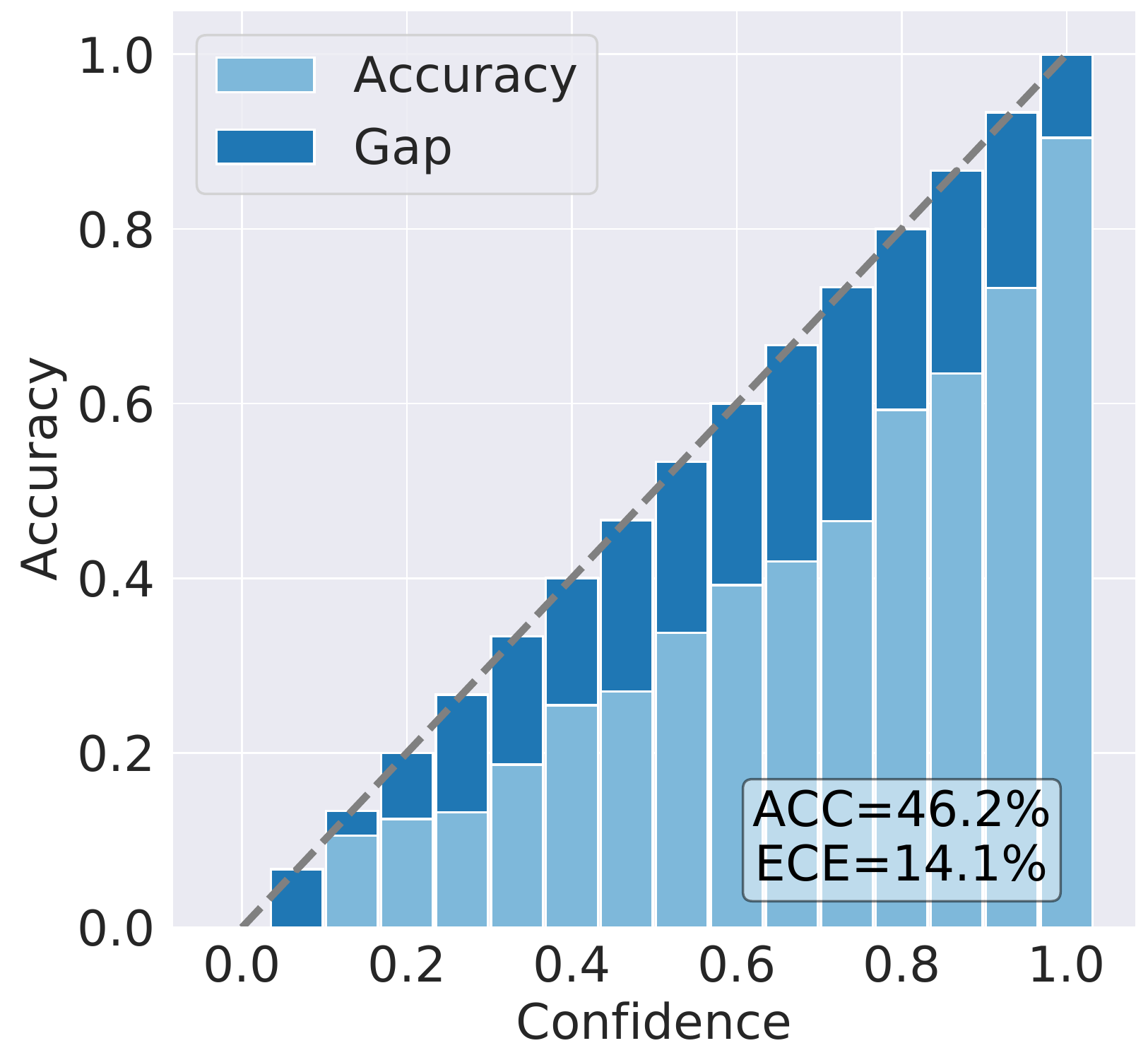}
    }
    \\
    \subfigure[B-Softmax with wd=0.0]{
        \label{fig:ECE-BSoftmax-0.0}
        \includegraphics[scale=0.23]{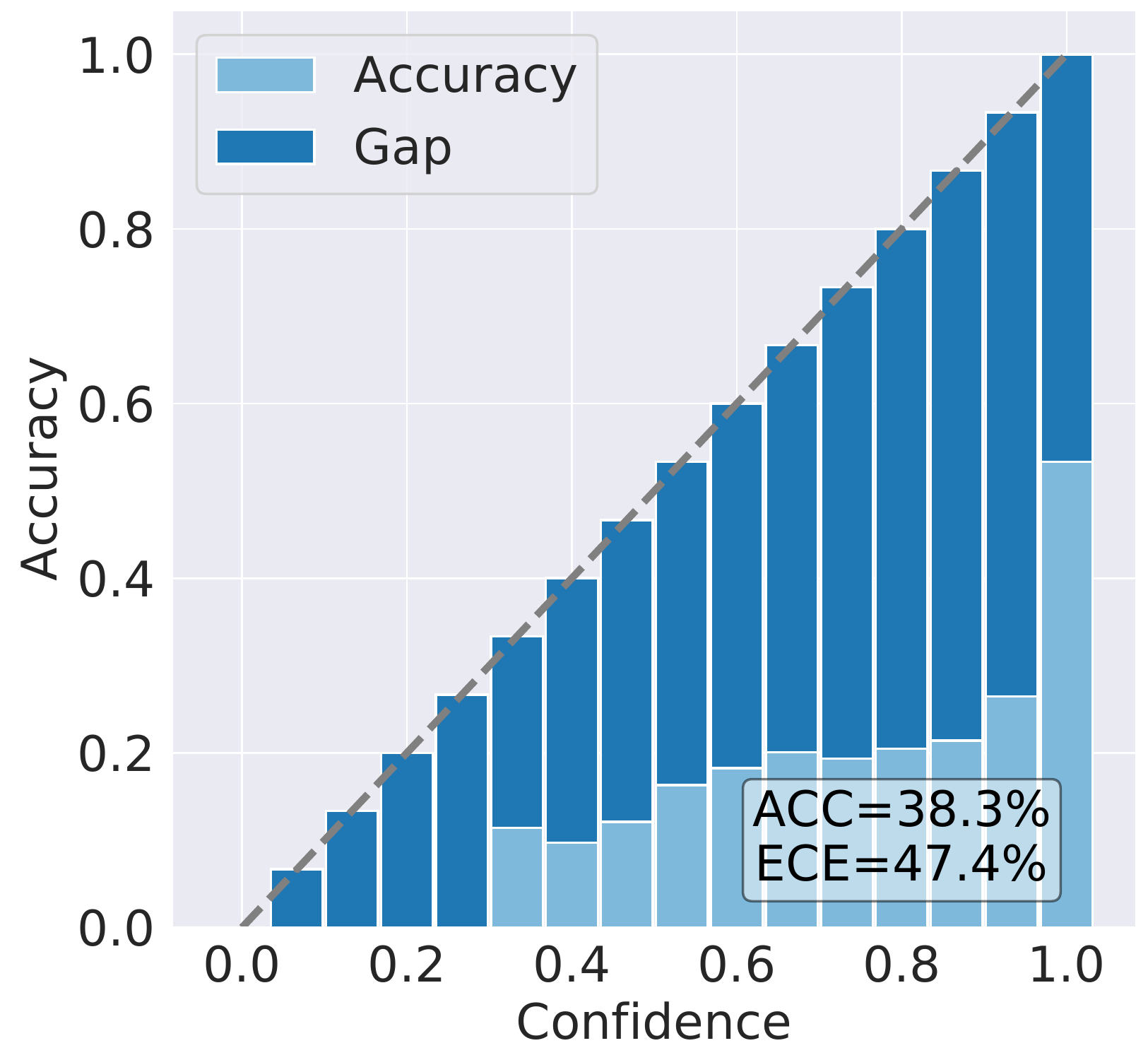}
    }
    \subfigure[B-Softmax with wd=5e-5]{
        \label{fig:ECE-BSoftmax-5e-5}
        \includegraphics[scale=0.23]{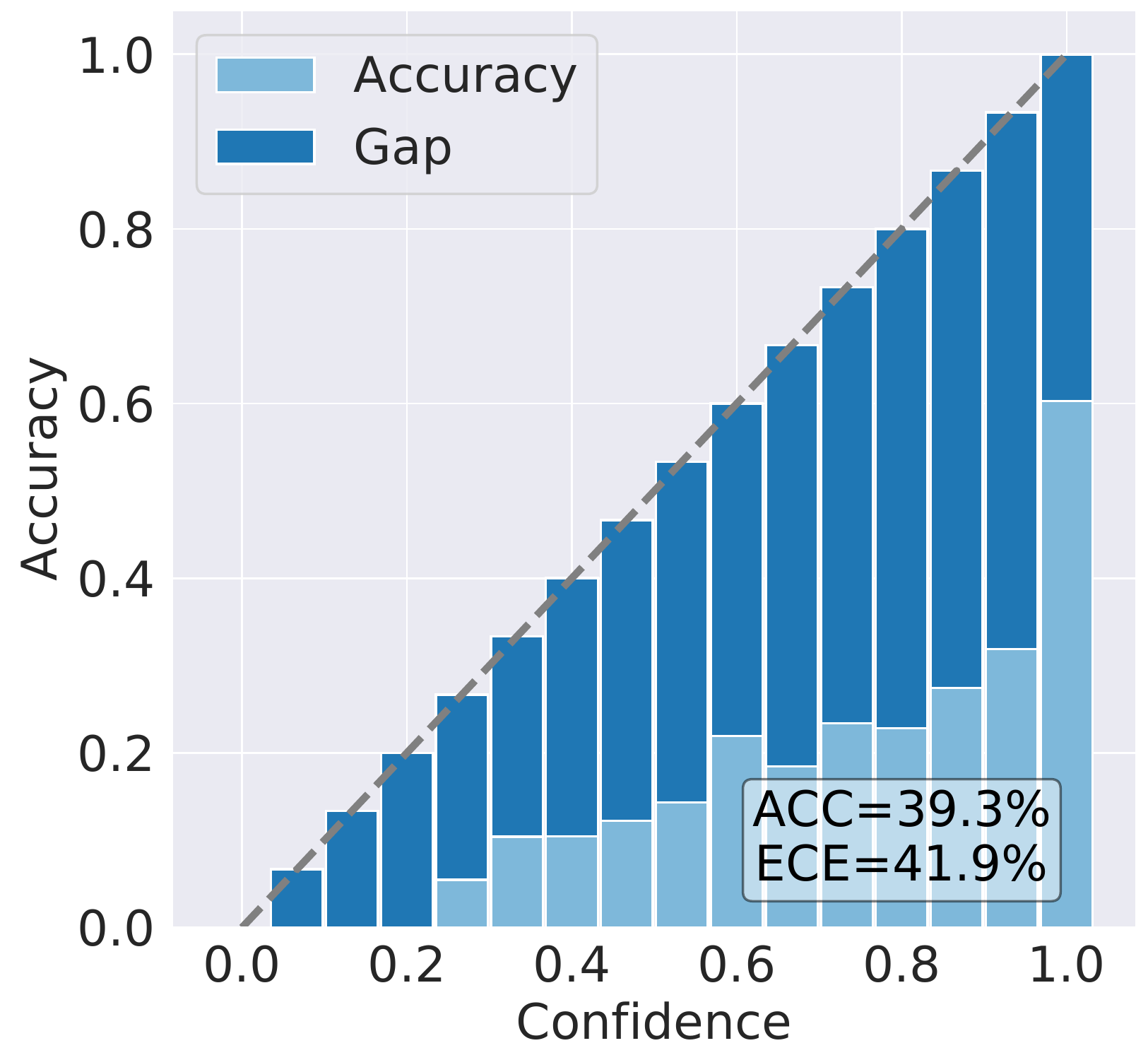}
    }
    \subfigure[B-Softmax with wd=5e-4]{
        \label{fig:ECE-BSoftmax-5e-4}
        \includegraphics[scale=0.23]{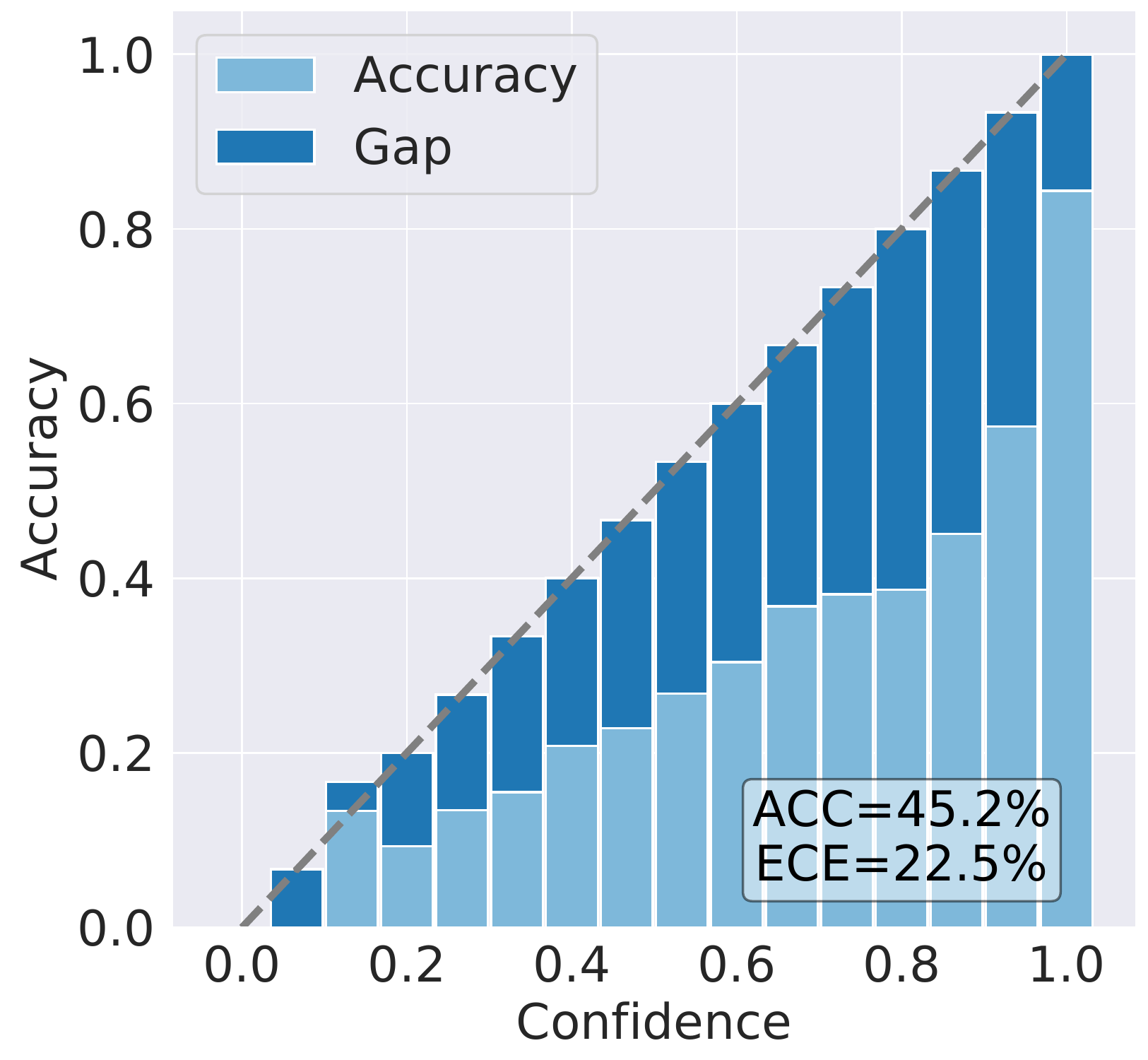}
    }
    \subfigure[B-Softmax with wd=1e-3]{
        \label{fig:ECE-BSoftmax-1e-3}
        \includegraphics[scale=0.23]{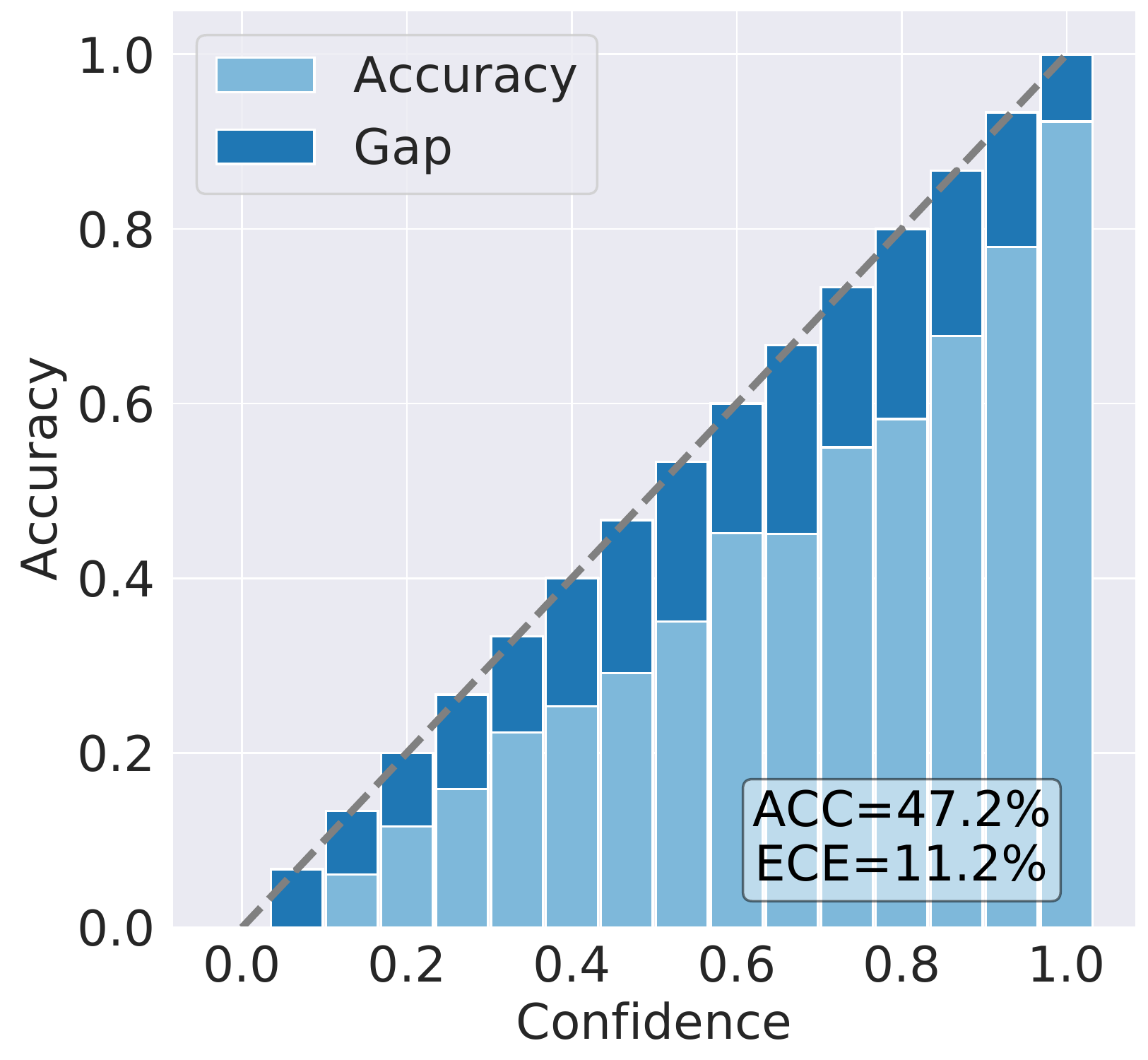}
    }
  \caption{Reliability diagrams of ResNet-32 trained on CIFAR-100-LT with imbalance ratio 100 under different weight decays (wds). (a-d), (e-h) and (i-l) denotes the results by CE, CE+DRW and balanced Softmax \cite{Ren2020balms}, respectively.}
  \label{fig:ece}
\end{figure*}

\begin{table}[htbp]
\centering
\caption{Top-1 validation accuracy on CIFAR-100-LT with imbalance ratio 0.01 under different values of weight decay in the SGD optimizer. We keep everything else the same except the weight decay. }
\label{tab:wd}
\begin{tabular}{c|cccccccccc}
    \toprule
     \multirow{2}{*}{Method} &  \multicolumn{10}{c}{weight decay}\\
     \cmidrule{2-11}
      & 0 & 5e-5 & 1e-4 & 2e-4 & 3e-4 & 5e-4 & 7e-4 & 1e-3 & 5e-3 & 1e-2\\
      \midrule
      CE & 37.36 & 37.31 & 37.76 & 38.69 & 39.65 & 40.49 & 40.61 & 40.78 & 24.10 & 13.19\\
      CE+DRW & 38.16 & 38.88 & 39.41 & 40.59 & 42.20 & 43.19 & 44.68 & 46.18 & 31.49 & 18.10\\
      Balanced Softmax & 38.27 & 39.31 & 41.16 & 42.68 & 43.89 & 45.19 & 46.71 & 47.21 & 30.02 & 17.5\\
     \bottomrule
\end{tabular}
\end{table}
\subsection{Rethinking the Role of Weight Decay in Mitigating Class Imbalance}
\label{role-of-weight-decay-in-imbalanced-learning}
 Based on the restriction of the norms of features and classifier weights, we rethink the role of weight decay that is often overlooked in mitigating class imbalance. As is well-known, a simple weight decay can improve generalization, since it suppresses any irrelevant components of model weights by choosing weights with the smallest norms when solving the learning problem \cite{krogh1992simple}. Actually, weight decay can facilitate not only explicitly learning smaller weights of networks, but also implicitly resulting in smaller feature representations. For a network $f(x)=L_n(L_{n-1}(\cdots L_2(L_1(x);W_1);W_2)\cdots;W_n)$ with bounded input $x$ and weights $W_1,...,W_n$, its output $f(x)$ is usually bounded by $W_1,...,W_n$. Back to long-tailed recognition, \textbf{an appropriately large weight decay can mitigate class imbalance} by preventing the excessively large norms of prototypes and features that cause over confidence, since networks trained on imbalanced datasets are usually miscalibrated and over-confident due to the imbalanced composition ratio of each class \cite{zhong2021improving}.

As shown in \cref{fig:CE-weights}, the norms of both prototypes obviously decrease as the weight decay increases, and the distribution of feature norms gradually moves to the left in \cref{fig:CE-vhist}. This is beneficial to alleviate the over-confident results and improve the performance of classification. We conduct an experiment on CIFAR-100-LT with imbalance ratio 100 to validate that an appropriately large weight decay can obviously improve confidence and accuracy, where \cref{fig:ece} shows that an appropriate large weight decay can not only largely enhance the network calibration but also greatly improve the performance for long-tailed recognition. For example, CE+DRW with weight decay $1e-3$ has obviously larger accuracy and smaller expected calibration error (ECE) than CE+DRW with weight decay $5e-5$. However, the fitting ability decreases as weight decay increases. as shown in \cref{tab:wd}.

Moreover, another easily overlooked to restrict weights and features is batch normalization (BatchNorm) \cite{ioffe2015batch}, which intuitively limits the distribution of features.  \citet{santurkar2018does} also prove that BatchNorm has minimax bound on weight-space Lipschitzness. Recall to the softmax loss in \cref{softmax-loss},  $\bm{z}_i$ and $\bm{w}_j$ implicitly have a small norm constraint with the role of both batch normalization and weight decay, \textit{i.e.}, $\exists r\in\sR$, \textit{s.t.}, $\|\bm{z}_i\|,\|\bm{w}_j\|\le r$, which shows that the domain of $\bm{z}_i$ and $\bm{w}_j$ belongs to a $2$-norm ball.

\subsection{Prototypes Generation}
To generate these prototypes, we randomly initialize $\{\bm{z}_i\}_{i=1}^N$ and $\{\bm{w}_i\}_{i=1}^k$ in $L_{\bm{\alpha}}$ ($\bm{\alpha}=0$) with a balanced setting (\textit{i.e.}, $N=k$ and $y_i=i$), and then directly minimize $L_{\bm{\alpha}}$ to obtain the optimal prototypes $\{\bm{w}_i\}_{i=1}^k$ according to Theorem \ref{balanced-case-for-per-margin}. We train $\{\bm{z}_i\}_{i=1}^N$ and $\{\bm{w}_i\}_{i=1}^k$ by SGD optimizer with momentum 0.9 and initial learning rate 0.1 for 100,000 epochs. We also use cosine learning rate annealing with $T_{\max}=20000$. The pseudo-code can be seen in \cref{pseudocode}
\begin{table}[htbp]
\caption{Pseudocode of prototypes generation in a PyTorch-like style}
\label{pseudocode}
\begin{python}
def generate_weight(n_classes, n_hiddens, use_relu=False):
    n_samples = n_classes
    scale = 5
    Z = torch.randn(n_samples, n_hiddens).cuda()
    Z.requires_grad = True
    W = torch.randn(n_classes, n_hiddens).cuda()
    W.requires_grad = True
    nn.init.kaiming_normal_(W)

    optimizer = SGD([Z, W], lr=0.1, momentum=0.9, weight_decay=1e-4)
    scheduler = CosineAnnealingLR(optimizer, T_max=20000, eta_min=0)

    criterion = nn.CrossEntropyLoss()
    for i in range(epochs):
        if use_relu:
            z = F.relu(Z)
        else:
            z = Z
        w = W
        L2_z = F.normalize(z, dim=1)
        L2_w = F.normalize(w, dim=1)
        out = F.linear(L2_z, L2_w)
        loss = criterion(out * scale, labels)
        optimizer.zero_grad()
        loss.backward()
        optimizer.step()
        scheduler.step()
    return W
\end{python}
\end{table}

\subsection{Details of Imbalanced Learning}
\paragraph{Imbalanced CIFAR-10 and CIFAR-100.} The original version of CIFAR-10 and CIFAR-100 contains 50,000 training images and 10,000 test images of size $32\times 32$ with 10 and 100 classes, respectively. To create their imbalanced version, we follow the setting in \citep{buda2018systematic, cui2019class, cao2019learning}, where we reduce the number of training examples per class, and keep the test set unchanged. To ensure that our methods apply to a variety of settings, we consider two types of imbalance: long-tailed imbalance \citep{cui2019class} and step imbalance \citep{buda2018systematic}. We use the imbalance ratio $\rho$ to denote the ratio between sample sizes of the most frequent and least frequent class, \textit{i.e.}, $\rho=\max_i\{n_i\}/\min_i\{n_i\}$. Long-tailed imbalance utilizes an exponential decay in sample sizes across different classes.

\paragraph{Networks and training.} We follow the settings in \cite{zhong2021improving}: ResNet-32 for CIFAR-100-LT and ResNet-50 for ImageNet-LT, but we train MiSLAS on CIFAR-100 with initial learning rate 0.15 and weight decay $5e-4$ for stage-1, and initial learning rate 0.01 for stage-2. The experiments with the same settings in \cite{cao2019learning} are reported in \cref{tab:cifar-imb}.

\begin{table*}[htbp]
\caption{Top-1 validation accuracy of ResNet-32 on imbalanced CIFAR-10 and CIFAR-100.}
    \label{tab:cifar-imb}
    \centering
    \begin{tabular}{c|cc|cc|cc|cc}
         \toprule
         Dataset & \multicolumn{4}{c|}{Imbalanced CIFAR-10} & \multicolumn{4}{c}{Imbalanced CIFAR-100}\\
         \midrule
         Imbalance Type  & \multicolumn{2}{c|}{long-tailed} & \multicolumn{2}{c|}{step} & \multicolumn{2}{c|}{long-tailed} & \multicolumn{2}{c|}{step}\\
         \midrule
         Imbalance Ratio & 100 & 10 & 100 & 10 & 100 & 10 & 100 & 10\\
         \midrule
         CE & 70.68 & 86.63 & 64.54 & 84.61 & 38.59 & 56.38 & 38.90 & 54.71\\
         CE-PAL & 70.60 & 86.39 & 62.81 & 84.58 & \textbf{39.13} & \textbf{57.89} & \textbf{39.15} & \textbf{55.47}\\
         CE-DRW & 74.85 & 87.69 & 69.14 & 86.89 & 39.38 & 58.01 & 42.64 & 58.45\\
         CE-DRW-PAL & 74.14 & \textbf{87.72} & \textbf{70.18} & \textbf{87.18} & \textbf{42.48} & \textbf{58.66} & \textbf{44.57} & \textbf{60.14}\\
         \midrule
         Norm & 70.65 & 86.15 & 62.39 & 84.41 & 35.68 & 54.79 & 39.72 & 53.76\\
         Norm-PAL & \textbf{71.27} & \textbf{86.83} & \textbf{62.81} & \textbf{85.05} & \textbf{38.14} & \textbf{56.09} & 39.28 & \textbf{54.10}\\
         Norm-DRW & 73.38 & 87.38 & 68.52 & 86.22 & 38.04 & 56.89 & 42.23 & 56.90\\
         Norm-DRW-PAL & \textbf{74.27} & \textbf{87.67} & \textbf{69.28} & \textbf{87.00} & \textbf{39.58} & \textbf{56.78} & \textbf{43.08} & \textbf{58.25}\\
         \midrule
         LDAM & 73.53 & 86.14 & 64.87 & 84.84 & 37.23 & 57.36 & 39.82 & 54.01\\
         LDAM-PAL & \textbf{73.66} & \textbf{86.79} & \textbf{65.37} & \textbf{85.57} & \textbf{39.70} & \textbf{57.76} & 39.09 & 53.30\\
         LDAM-DRW & 76.00 & 87.78 & 72.22 & 87.69 & 41.28 & 58.98 & 45.43 & 59.11\\
         LDAM-DRW-PAL & \textbf{77.79} & \textbf{87.82} & \textbf{75.54} & \textbf{88.30} & \textbf{42.98} & \textbf{59.26} & \textbf{46.17} & 58.69\\
         \bottomrule
    \end{tabular}

\end{table*}

\subsection{Details of Learning with Noisy Labels}
\paragraph{Noise generation.} The noisy labels are generated following standard approaches in previous works \cite{ma2020normalized, 8099723}. For symmetric noise, we corrupt the training labels by flipping labels in each class randomly to incorrect labels to other classes with flip probability $\eta\in\{0.2, 0.3, 0.6, 0.8\}$. For asymmetric noise, we flip the labels within a specific set of classes. For MNIST, flipping 7 $\rightarrow$ 1, 2 $\rightarrow$ 7, 5 $\leftrightarrow$ 6, 3 $\rightarrow$ 8. For CIFAR-10, flipping TRUCK $\rightarrow$ AUTOMOBILE, BIRD $\rightarrow$ AIRPLANE, DEER $\rightarrow$ HORSE, CAR $\leftrightarrow$ DOG. For CIFAR-100, the 100 classes are grouped into 20 super-classes with each having 5 sub-classes, and each class are flipped  within the same super-class into the next in a circular fashion.

\paragraph{Networks and training.} We follow the settings in \cite{ma2020normalized}: 4-layer CNN for MNIST, an 8-layer CNN for CIFAR-10 and a ResNet-34 \cite{he2016deep} for CIFAR-100. The networks are trained for 50, 120, 200 epochs for MNIST, CIFAR-10, CIFAR-100, respectively. For all the training, we use SGD optimizer with momentum 0.9 and cosine learning rate annealing. The initial learning rate is set to $0.01$, $0.01$ and  $0.1$ for MNIST, CIFAR-10 and CIFAR-100, respectively. Weight decay is set to $1\times 10^{-3}$, $1\times 10^{-4}$ and $1\times 10^{-5}$ for MNIST, CIFAR-10 and CIFAR-100, respectively. The initial learning rate is set to 0.01 for MNIST/CIFAR-10 and 0.1 for CIFAR-100. Batch size is set to 128. Typical data augmentations including random width/height shift and horizontal flip are applied.

\begin{figure}
    \centering
    \includegraphics[width=2in]{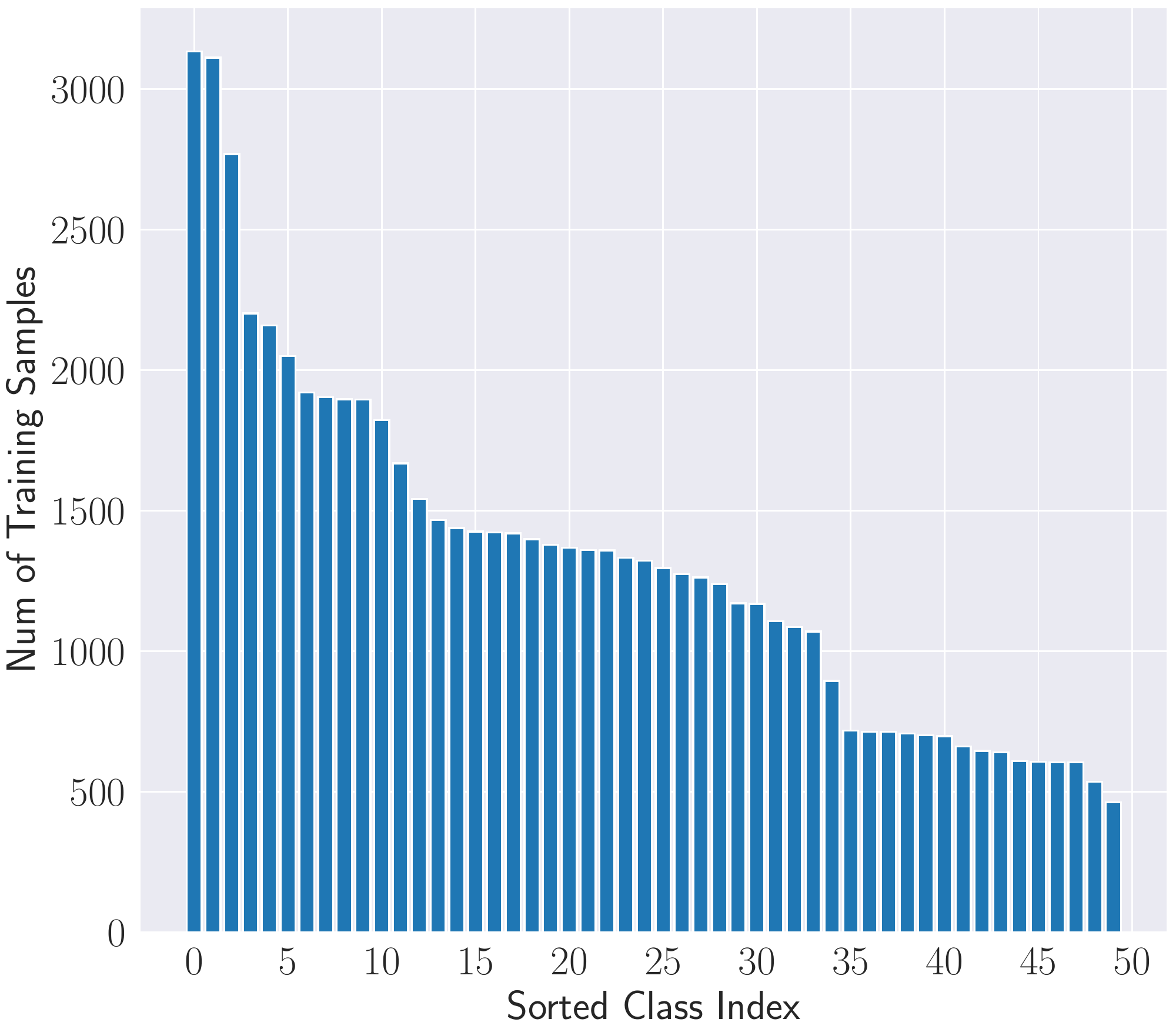}
    \caption{The label distribution on mini-WebVision.}
    \label{fig:webvision-stat}
\end{figure}
\subsection{Details of experiments on WebVision}
WebVision 1.0 \cite{li2017webvision} contains 2.4 million images of real-world noisy labels, crawled from the web using 1,000 concepts in ImageNet ILSVRC12. Since the dataset is very big, for quick experiments, we follow the training setting in \cite{jiang2018mentornet, ma2020normalized} that only takes the first 50 classes of the Google resized image subset. We evaluate the trained networks on the same 50 classes of WebVision 1.0 validation set, where each concept in validation set contains 50 instances and the labels can be considered as a clean validation. ResNet-50 \cite{he2016deep} is the model to be learned. The distribution of labels in mini-WebVision is shown in \cref{fig:webvision-stat}, where the imbalance ratio is $3134/462$.

We add our proposed PAL and FNPAL in some commonly-used but not sufficiently robust loss functions, such as CE, Focal loss (FL), SCE and compare them with GCE, SCE and NCE+RCE. The training details follow \cite{ma2020normalized}, where for each method, we train a ResNet-50 \cite{he2016deep} using SGD for 250 epochs with initial learning rate 0.4, nesterov momentum 0.9 and weight decay $3\times 10^{-5}$ and batch size $256$. The learning rate is multiplied by $0.97$ after every epoch of training. All the images are resized to $224 \times 224$. Typical data augmentations including random width/height shift, color jittering, and random horizontal flip are applied. More detailed results than \Cref{lnl-webvision} are reported in \Cref{full-lnl-webvision}.

\begin{table*}[htbp]
\small
\centering
\caption{Top-1 validation accuracies (\%) on WebVision validation set using different loss functions.}
\label{full-lnl-webvision}
\begin{tabular}{c|cccccccccc}
    \toprule
    Method & CE & FL & GCE & SCE & NCE+RCE & \textbf{NSL} & \textbf{CE+PAL} & \textbf{CE+FNPAL} & \textbf{FL+FNPAL} & \textbf{SCE+FNPAL}\\
    \midrule
    Accuracy & 62.60 & 63.80 & 61.76 & 66.92 & 66.32 & \textbf{69.56} & \textbf{68.92} & \textbf{69.69} & \textbf{69.64} & \textbf{69.92} \\ 
    \bottomrule
\end{tabular}
\end{table*}

\begin{table*}[!t]
\small
\centering
\caption{Validation accuracies (\%) of different methods on benchmark datasets with clean or symmetric label noise ($\eta\in[0.2, 0.4, 0.6, 0.8]$). The results (mean$\pm$std) are reported over 3 random runs. The results with positive gains are \textbf{boldfaced} and the best one is \underline{underlined}.}
\label{full-symmetric-noise}
\begin{tabular}{c|c|c|cccc}
    \toprule
     \multirow{2}*{Dataset} & \multirow{2}*{Method} & \multirow{2}*{Clean ($\eta=0.0$)} & \multicolumn{4}{c}{Symmetric Noise Rate ($\eta$)}  \\
     ~ & ~ & ~ & 0.2 & 0.4 & 0.6 & 0.8\\
     \midrule\midrule
     \multirow{10}*{MNIST}
      & CE & 99.17 $\pm$ 0.04 & 91.40 $\pm$ 0.11 & 74.36 $\pm$ 0.29 & 49.32 $\pm$ 0.70 & 22.32 $\pm$ 0.15\\
    ~ & FL & 99.16 $\pm$ 0.02 & 91.49 $\pm$ 0.20 & 75.28 $\pm$ 0.10 & 50.25 $\pm$ 0.70 & 22.68 $\pm$ 0.14 \\
    ~ & GCE & 99.15 $\pm$ 0.02 & 98.90 $\pm$ 0.03 & 96.81 $\pm$ 0.23 & 81.39 $\pm$ 0.64 & 33.07 $\pm$ 0.31\\
    ~ & SCE & 99.28 $\pm$ 0.07 & 98.91 $\pm$ 0.12 & 97.60 $\pm$ 0.22 & 88.00 $\pm$ 0.50 & 47.32 $\pm$ 0.99\\
    ~ & NCE+MAE & 99.42 $\pm$ 0.02 & 99.18 $\pm$ 0.08 & 98.47 $\pm$ 0.21 & 95.52 $\pm$ 0.04 & 73.05 $\pm$ 0.59 \\
    ~ & NCE+RCE & 99.40 $\pm$ 0.04 & \underline{99.24} $\pm$ 0.01 & 98.44 $\pm$ 0.11 & 95.77 $\pm$ 0.09 & 74.80 $\pm$ 0.28\\
    ~ & NFL+RCE & 99.37 $\pm$ 0.01 & 99.16 $\pm$ 0.03 & 98.55 $\pm$ 0.05 & 95.62 $\pm$ 0.24 & 74.67 $\pm$ 0.97\\
    \cmidrule{2-7}
    ~ & NSL & 99.24 $\pm$ 0.03 & 98.99 $\pm$ 0.03 & 98.58 $\pm$ 0.11 & 95.99 $\pm$ 0.24 & 59.77 $\pm$ 1.98\\
    ~ & \textbf{CE+FNPAL} & 99.24 $\pm$ 0.05 & \textbf{99.05 $\pm$ 0.04} & \textbf{98.66 $\pm$ 0.04} & \textbf{97.62 $\pm$ 0.15} & \textbf{79.23 $\pm$ 0.87}\\
    ~ & \textbf{FL+FNPAL} & 99.16 $\pm$ 0.03 & \textbf{98.97 $\pm$ 0.09} & \textbf{98.61 $\pm$ 0.07} & \textbf{97.69 $\pm$ 0.03} & \textbf{80.63 $\pm$ 0.28}\\
    ~ & \textbf{GCE+FNPAL} & 99.07 $\pm$ 0.06 & \textbf{99.04 $\pm$ 0.01} & \textbf{98.62 $\pm$ 0.06} & \textbf{97.89 $\pm$ 0.13} & \textbf{94.65 $\pm$ 0.52}\\
    ~ & \textbf{SCE+FNPAL} & 99.27 $\pm$ 0.04 & \textbf{99.06 $\pm$ 0.05} & \underline{\textbf{98.76 $\pm$ 0.09}} & \underline{\textbf{97.94 $\pm$ 0.07}} & \underline{\textbf{88.56 $\pm$ 1.07}}\\
    ~ & \textbf{NCE+MAE+FNPAL} & 99.34 $\pm$ 0.06 & 99.10 $\pm$ 0.09 & 98.40 $\pm$ 0.09 & \textbf{95.70 $\pm$ 0.05} & \textbf{80.21 $\pm$ 0.43}\\
    ~ & \textbf{NCE+RCE+FNPAL} & 99.29 $\pm$ 0.04 & 99.04 $\pm$ 0.07 & 98.11 $\pm$ 0.09 & 94.84 $\pm$ 0.08 & \textbf{79.70 $\pm$ 1.06}\\
    ~ & \textbf{NFL+RCE+FNPAL} & 99.29 $\pm$ 0.06 & 99.02 $\pm$ 0.05 & 98.32 $\pm$ 0.14 & 95.38 $\pm$ 0.11 & \textbf{76.06 $\pm$ 0.58}\\
    \midrule\midrule
    \multirow{10}*{CIFAR10} 
      & CE & 90.36 $\pm$ 0.25 & 74.78 $\pm$ 0.68 & 57.95 $\pm$ 0.12 & 38.21 $\pm$ 0.12 & 18.89 $\pm$ 0.43\\
    ~ & FL & 89.69 $\pm$ 0.25 & 74.19 $\pm$ 0.23 & 57.35 $\pm$ 0.27 & 38.11 $\pm$ 0.76 & 19.39 $\pm$ 0.44\\
    ~ & GCE & 89.37 $\pm$ 0.29 & 87.05 $\pm$ 0.21 & 82.43 $\pm$ 0.10 & 68.05 $\pm$ 0.07 & 25.21 $\pm$ 0.28\\
    ~ & SCE & 91.24 $\pm$ 0.19 & 87.34 $\pm$ 0.01 & 79.84 $\pm$ 0.43 & 61.09 $\pm$ 0.19 & 27.19 $\pm$ 0.34\\
    ~ & NCE+MAE & 89.02 $\pm$ 0.09 & 87.06 $\pm$ 0.17 & 83.92 $\pm$ 0.16 & 76.47 $\pm$ 0.25 & 45.01 $\pm$ 0.31\\
    ~ & NCE+RCE & 91.12 $\pm$ 0.14 & 89.21 $\pm$ 0.00 & 86.03 $\pm$ 0.14 & 80.04 $\pm$ 0.26 & 51.67 $\pm$ 1.38\\
    ~ & NFL+RCE & 91.03 $\pm$ 0.15 & 89.10 $\pm$ 0.16 & 86.20 $\pm$ 0.19 & 79.58 $\pm$ 0.08 & 50.03 $\pm$ 2.78\\
    \cmidrule{2-7}
    ~ & NSL & 88.07 $\pm$ 0.12 & 86.46 $\pm$ 0.02 & 83.27 $\pm$ 0.13 & 76.17 $\pm$ 0.40 & 46.74 $\pm$ 0.72\\
    ~ & \textbf{CE+FNPAL} & 90.69 $\pm$ 0.11 & \textbf{86.34 $\pm$ 0.37} & \textbf{81.30 $\pm$ 0.29} & \textbf{72.77 $\pm$ 0.41} & \textbf{51.46 $\pm$ 1.10} \\
    ~ & \textbf{FL+FNPAL} & 90.02 $\pm$ 0.26 & \textbf{86.17 $\pm$ 0.49} & \textbf{80.96 $\pm$ 0.15} & \textbf{72.36 $\pm$ 0.34} & \textbf{51.54 $\pm$ 0.13}\\
    ~ & \textbf{SCE+FNPAL} & 91.11 $\pm$ 0.13 & 87.30 $\pm$ 0.06 & \textbf{82.68 $\pm$ 0.22} & \textbf{73.49 $\pm$ 0.42} & \textbf{51.99 $\pm$ 1.10}\\
    ~ & \textbf{NCE+MAE+FNPAL} & 89.18 $\pm$ 0.05 & \textbf{87.26 $\pm$ 0.07} & \textbf{84.55 $\pm$ 0.02} & \textbf{78.39 $\pm$ 0.49} & \textbf{52.15 $\pm$ 0.65}\\
    ~ & \textbf{NCE+RCE+FNPAL} & 90.88 $\pm$ 0.10 & \textbf{89.34 $\pm$ 0.15} & \textbf{86.65 $\pm$ 0.21} & \textbf{80.28 $\pm$ 0.07} & \underline{\textbf{57.21 $\pm$ 0.22}}\\
    ~ & \textbf{NFL+RCE+FNPAL} & 91.16 $\pm$ 0.25 & \underline{\textbf{89.49 $\pm$ 0.32}} & \underline{\textbf{86.66 $\pm$ 0.08}} & \underline{\textbf{80.33 $\pm$ 0.15}} & \textbf{56.23 $\pm$ 0.15}\\
     \midrule\midrule
     \multirow{10}*{CIFAR100} 
       & CE & 70.41 $\pm$ 1.17 & 55.64 $\pm$ 0.17 & 40.39 $\pm$ 0.46 & 22.00 $\pm$ 1.23 & 7.37 $\pm$ 0.16\\
     ~ & FL & 70.56 $\pm$ 0.59 & 56.02 $\pm$ 0.80 & 40.41 $\pm$ 0.39 & 22.11 $\pm$ 0.30 & 7.70 $\pm$ 0.20\\
     ~ & GCE & 63.06 $\pm$ 1.00 & 62.15 $\pm$ 0.66 & 57.11 $\pm$ 1.43 & 45.99 $\pm$ 1.00 & 18.32 $\pm$ 0.36\\
     ~ & SCE & 70.41 $\pm$ 0.63 & 55.05 $\pm$ 0.68 & 39.60 $\pm$ 0.14 & 21.53 $\pm$ 0.72 & 7.82 $\pm$ 0.30 \\
     ~ & NCE+MAE & 67.16 $\pm$ 0.13 & 52.34 $\pm$ 0.12 & 35.81 $\pm$ 0.42 & 19.29 $\pm$ 0.29 & 7.31 $\pm$ 0.23\\
     ~ & NCE+RCE & 68.09 $\pm$ 0.26 & 64.32 $\pm$ 0.40 & 58.11 $\pm$ 0.63 & 45.94 $\pm$ 1.31 & \underline{25.22 $\pm$ 0.08}\\
     ~ & NFL+RCE & 67.58 $\pm$ 0.39 & 64.48 $\pm$ 0.50 & 57.86 $\pm$ 0.12 & 46.74 $\pm$ 0.59 & 24.55 $\pm$ 0.47\\
     \cmidrule{2-7}
     ~ & NSL & 70.08 $\pm$ 0.19 & 65.30 $\pm$ 0.36 & 56.77 $\pm$ 0.52 & 41.21 $\pm$ 1.01 & 12.16 $\pm$ 0.96\\
     ~ & \textbf{CE+FNPAL} & 71.69 $\pm$ 0.27 & \textbf{65.38 $\pm$ 0.17} & \textbf{57.24 $\pm$ 0.36} & \textbf{41.35 $\pm$ 0.19} & \textbf{12.12 $\pm$ 0.88}\\
     ~ & \textbf{FL+FNPAL} & 71.79 $\pm$ 0.26 & \textbf{65.24 $\pm$ 0.05} & \textbf{56.05 $\pm$ 0.15} & \textbf{40.41 $\pm$ 0.47} & \textbf{13.16 $\pm$ 0.25}\\
     ~ & \textbf{GCE+FNPAL} & 70.22 $\pm$ 0.05 & \textbf{63.40 $\pm$ 0.12} & 50.86 $\pm$ 0.38 & 34.01 $\pm$ 0.44 & 15.08 $\pm$ 0.47\\
     ~ & \textbf{SCE+FNPAL} & 70.87 $\pm$ 0.45 & \textbf{65.30 $\pm$ 0.15} & \textbf{55.10 $\pm$ 0.45} & \textbf{39.73 $\pm$ 0.04} & \textbf{11.70 $\pm$ 0.53}\\
     ~ & \textbf{NCE+MAE+FNPAL} & 67.30 $\pm$ 0.07 & \textbf{62.00 $\pm$ 0.20} & \textbf{52.03 $\pm$ 0.18} & \textbf{33.78 $\pm$ 0.12} & \textbf{10.92 $\pm$ 0.77}\\
     ~ & \textbf{NCE+RCE+FNPAL} & 69.29 $\pm$ 0.32 & \textbf{65.53 $\pm$ 0.30} & \textbf{60.53 $\pm$ 0.27} & \textbf{49.73 $\pm$ 0.64} & 24.54 $\pm$ 0.28\\
     ~ & \textbf{NFL+MAE+FNPAL} & 69.53 $\pm$ 0.05 & \underline{\textbf{65.94 $\pm$ 0.32}} & \underline{\textbf{60.89 $\pm$ 0.60}} & \underline{\textbf{50.10 $\pm$ 0.40}} & 24.15 $\pm$ 1.06\\
     \bottomrule
\end{tabular}
\vskip-15pt
\end{table*}

\begin{table*}[!t]
\small
\centering
\caption{Validation accuracies (\%) of different methods on benchmark datasets with asymmetric label noise ($\eta\in[0.1, 0.2, 0.3, 0.4]$). The results (mean$\pm$std) are reported over 3 random runs. The results with positive gains are \textbf{boldfaced} and the best one is \underline{underlined}.}
\label{full-asymmetric-noise}
\begin{tabular}{c|c|cccc}
    \toprule
     \multirow{2}*{Dataset} & \multirow{2}*{Method} & \multicolumn{4}{c}{Asymmetric Noise Rate ($\eta$)}  \\
     ~ & ~ & 0.1 & 0.2 & 0.3 & 0.4\\
     \midrule\midrule
     \multirow{9}*{MNIST} & 
        CE & 97.66 $\pm$ 0.05 & 94.42 $\pm$ 0.28 & 88.74 $\pm$ 0.07 & 81.45 $\pm$ 0.09\\
    ~ & FL & 97.68 $\pm$ 0.20 & 94.18 $\pm$ 0.05 & 88.87 $\pm$ 0.30 & 81.79 $\pm$ 0.36\\
    ~ & GCE & 99.05 $\pm$ 0.03 & 96.57 $\pm$ 0.15 & 89.20 $\pm$ 0.11 & 81.92 $\pm$ 0.45\\
    ~ & SCE & 99.17 $\pm$ 0.02 & 98.13 $\pm$ 0.17 & 93.59 $\pm$ 0.24 & 84.42 $\pm$ 0.57\\
    ~ & NCE+MAE & 99.28 $\pm$ 0.08 & 98.87 $\pm$ 0.10 & 97.00 $\pm$ 0.15 & 91.33 $\pm$ 0.65\\
    ~ & NCE+RCE & 99.26 $\pm$ 0.06 & 98.91 $\pm$ 0.08 & 97.37 $\pm$ 0.20 & 86.95 $\pm$ 4.27\\
    ~ & NFL+RCE & \underline{99.33 $\pm$ 0.02} & 98.87 $\pm$ 0.05 & 97.41 $\pm$ 0.12 & 92.35 $\pm$ 0.26\\
    \cmidrule{2-6}
    ~ & NSL & 99.15 $\pm$ 0.02 & 98.79 $\pm$ 0.05 & 95.85 $\pm$ 0.12 & 87.41 $\pm$ 0.62\\
     ~ & \textbf{CE+FNPAL} & \textbf{98.99 $\pm$ 0.01} & \textbf{98.58 $\pm$ 0.05} & \textbf{96.94 $\pm$ 0.27} & \textbf{91.77 $\pm$ 0.94}\\
     ~ & \textbf{FL+FNPAL} & \textbf{99.02 $\pm$ 0.09} & \textbf{98.52 $\pm$ 0.06} & \textbf{97.06 $\pm$ 0.05} & \textbf{92.17 $\pm$ 0.4}4\\
     ~ & \textbf{GCE+FNPAL} & 99.11 $\pm$ 0.07 & 99.05 $\pm$ 0.03 & 98.82 $\pm$ 0.09 & 97.27 $\pm$ 0.23\\
     ~ & \textbf{SCE+FNPAL} & \textbf{99.23 $\pm$ 0.05} & \underline{\textbf{99.08 $\pm$ 0.02}} & \underline{\textbf{98.43 $\pm$ 0.13}} & \textbf{92.11 $\pm$ 0.65}\\
     ~ & \textbf{NCE+MAE+FNPAL} & 99.23 $\pm$ 0.09 & \textbf{99.00 $\pm$ 0.01} & \textbf{97.71 $\pm$ 0.20} & \textbf{93.95 $\pm$ 0.43}\\
     ~ & \textbf{NCE+RCE+FNPAL} & \textbf{99.26 $\pm$ 0.05} & \textbf{98.93 $\pm$ 0.04} & \textbf{97.91 $\pm$ 0.29} & \textbf{94.37 $\pm$ 0.54}\\
     ~ & \textbf{NFL+RCE+FNPAL} & 99.17 $\pm$ 0.10 & \textbf{99.02 $\pm$ 0.10} & \textbf{98.18 $\pm$ 0.19} & \underline{\textbf{94.86 $\pm$ 0.18}}\\
    \midrule\midrule
     \multirow{9}*{CIFAR-10} & CE & 86.89 $\pm$ 0.04 & 83.34 $\pm$ 0.32 & 79.58 $\pm$ 0.22 & 74.61 $\pm$ 0.42\\
     ~ & FL & 86.36 $\pm$ 0.30 & 83.31 $\pm$ 0.23 & 79.67 $\pm$ 0.08 & 74.19 $\pm$ 0.40\\
     ~ & GCE & 88.25 $\pm$ 0.47 & 85.91 $\pm$ 0.02 & 80.56 $\pm$ 0.30 & 74.49 $\pm$ 0.34\\
     ~ & SCE & 89.73 $\pm$ 0.22 & 86.47 $\pm$ 0.14 & 81.61 $\pm$ 0.17 & 75.52 $\pm$ 0.28\\
     ~ & NCE+MAE & 88.41 $\pm$ 0.28 & 86.52 $\pm$ 0.28 & 83.74 $\pm$ 0.19 & 76.75 $\pm$ 0.33\\
    ~ & NCE+RCE & \underline{90.19 $\pm$ 0.20} & 88.61 $\pm$ 0.27 & 85.55 $\pm$ 0.01 & 79.25 $\pm$ 0.33\\
    ~ & NFL+RCE & 89.97 $\pm$ 0.17 & 88.58 $\pm$ 0.15 & 85.53 $\pm$ 0.26 & 79.61 $\pm$ 0.07\\
     \cmidrule{2-6}
     ~ & NSL & 87.50 $\pm$ 0.04 & 86.23 $\pm$ 0.23 & 83.56 $\pm$ 0.15 & 78.63 $\pm$ 0.28\\
     ~ & \textbf{CE+FNPAL} & \textbf{88.55 $\pm$ 0.11} & \textbf{86.11 $\pm$ 0.19} & \textbf{83.06 $\pm$ 0.06} & \textbf{77.41 $\pm$ 0.27}\\
     ~ & \textbf{FL+FNPAL} & \textbf{88.12 $\pm$ 0.18} & \textbf{85.56 $\pm$ 0.11} & \textbf{82.90 $\pm$ 0.09} & \textbf{76.74 $\pm$ 0.52}\\
     ~ & \textbf{SCE+FNPAL} & 89.24 $\pm$ 0.09 & \textbf{87.32 $\pm$ 0.21} & \textbf{84.18 $\pm$ 0.10} & \textbf{78.15 $\pm$ 0.34}\\
     ~ & \textbf{NCE+MAE+FNPAL} & 88.21 $\pm$ 0.04 & \textbf{87.13 $\pm$ 0.27} & \textbf{85.43 $\pm$ 0.41} & \textbf{78.96 $\pm$ 0.34}\\
     ~ & \textbf{NCE+RCE+FNPAL} & 89.94 $\pm$ 0.15 & \underline{\textbf{89.01 $\pm$ 0.18}} & \textbf{85.97 $\pm$ 0.19} & \textbf{79.51 $\pm$ 0.19}\\
     ~ & \textbf{NFL+RCE+FNPAL} & \textbf{90.05 $\pm$ 0.06} & \textbf{88.64 $\pm$ 0.17} & \underline{\textbf{86.20 $\pm$ 0.09}} & \underline{\textbf{79.66 $\pm$ 0.60}}\\
     \midrule\midrule
     \multirow{9}*{CIFAR-100} & CE & 65.03 $\pm$ 0.07 & 57.67 $\pm$ 0.89 & 50.63 $\pm$ 0.57 & 41.95 $\pm$ 0.12\\
     ~ & FL & 64.83 $\pm$ 1.13 & 58.37 $\pm$ 0.56 & 51.44 $\pm$ 0.79 & 42.16 $\pm$ 0.53\\
     ~ & {GCE} & 62.46 $\pm$ 0.50 & 59.56 $\pm$ 0.64 & 54.22 $\pm$ 1.18 & 42.18 $\pm$ 0.77\\
     ~ & SCE & 64.55 $\pm$ 0.58 & 58.15 $\pm$ 0.54 & 50.58 $\pm$ 0.27 & 41.57 $\pm$ 0.19\\
     ~ & NCE+MAE & 60.55 $\pm$ 0.42 & 52.03 $\pm$ 0.15 & 44.45 $\pm$ 0.22 & 36.86 $\pm$ 0.46 \\
    ~ & NCE+RCE & 66.31 $\pm$ 0.59 & 62.93 $\pm$ 0.42 & 55.70 $\pm$ 0.14 & 42.73 $\pm$ 0.58\\
    ~ & NFL+RCE & 66.41 $\pm$ 0.20 & 62.94 $\pm$ 0.39 & 55.83 $\pm$ 0.46 & 42.63 $\pm$ 0.13\\
     \cmidrule{2-6}
     ~ & NSL & 65.79 $\pm$ 0.27 & 58.16 $\pm$ 0.46 & 50.35 $\pm$ 0.41 & 40.38 $\pm$ 0.60\\
     ~ & \textbf{CE+FNPAL} & \textbf{65.99 $\pm$ 0.06} & \textbf{57.93 $\pm$ 0.18} & \textbf{51.45 $\pm$ 0.92} & \textbf{47.53 $\pm$ 0.94}\\
     ~ & \textbf{FL+FNPAL} &\textbf{ 65.54 $\pm$ 0.13} & \textbf{57.17 $\pm$ 0.43} & \textbf{51.74 $\pm$ 0.64} & \textbf{44.63 $\pm$ 0.14}\\
     ~ & \textbf{SCE+FNPAL} & \textbf{64.65 $\pm$ 0.37} & \textbf{57.61 $\pm$ 0.22} & \textbf{50.74 $\pm$ 0.51} & \textbf{44.31 $\pm$ 0.22}\\
     ~ & \textbf{NCE+MAE+FNPAL} & 60.53 $\pm$ 0.73 & \textbf{52.58 $\pm$ 0.38} & \textbf{44.96 $\pm$ 0.29} & \textbf{36.88 $\pm$ 0.40}\\
     ~ & \textbf{NCE+RCE+FNPAL} & \underline{\textbf{67.19 $\pm$ 0.10}} & \underline{\textbf{64.63 $\pm$ 0.32}} & \textbf{57.43 $\pm$ 0.53} & \textbf{49.77 $\pm$ 1.11}\\
     ~ & \textbf{NFL+RCE+FNPAL} & \textbf{67.09 $\pm$ 0.30} & \textbf{64.32 $\pm$ 0.59} & \underline{\textbf{57.92 $\pm$ 0.65}} & \underline{\textbf{50.19 $\pm$ 1.60}}\\
     \bottomrule
\end{tabular}
\vskip-15pt
\end{table*}

\end{document}